\documentclass[acmtog, dvipsnames]{acmart}

\usepackage{booktabs} 
\usepackage{multirow}
\usepackage{xcolor}
\usepackage{bm}

\usepackage{adjustbox}
\usepackage{array}
\newcolumntype{R}[2]{%
	>{\adjustbox{angle=#1,lap=\width-(#2)}\bgroup}%
	l%
	<{\egroup}%
}
\newcommand*\rot{\multicolumn{1}{R{45}{1em}}}
	
\usepackage{enumerate}
	
\usepackage[tight,footnotesize,sf,SF]{subfigure}
	
\acmSubmissionID{360}

\usepackage{pifont}
\newcommand{\cmark}{{\color{ForestGreen} \ding{51}}}%
\newcommand{\xmark}{{\color{red} \ding{55}}}%

\definecolor{forestgreen}{rgb}{0.13, 0.55, 0.13}
\definecolor{green(munsell)}{rgb}{0.0, 0.66, 0.47}
\definecolor{revised}{rgb}{0.9, 0.2, 0.0}
\definecolor{revised2}{rgb}{0.0, 0.8, 0.4}

\newcommand{\ie}{\textit{i.e.,}~}

\acmPrice{15.00}

\setcopyright{acmlicensed}
\acmJournal{TOG}
\acmYear{2020}
\acmVolume{39}
\acmNumber{6}
\acmArticle{218}
\acmMonth{12} \acmDOI{10.1145/3414685.3417852}

\setcitestyle{square}

\begin{document}

\title{RGB2Hands: Real-Time Tracking of 3D Hand Interactions from Monocular RGB Video}


 \author{Jiayi Wang}
 \author{Franziska Mueller}
 \affiliation{%
   \institution{MPI Informatics, Saarland Informatics Campus}}
 \author{Florian Bernard}
 \affiliation{%
   \institution{MPI Informatics, Saarland Informatics Campus, Technical University of Munich}}
 \author{Suzanne Sorli}
 \affiliation{%
   \institution{Universidad Rey Juan Carlos}}
 \author{Oleksandr Sotnychenko}
 \author{Neng Qian}
 \affiliation{%
   \institution{MPI Informatics, Saarland Informatics Campus}}
 \author{Miguel A. Otaduy}
 \author{Dan Casas}
 \affiliation{%
   \institution{Universidad Rey Juan Carlos}}

 \author{Christian Theobalt}
 \affiliation{%
   \institution{MPI Informatics, Saarland Informatics Campus}}
   
\renewcommand{\shortauthors}{}

\begin{abstract}
Tracking and reconstructing the 3D pose and geometry of two hands in interaction is a challenging problem that has a high relevance for several human-computer interaction applications, including AR/VR, robotics, or sign language recognition. Existing works are either limited to simpler tracking settings (\textit{e.g.}, considering only a single hand or two spatially separated hands), or rely on less ubiquitous sensors, such as depth cameras. In contrast, in this work we present the first real-time method for motion capture of skeletal pose and 3D surface geometry of hands from a single RGB camera that explicitly considers close interactions. In order to address the inherent depth ambiguities in RGB data, we propose a novel multi-task CNN that regresses multiple complementary pieces of information, including segmentation, dense matchings to a 3D hand model, and 2D keypoint positions, together with newly proposed intra-hand relative depth and inter-hand distance maps. These predictions are subsequently used in a generative model fitting framework in order to estimate pose and shape parameters of a 3D hand model for both hands. We experimentally verify the individual components of our RGB two-hand tracking and 3D reconstruction pipeline through an extensive ablation study. Moreover, we demonstrate that our approach offers previously unseen two-hand tracking performance from RGB, and quantitatively and qualitatively outperforms existing RGB-based methods that were not explicitly designed for two-hand interactions. Moreover, our method even performs on-par with depth-based real-time methods.
\end{abstract}

\ccsdesc[500]{Computing methodologies~Tracking}
\ccsdesc[300]{Computing methodologies~Computer vision}
\ccsdesc[100]{Computing methodologies~Neural networks}

\keywords{hand tracking, hand pose estimation, hand reconstruction, two hands, monocular RGB, RGB video, computer vision}

\begin{teaserfigure}
  \centering
  \includegraphics[width=\textwidth,trim={0 12cm 0 0},clip]{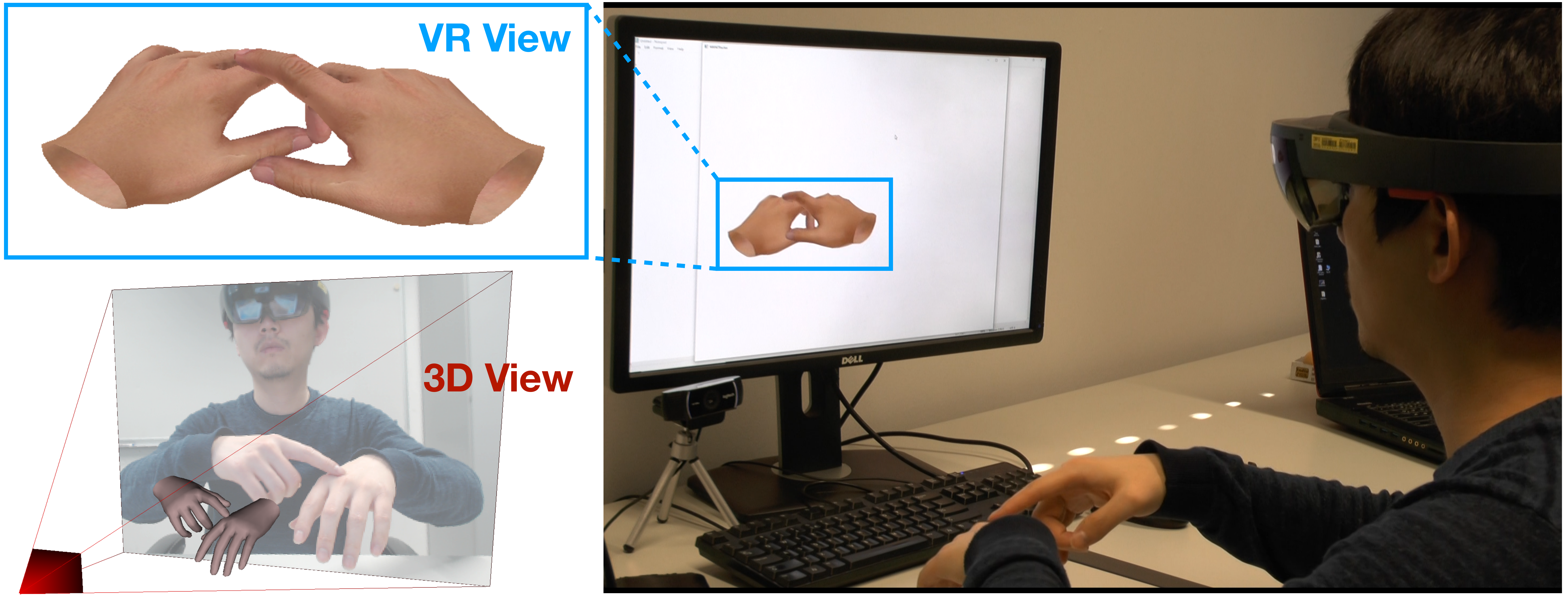}
  \caption{Our RGB2Hands approach tracks and reconstructs the 3D pose and shape of two interacting hands in real time based on a single RGB camera (right). We obtain global 3D pose and shape (bottom left), which can be used to visualize interacting hands in VR (upper left), among many other applications.
  }
  \label{fig:teaser}
\end{teaserfigure}

\maketitle


\section{Introduction}
Marker-less 3D hand motion capture is a challenging and important problem.
With the abundance of smart and mobile devices, interaction paradigms with computers are changing rapidly and moving farther away from the traditional desktop setting.
With the recent progress on virtual and augmented reality (VR/AR), hand pose estimation has gained further attention as direct, natural, and immersive way to interact.
The numerous opportunities for application also include robotics, activity recognition, or sign language recognition and translation.
Hence, hand pose estimation has been an actively researched topic for years.
Depending on the application, several properties are desirable for the method, \textit{e.g.}, marker-less capture, real time performance, capabilities for tracking two interacting hands, automatically adapting to the users' hand shape, or the use of a single RGB camera.
However, due to a range of challenges, such as frequent occlusion, depth-scale ambiguity, and self-similarity of hand parts, achieving all of these properties is a difficult task.

To ease the problem, many previous works on 3D hand pose estimation use special depth cameras providing partial 3D information.
Nevertheless, many of them focused on tracking a single isolated hand \cite{Yuan_2018_CVPR}, with only a few exceptions that are able to handle object interactions \cite{panteleris20153d,tzionas20153d,sridhar_eccv2016} or interactions with a second hand~\cite{taylor_siggraph2016,Taylor_siggraphasia2017,mueller_siggraph2019}.
In recent years, the research focus has shifted towards methods that use a single RGB camera since these sensors are ubiquitous \cite{Zimmermann_2019_ICCV,mueller_cvpr2018,cai2018weakly}.
Despite tremendous progress, to date there is no method explicitly designed for and capable of reconstructing close two-hand interactions from single RGB input.
However, humans naturally use both of their hands for interaction with real and virtual surroundings, and for gesturing and communication. 
Therefore, many applications require hand pose estimation of both hands in close interaction simultaneously.

To this end, we present the first method for marker-less capture of 3D hand motion and shape from monocular RGB input that successfully handles two closely interacting hands.
Our real-time approach automatically adapts to the user's hand shape, and reliably captures collision-resolved poses also under difficult occlusions. 
Since color images carry no explicit 3D information, we also have to cope with scale and depth ambiguities. 
A proper handling of these ambiguities, which are inherent to monocular RGB data, is particularly important in the two-hand case, since mismatches in per-hand depth estimates would lead to incorrectly captured interactions in 3D.
Hence, our setting with a monocular RGB camera is significantly more challenging compared to previous works that make use of depth data, such as~\cite{tzionas_ijcv2016,mueller_siggraph2019}.
To achieve our goal, and thus overcome the challenges and ambiguities of monocular RGB data, we propose a novel multi-task CNN which regresses multiple variables simultaneously: per-pixel left/right hand segmentation masks, dense vertex matchings to a parametric hand model, intra-hand relative depth maps, inter-hand distance, as well as occlusion-robust 2D keypoint positions.
Our regression targets are designed to explicitly consider the challenges of monocular two-hand reconstruction like strong occlusions and ambiguous relative 3D placement of the hands.
We use these predictions in a generative model fitting framework to robustly estimate for both hands the pose and shape parameters of a 3D hand model.

For training our multi-task network we combine real and synthetic data from different sources to bridge the domain gap.
Since none of the publicly available datasets are sufficient for our purposes, in addition we create our own dataset comprising both real and synthetic images.
To obtain real data with (possibly noisy) annotations, we use the depth-based CNN from \citet{mueller_siggraph2019} and an RGB-D sensor.
To obtain perfectly annotated synthetic data, we develop the first system simulating physically correct two-hand interactions with personalized hand shape, based on the parametric MANO hand model~\cite{Romero_siggraphasia2017}, and diverse appearances.
We experimentally show that our proposed mixed-data training set in conjunction with the multi-task CNN is crucial for a successful optimization of the hand model parameters on monocular RGB images.
Our extensive evaluation, in both 2D and 3D, is enabled by our new benchmark dataset \textsc{RGB2Hands} that contains significantly stronger hand interactions compared to previous benchmarks.

In summary, we propose the first monocular-RGB-based method for 3D motion capture of two strongly interacting hands, which simultaneously estimates hand pose and shape, while running in real time.
The technical contributions in order to achieve this include:

\begin{itemize}
    \item A \emph{generative model fitting formulation} that is specifically tailored towards fitting parametric 3D hand models of two interacting hands to an RGB image, while taking inherent depth ambiguities and occlusions into account.
    To this end, we extract information from the input image based on a  machine learning pipeline, which is then used as fitting target.
    \item We propose an \emph{alternative image-based representation of 3D geometry information}, namely intra-hand relative depth, and inter-hand distance, which can be extracted directly from RGB images using our novel \emph{multi-task CNN} and is scalable to dense hand surfaces. 
    In combination with 2D keypoint predictions, and an image-to-hand-model matching prediction, this allows to effectively fit the parametric model.
    \item To train our machine learning predictors, we use synthetic data to complement a real dataset that has possibly noisy annotations. For the former,
  we introduce a \emph{physically-correct synthetic data generation framework}, which is able to account for interacting hands with varying hand identities, both in terms of shape and appearance. 
    \item For performance evaluation we introduce a new benchmark dataset \textsc{RGB2Hands} of real two-hand image sequences that comes with manual keypoint annotations of position and occlusion state.
    Synchronously recorded depth data enables 3D evaluation.
\end{itemize}

\section{Related Work}

Marker-less 3D hand pose estimation has been an actively researched problem for decades, which can be explained by the fact that it enables many important applications, e.\,g.\, in human--computer interaction, activity recognition, or robotics.
In our review of related work we focus on methods using a single depth or RGB camera that are most related to our approach.

\paragraph{Depth-Based Methods}
The majority of previous work on 3D hand pose estimation with a single camera has used depth data \cite{Yuan_2018_CVPR}.
These approaches can generally be classified as generative, discriminative, or hybrid approaches.
Generative methods use a parametric generative hand model and compare current pose hypotheses to the observed image \cite{melax2013dynamics, tagliasacchi_sgp2015,tkach2016sphere}.
The parameters of the hand model are commonly found by minimizing an energy function which describes the discrepancy between the current pose and the observation, commonly in conjunction with suitable priors to serve es regularization.
The employed optimization strategies require a good initialization, so that oftentimes tracking failures occur when there is large inter-frame motion.
With an increasing popularity of machine learning techniques in computer vision, researchers started to investigate hybrid methods.
These approaches add a discriminative component to a generative method to improve the overall robustness, e.\,g.\, through regression of finger-part labels \cite{sridhar_cvpr2015} or partial pose \cite{tang_iccv2015}.
Especially due to the success of neural networks and deep learning, most of the recent work has focused on purely discriminative methods for 3D hand pose estimation.
These approaches are generally based on regressing joint locations from depth data \cite{tompson_tog2014,oberweger_iccv2015,wan2017crossing,Baek_2018_CVPR,Ge_2018_CVPR,Li_2019_CVPR,Chen_2019_ICCV}.
Most of these approaches are single-frame methods and therefore independent of an initialization (in contrast to the mentioned generative methods), however, they are dependent on the training data, which are not trivial to obtain, and independent per-frame estimates may exhibit temporal noise on sequences.
The majority of depth-based hand pose estimation methods are limited to tracking or reconstructing a single hand in free space, and only few approaches have tackled the harder problems of estimating hands and objects, or two interacting hands.
The methods by \cite{mueller_iccv2017,rogez_eccv2014workshop} work for a strongly occluded hand in cluttered scenes with arbitrary objects.
Other methods, like \cite{sridhar_eccv2016,tzionas_ijcv2016}, jointly reconstruct hand and object motion, and are thus able to exploit mutual constraints like physically stable grasps.
Pose estimation methods for two hands often have a trade-off between real-time runtime \cite{Taylor_siggraphasia2017} and accurate collision resolution \cite{tzionas_ijcv2016,oikonomidis2012tracking,kyriazis_cvpr2014}.
The most recent method by \cite{mueller_siggraph2019} runs in real time while providing coarse interpenetration avoidance. All methods discussed in this paragraph have the shortcoming that they rely on specialized camera hardware. In contrast, we address the much more difficult setting of using only more common RGB data, as will be addressed in the next paragraph.

\paragraph{RGB-Based Methods}
Due to the ubiquity of RGB cameras, research on 3D hand pose estimation has shifted towards monocular RGB methods.
Earlier approaches \cite{simon2017hand} estimate 2D hand pose from a single RGB image but require multi-view RGB for 3D pose.
More recent methods are able to estimate normalized 3D pose \cite{Zimmermann:2017um,Spurr_2018_CVPR,cai2018weakly,Yang_2019_ICCV} or even global 3D pose with respect to the camera \cite{mueller_cvpr2018,panteleris2018using,iqbal2018hand}.
The approach by Ge et al.~\shortcite{Ge_2019_CVPR} estimates a full 3D hand mesh directly.
With the increasing popularity of the MANO hand model \cite{Romero_siggraphasia2017} several methods that regress both shape and pose parameters have been proposed \cite{Boukhayma_2019_CVPR,Baek_2019_CVPR,Zhang_2019_ICCV}.
Zimmermann et al.~\shortcite{Zimmermann_2019_ICCV} recently built an extensive dataset of RGB images with fitted MANO models.
However, all the aforementioned monocular-RGB methods only work for a single isolated hand.

Alternatively, a few existing methods track the full 3D body from RGB-only input, including the two articulated hands, by fitting a parametric human model \cite{pavlakos2019SMPLX,xiang2019monocular}. Despite the impressive results, they often fail in close hand-to-hand interaction since they have not been explicitly designed for this setting.
Two-hand tracking has also been attempted by performing per-hand tracking based on tight crops around each hand,~\textit{e.g.},~by \citet{panteleris2018using}. Similarly to the full body methods, this strategy performs poorly in close two-hand interaction.
There are few methods \cite{Tekin_2019_CVPR,hasson19_obman} that jointly reconstruct the pose of a single hand and a manipulated object, but to the best of our knowledge there is no method that reconstructs very close two-hand interactions from monocular RGB images.

In this work we fill this gap and propose the first method to jointly estimate global 3D hand pose and shape of two strongly interacting hands from monocular RGB video.
In addition, our approach runs in real time and resolves collisions for fast and physically accurate results.
\begin{figure*}
  \includegraphics[width=\textwidth]{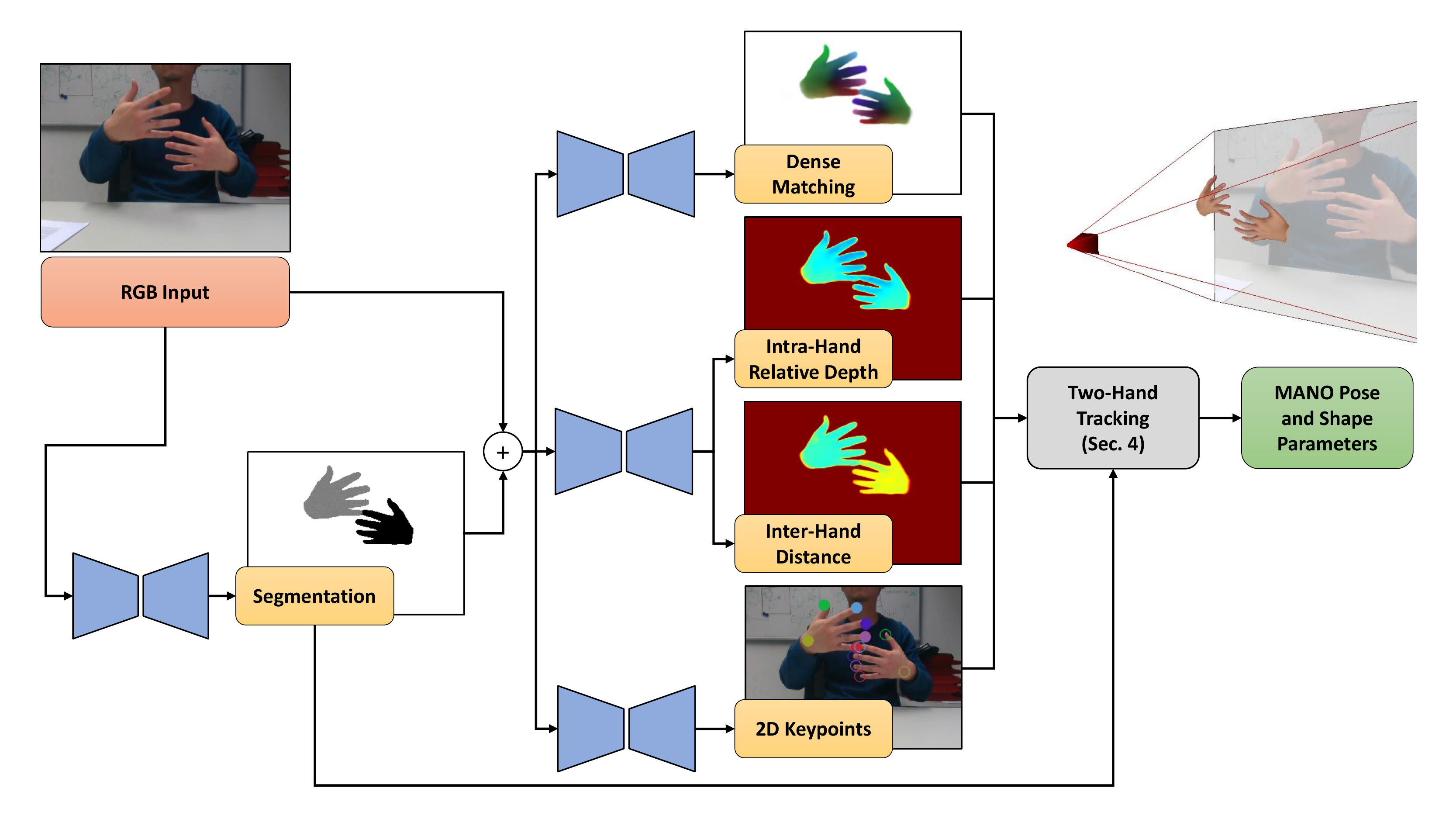}
  \caption{Illustration of our RGB2Hands  approach. The RGB input image is processed by neural predictors that estimate segmentation, dense matching, intra-hand relative depth, inter-hand distances, as well as 2D keypoints. This is then used within our two-hand tracking energy minimization framework. The output are pose and shape parameters of the 3D MANO model~\cite{Romero_siggraphasia2017} of both hands, which directly give rise to a bimanual 3D reconstruction.}
  \label{fig:overview}
\end{figure*}
\section{Overview}
\label{sec:overview}
We present an overview of our approach in Figure \ref{fig:overview}. 
Given a monocular RGB image that depicts a two-hand interaction scenario, our goal is to recover the global 3D pose and 3D surface geometry by fitting a parametric hand model to both hands in the input image, as described in Sec. \ref{sec:pose_estimation}. 
Such a model-fitting task requires information extracted from the input image to be used as a fitting target, which however represents a major challenge when using monocular RGB data only.
Previous methods that rely on depth data \cite{mueller_siggraph2019,Taylor_siggraphasia2017} are implicitly provided with a much richer input (\textit{i.e.}, global depth), which is the fundamental ingredient for an accurate 3D pose and shape fit.
Per-pixel estimation of correct 3D hand depth from a single RGB image is very challenging.

Note that, in particular in the two-hand case, inconsistent depth estimates per hand would lead to incorrectly captured interactions in 3D.
Thus, the method and the scene representation need to be able to handle these ambiguities well.
Therefore, in Sec. \ref{sec:network}, we design an alternative representation of dense 3D geometry information, tailored for a two-hand scenario, which is amenable to be directly extracted from RGB images based on a machine learning pipeline. 
This is in contrast to existing representations which are limited to sparse (\textit{i.e.}, per-hand and/or per-joint) information and cannot be extended to dense geometry in a scalable way, such as joint heatmaps \cite{Zimmermann:2017um,mueller_cvpr2018,panteleris2018using} or part orientation fields \cite{xiang2019monocular}.
To this end, we opt to regress inter-hand distance and intra-hand depth maps, in combination with robust 2D keypoints.
This design choice explicitly provides sufficient information to resolve depth ambiguities in our model-fitting step.
Furthermore, we also regress dense per-pixel surface matchings to the parametric hand model directly from input images.
This step is designed to be robust against the significant skin tone and illumination variability in RGB images. 
Finally,  we describe the training data that we used to train our machine learning components in Sec.~\ref{sec:training_data}, where we also introduce a novel methodology to generate photorealistic and physically accurate synthetic data of sequences with interacting hand motions. 
To this end, we employ a motion capture-driven physics-based simulation to generate physically-correct sequences of hands with varying identities (skin tone and shape). 

\section{Two-Hand Tracking Framework}
\label{sec:pose_estimation}

Our hand representation builds on the parametric surface hand model MANO proposed by \citet{Romero_siggraphasia2017}, which we summarize below. Subsequently, we will derive our model-based fitting framework.

\subsection{Parametric Pose and Shape Model}
MANO was built from more than 1,000 scans of 30 subjects performing a large variety of poses, and consequently the model is capable of reproducing hand shape variability and surface deformations of articulated hands with high detail. Specifically, for a single hand, MANO outputs a set of 3D vertex positions $\mathcal{X}$ of an articulated 3D hand mesh, \ie
\begin{eqnarray}
\mathcal{X}(\bm{\beta},\bm{\theta}) = W(T(\bm{\beta},\bm{\theta}),J(\bm{\beta}),\mathbf{W})\,,
\end{eqnarray}
where $\bm{\beta} \in \mathbb{R}^{10}$ and $\bm{\theta} \in \mathbb{R}^{51}$ are the shape and pose parameters with the latter consisting of 45 articulation parameters and 6 global rotation and translation parameters.
$T(\cdot)$ is a parametric hand template in rest pose with pose-dependent corrections to reduce skinning artifacts, $J(\cdot)$ computes the 3D position of the hand joints, and $\mathbf{W}$ is a matrix of rigging weights used by the skinning function $W$ (based on linear blend skinning). See \cite{Romero_siggraphasia2017} for further details.
 
 As we are targeting a two-hand scenario, we use two sets of shape and pose parameters $(\bm{\beta}_h,\bm{\theta}_{h}), h \in \{\text{left},\text{right}\}$, for the left and right hand respectively.
 To simplify the notation, we stack the   parameters of both hands as $\bm{\beta} = ({\bm{\beta}_{\text{left}}},{\bm{\beta}_{\text{right}}}) \in \mathbb{R}^{20}$ and $\bm{\theta} = ({\bm{\theta}_{\text{left}}},{\bm{\theta}_{\text{right}}}) \in \mathbb{R}^{102}$, and define the unique set of vertices $\mathcal{X} =(\mathcal{X}_{\text{left}},\mathcal{X}_{\text{right}})$, where we may omit the dependence of $\mathcal{X}$ on $\bm{\beta}$ and $\bm{\theta}$ for brevity.

\subsection{Overview of Model-Based Fitting Formulation}
In order to track two interacting hands in an image sequence we use the parametric MANO model within an energy minimization framework. To this end we introduce the fitting energy $f(\bm{\beta},\bm{\theta})$ as
\begin{align}\label{eq:fit}
    f(\bm{\beta},\bm{\theta}) &= \Phi(\bm{\beta},\bm{\theta}) + \Omega(\bm{\beta},\bm{\theta})
    \,,
\end{align}
where $\Phi(\cdot)$  is the image fitting term that accounts for fitting the model to the observed RGB image, and $\Omega(\cdot)$ is the regularizer that has the purpose of obtaining a plausible and well-behaved tracking result. 
By minimizing the fitting energy $f$ we jointly estimate the pose and shape parameters $\bm{\theta} \in \mathbb{R}^{102}, \bm{\beta} \in \mathbb{R}^{20}$ (of both hands) for each frame of the image sequence.

\subsection{Image-fitting Term}
Due to the 2D nature of RGB images and the so-resulting depth ambiguities, as well as the additional level of difficulty caused by interactions between the left and right hand, our novel image-fitting term $\Phi$ is designed carefully in order to allow for a reliable fit of the parametric hand model. 
In particular it uses specific information that our multi-task CNN (see Sec.~\ref{sec:network}) extracts from 2D images that enables us to estimate the correct and coherent 3D pose of both hands in interaction, and minimizes the risk of implausible interaction capture due to ambiguous 3D pose estimates of each individual hand.
We propose to combine five components, where we use  
\begin{enumerate}
    \item the dense 2D fitting term $\Phi_{\text{dense}}$,
    \item the silhouette term $\Phi_{\text{sil}}$, 
    \item the 2D keypoint term $\Phi_{\text{key}}$, 
    \item the intra-hand relative depth term $\Phi_{\text{intra}}$, and
    \item the inter-hand distance term $\Phi_{\text{inter}}$.
\end{enumerate}
We emphasize that existing methods that are capable of tracking \textit{two hands in interaction} avoid 3D pose ambiguities by heavily relying on depth-based input data that is used in their image-fitting term, 
which, however severely simplifies the problem. 
In contrast, our energy terms $\Phi_{\text{dense}}$, $\Phi_{\text{intra}}$, $\Phi_{\text{inter}}$ have the purpose of compensating the lack of available depth information and enable 3D consistent two-hand reconstructions by using a strong neural prior that extracts suitable information from RGB images only.

With that, the complete image fitting term that accounts for the model-to-image fitting reads
\begin{equation}
    \Phi(\bm{\beta},\bm{\theta}) = 
    \Phi_{\text{dense}} + 
    \Phi_{\text{sil}} + 
     \Phi_{\text{key}} + 
     \Phi_{\text{intra}} + 
   \Phi_{\text{inter}} \,,
\end{equation}
where we have omitted the explicit dependence on $(\bm{\beta},\bm{\theta})$ of the individual terms for the sake of readability.

We assume known camera intrinsics and define $\Pi: \mathbb{R}^3 \rightarrow \Gamma$ to be the projection from camera space onto the image plane. When this is not available, plausible intrinsics can be provided to obtain results accurate up to a scale.

One crucial part for defining the image fitting term is the \emph{dense matching map}  $\psi: \mathcal{X} \rightarrow \Gamma$, which predicts for each vertex $\bm{x} \in \mathcal{X}$ the corresponding pixel position $(u,v) \in \Gamma$ in the input image. For the time being we will assume that $\psi$ is known, and later in Sec.~\ref{sec:network} we will explain how we obtain it.
In the following, when we sum over vertices in the set $\mathcal{X}$, we only consider those vertices that are visible, where a vertex $\bm{x}$ is considered to be visible whenever $\psi(\bm{x}) \neq \emptyset$. 

We will now explain the individual components in depth.

\paragraph{Dense 2D Fitting:}
Since an RGB image does not contain explicit 3D information, the actual depth of a model vertex is unknown. 
Hence, we penalize the 2D image-plane distance between a projected visible vertex $\Pi(\bm{x})$ and its corresponding pixel $\psi(\bm{x})$. We define the dense 2D fitting term as
\begin{equation}
    \Phi_{\text{dense}}(\bm{\beta},\bm{\theta}) =
    \lambda_{\text{d}}  \sum_{\bm{x} \in \mathcal{X}}
    \| \Pi(\bm{x}) -  \psi(\bm{x}) \|_2^2\,,
\end{equation}
where $\lambda_{\text{d}}$ is the relative weight of this term.
\paragraph{Silhouettes:}

\begin{figure}
    \centering
    \includegraphics[width=0.49\linewidth,height=3cm]{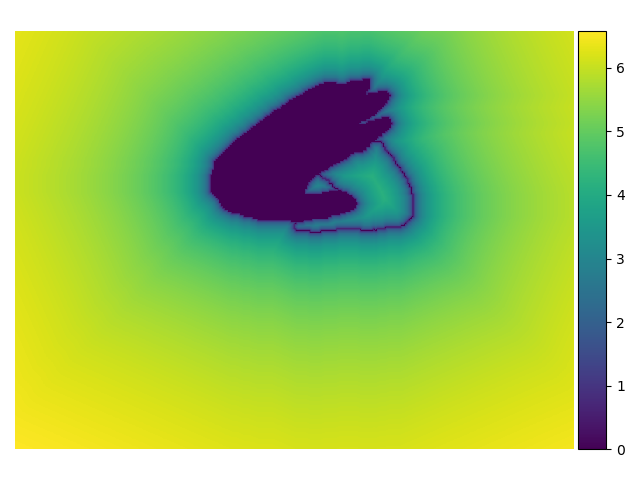}
    \includegraphics[width=0.49\linewidth,height=3cm]{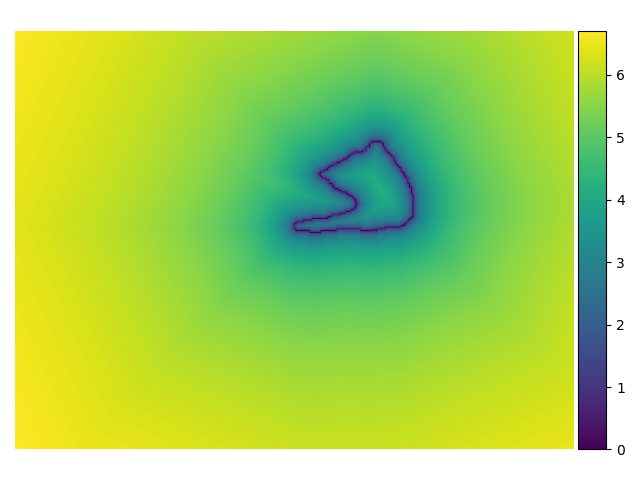}
    \includegraphics[width=0.49\linewidth,height=3cm]{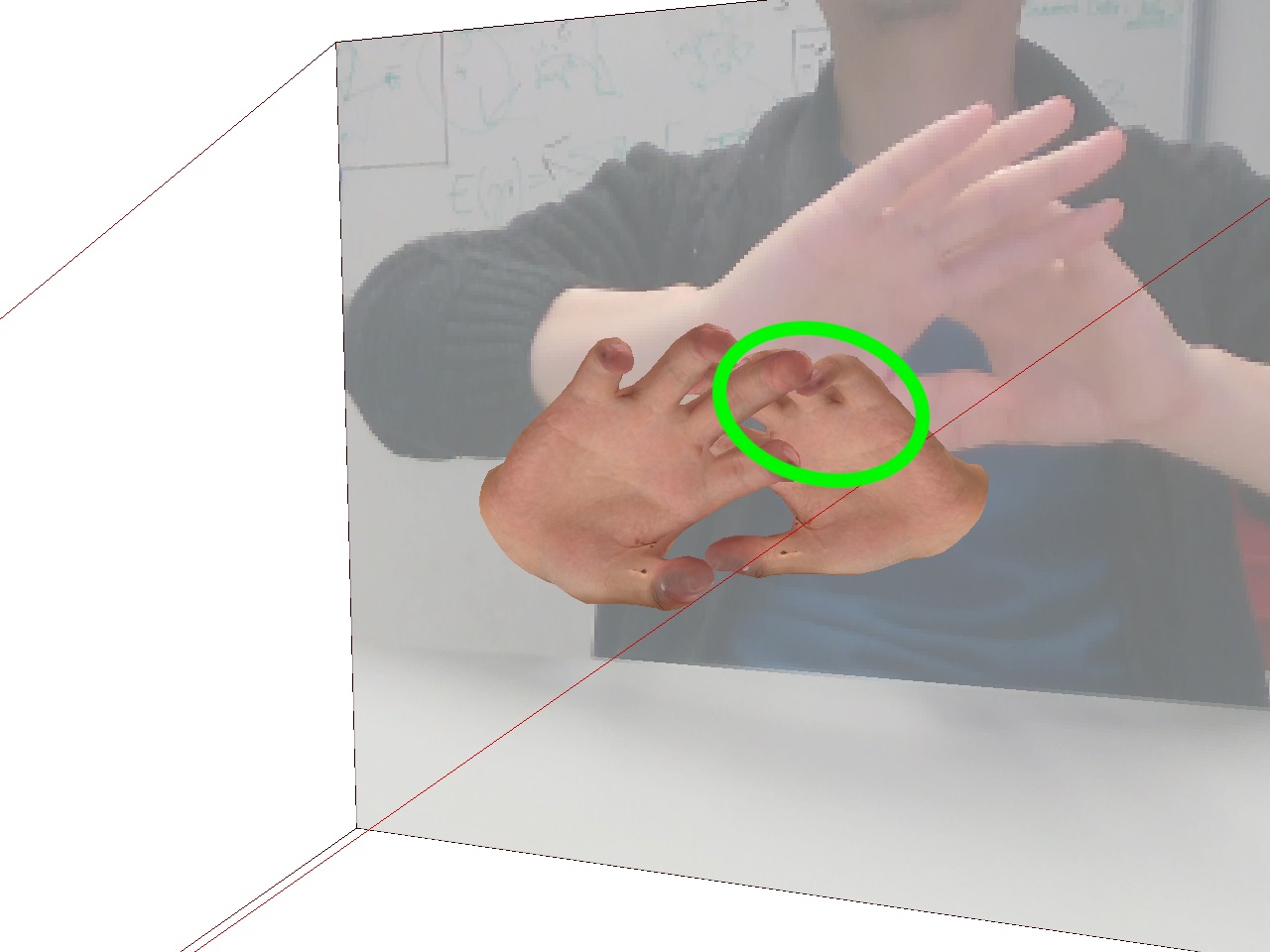}
    \includegraphics[width=0.49\linewidth,height=3cm]{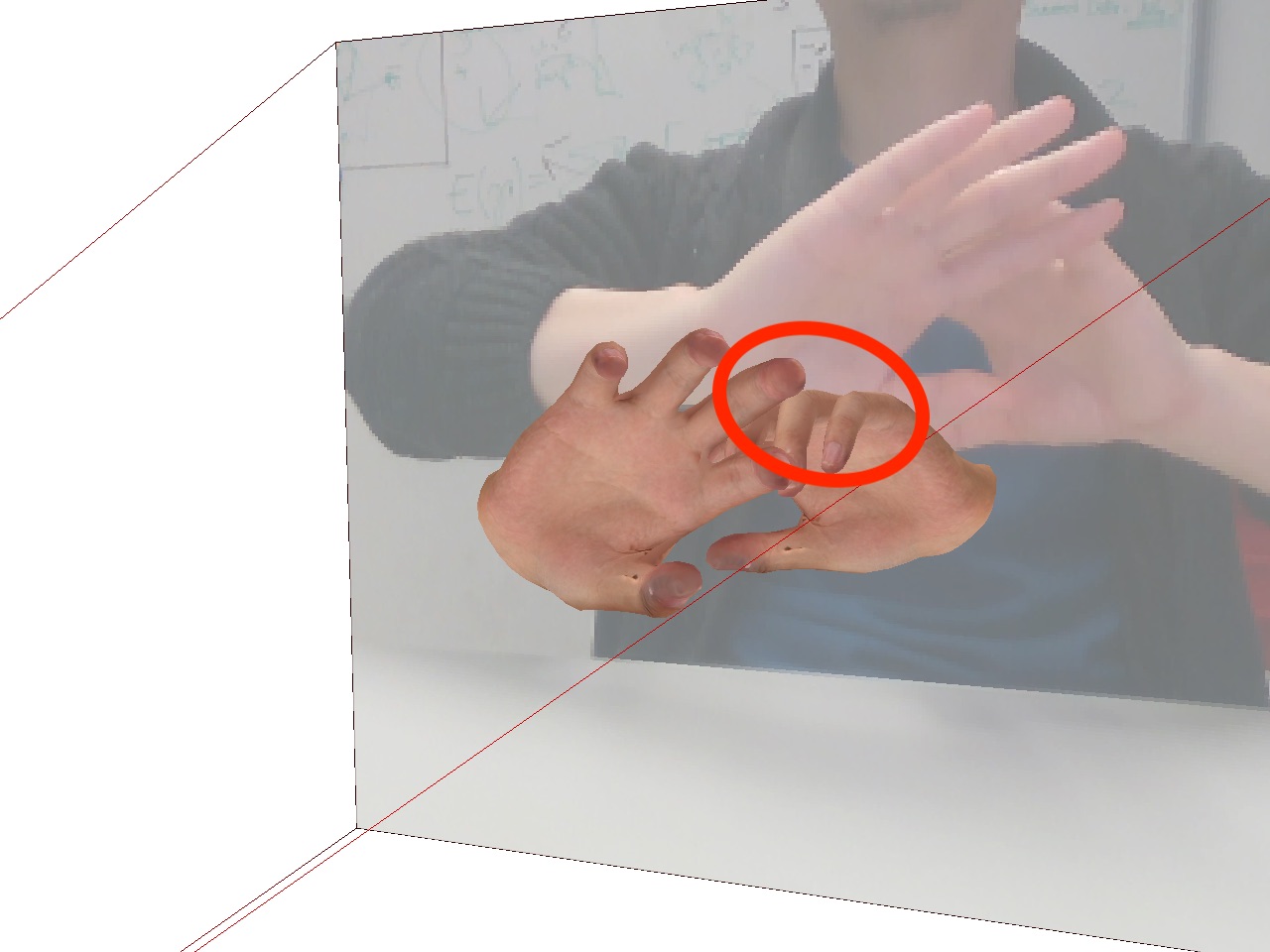}

    \caption{
    Visualization of $log(DT_{\text{left}}+1)$ with (top left) and without (top right) occlusion handling. The reconstructed hand without occlusion handling (bottom right) incorrectly articulates to explain an occlusion boundary, while our proposed method (bottom left) correctly handles the occlusion. 
    }
    
    \label{fig:distance_transform}
\end{figure}

Since the dense matching map might not be perfectly precise for neighboring vertices and pixels, we introduce an occlusion-aware silhouette term to improve the projection error of the estimated hand models in the input image.
Similar to previous work \cite{Habermann2019livecap}, we define a set of \emph{boundary vertices} $\mathcal{X}_b$ and penalize their distance from the silhouette edges in the input image.
We determine the set of boundary vertices based on the current pose and shape estimate in every iteration of the optimization. 
We choose all hand model vertices that lie close to model-to-background edges in the projected view.
To efficiently represent the distance to the silhouette edges without explicit correspondences, a Euclidean distance transform representation is used.
Since we need to distinguish the right and left hand, we create two distance transform images $DT_{\text{right}}$ and $DT_{\text{left}}$, one for each hand respectively.
To this end, we make use of the predicted segmentation mask $\mathcal{S}$ (see Section \ref{sec:network_outputs}) to extract silhouette edges per hand.
Since we specifically target close two-hand interactions, the segmentation mask does not only contain silhouette edges but also occlusion boundaries (\textit{i.e.}, hand-hand boundaries).
Without proper handling, vertices that are occluded by the other hand would be drawn towards the occlusion boundary, which in turn would encourage shrinking of the occluded hand.
Thus, we set the distance transform image for each hand to 0 at all pixels that are predicted to belong to the other hand (see Fig.~\ref{fig:distance_transform}).
With that, boundary vertices that project onto the other hand in the input image are not pulled towards the occlusion boundary, which would produce an undesirable distortion effect, leading to a grasping pose, everytime a hand is occluded. 
Mathematically, our occlusion-aware silhouette term is formulated as
\begin{equation}
    \Phi_{\text{sil}}(\bm{\beta},\bm{\theta}) =
    \lambda_{\text{sil}}  \sum_{\bm{x}_b \in \mathcal{X}_b}
    \left( DT_{h(\bm{x}_b)}(\Pi(\bm{x}_b)) \right)^2 \,,
\end{equation}
where $h(\bm{x}_b)$ gives the handedness of boundary vertex $\bm{x}_b$.
Note that we use an additional normal-based weight for each summand as introduced by \citet{Habermann2019livecap}.
Please refer to this paper for further details.

\paragraph{2D keypoints:} 
Since the dense 2D fitting only constrains visible parts of the hand model, we add an occlusion-robust 2D keypoint term.
We penalize the discrepancy between corresponding keypoint predictions on the RGB image and the hand model projected to the image plane. 
Our keypoint detection is designed to also be available under occlusion, increasing the robustness to strong occlusions that frequently occur in the two-hand scenario.
For each hand we use the center of the wrist and the 5 fingertip positions as keypoints, leading to a total number of 12 keypoints across both hands. 
We use $\bm{x}_j \in \mathbb{R}^3$ to denote the 3D position of the $j$-th keypoint of the hand model. Similarly, by $\mathcal{Q}_{\text{key}}(j) \in \Gamma$ we denote the pixel position of the $j$-th keypoint in the image, which is obtained based on the keypoint predictor $\mathcal{Q}_{\text{key}}$ that we will define in Sec.~\ref{sec:network}. Let $\mathcal{J}$ be the set of \emph{detected} keypoints, which may have less than 12 elements whenever some keypoints do not meet the confidence threshold (see Section \ref{sec:network_outputs}).
With that, our 2D keypoint term reads
\begin{align}
    \Phi_{\text{key}}(\bm{\beta},\bm{\theta}) = 
    \lambda_{\text{key}} \sum_{ j \in \mathcal{J}} \| \Pi(\bm{x}_j) - \mathcal{Q}_{\text{key}}(j) \|_2^2\,.
\end{align}

\paragraph{Intra-hand relative depth:}
In order to address depth ambiguities within estimated 3D pose and shape of each individual hand (cf.~the Bas-Relief Ambiguity~\cite{belhumeur1999bas}), we introduce the \emph{intra-hand relative depth term} that penalizes the differences between per-hand root-relative depth values of the 3D hand model and per-hand relative depth predictions obtained from the RGB image.
To this end, we compare the estimated distance along the camera direction (which we refer to as \emph{z-direction})
from the hand root joint in the model to an analogous output of a machine learning predictor (Sec.~\ref{sec:network}) that serves as relative depth prior conditioned on the RGB image.
Let the function $\operatorname{root}(\bm{x})$ compute the 3D position of the root joint
of the hand to which the vertex $\bm{x}$ belongs to, and let $(\cdot)_z$ denote the extraction of the z-component of a 3D vector. Moreover, by $\mathcal{Q}_{\text{intra}}(u,v)$ we denote the  relative depth that was predicted by a neural network in the image at the pixel $(u,v)$. With that, we define the intra-hand relative depth term $\Phi_{\text{intra}}$ as
\begin{equation}
\label{eq:intra}
    \Phi_{\text{intra}}(\bm{\beta},\bm{\theta}) =
    \lambda_{\text{intra}} \sum_{x \in \mathcal{X}}
    ( \mathcal{Q}_{\text{intra}}(\psi(\bm{x})) - (\bm{x}_z - \operatorname{root}(\bm{x})_z) )^2
    \,.
\end{equation}

\paragraph{Inter-hand distance:}
In addition to the intra-hand relative depth, we also take the \emph{inter-hand distance} into account, where we compare the estimated distance between the root of both hands to the output of a trained learning system predicting the same conditioned on the RGB image. 
Note that this term is crucial to obtain correct relative placement of the two hands in 3D from monocular RGB data.
Let ${\operatorname{root}_h}, h \in \{\text{left, right}\}$ be the 3D position of the root joint of a hand 
and let $q_{\text{inter}}$ denote the relative distance of the left hand from the right hand as predicted by a neural network. 
With that, we define the inter-hand distance term as
\begin{align}
\label{eq:inter}
     \Phi_{\text{inter}}(\bm{\beta},\bm{\theta}) = 
     \lambda_{\text{inter}} \left(
     (\operatorname{root}_{\text{left}})_z - (\operatorname{root}_{\text{right}})_z  - q_{\text{inter}} \right)^2 \,.
\end{align}

\subsection{Hand Model and Tracking Regularization}
In order to enable a plausible and realistic tracking, we define a regularizer $\Omega(\bm{\beta},\bm{\theta})$ that combines different terms to account for an appropriate regularization of the parametric hand model:
\begin{align}
\Omega(\bm{\beta},\bm{\theta}) = \Omega_0(\bm{\beta},\bm{\theta}) + \Omega_{\text{overlap}}(\bm{\beta},\bm{\theta}) + \Omega_{\text{scale}}(\bm{\beta}) \,.
\end{align}
Below we first summarize the \emph{structural regularizers} $\Omega_0$ and $\Omega_{\text{overlap}}$, which are a well-established terms in hand tracking and reconstruction settings.
We refer the interested reader to previous works for a more detailed description, such as~\cite{tagliasacchi_sgp2015,tan_cvpr2016,Romero_siggraphasia2017,mueller_siggraph2019}.
Subsequently, we introduce the new (optional) \emph{hand scale prior} $\Omega_{\text{scale}}$, which we designed in order to address the scale ambiguity that arises specifically when performing 3D reconstruction in monocular RGB data. If this prior is provided, our method is able to obtain \emph{metric} 3D pose and shape reconstruction results.

\paragraph{Structural Regularization:} We impose Tikhonov regularization upon the shape parameter $\bm{\beta}$, which accounts for it following a multivariate standard normal distribution. Similarly, we use a thresholded version thereof for the pose paraemter  $\bm{\theta}$, so that poses close to the mean pose are not penalized. 
Furthermore, we impose a temporal regularization that penalizes the difference between the parameters at the current and the previous frame. Moreover, in order to ensure that the shapes of the left and right hand are similar, we penalize discrepancies between the hand shapes. We write these structural regularizers in terms of the squared $\ell_2$-norm summarily as 
\begin{align}
    \Omega_{0}(\bm{\beta},\bm{\theta}) = \big\| 
    \begin{bmatrix}
    {\lambda_{\beta}} \bm{\beta} \\
    {\lambda_{\theta}} \,\bm{1}_{>t_{\theta}} (\bm{\theta}) \\
    {\lambda_{\tau}} (\bm{\beta'} - \bm{\beta}) \\
    {\lambda_{\tau}} (\bm{\theta'} - \bm{\theta}) \\
    \lambda_{\text{sym}} (\bm{\beta}_{\text{left}} - \bm{\beta}_{\text{right}})
    \end{bmatrix}
    \big\|_2^2  \,,
\end{align}
where $\bm{1}_{>t_{\theta}}(\bm{\theta})$ is a  function yielding $\bm{\theta}$ if $||\bm{\theta}||_2 > t_{\theta}$, and $\bm{0}$ otherwise. The variables
$\bm{\beta'}$ and $\bm{\theta'}$ denote the shape and pose parameters from the previous frame, and $\lambda_{\bullet}$ are the respective weights.

For avoiding collisions between the two hands, as well as within each hand, we penalize mesh overlaps as approximated with 3D Gaussians that are attached to the parametric hand model.
The position and size of the Gaussians change according to the shape and pose parameters $(\bm{\beta},\bm{\theta})$~\cite{mueller_siggraph2019}. For  $\mathcal{N }_i(\bm{z} | \bm{\beta},\bm{\theta})$ denoting the $i$-th 3D Gaussian evaluated at the position $\bm{z} \in \mathbb{R}^3$, we compute the overlap between all pairs $(i,j)$ of Gaussians as
\begin{equation}
    \Omega_{\text{overlap}}(\bm{\beta},\bm{\theta}) = \lambda_{\mathcal{N}} \sum_{i,j} \big(\int_{\mathbb{R}^3} \mathcal{N }_i(\bm{z}| \bm{\beta},\bm{\theta}) \cdot \mathcal{N}_j(\bm{z}  | \bm{\beta},\bm{\theta}) \, d\bm{z} \big)^2 \,.
\end{equation}

\paragraph{Hand Scale Prior:}
Since reconstruction from monocular RGB data is inherently ambiguous up to a single scalar factor, we give the option to provide a single metric measurement of the user's hand in order to produce metric results.
We choose this measurement to be the length of the palm, defined as the distance between the middle finger metacarpophalangeal joint (MCP) and the wrist.
If the user does not provide this measurement, we assume the  palm length is given by the mean shape of the MANO model, i.e., for $\bm{\beta} =\bm{0}$.
We formulate the hand scale prior  to penalize deviations from the pre-defined palm length $\alpha$ as
\begin{equation}
    \Omega_{\text{scale}}(\bm{\beta}) = \lambda_{s} \sum_{h \in \{\text{left},\text{right}\}}(\operatorname{palmlength}\,(\bm{\beta}_h) - \alpha)^2
    \quad ,
\end{equation}
where the function $\operatorname{palmlength}(\cdot)$ computes the length of the palm of the hand model given a set of shape parameters.

\subsection{Numerical Optimization}
For the numerical optimization of the fitting energy $f$ in~\eqref{eq:fit} we use a Levenberg-Marquardt (LM) approach. The main idea here is to iteratively update the parameters $\bm{\nu}  := (\bm{\beta}, \bm{\theta})$ using the Jacobian matrix $\bm{J}_f$ of $f$ as
\begin{align}
    \bm{\nu} =  \bm{\nu}^{\text{old}} - (\bm{J}_f^T \bm{J}_f + \mu \mathbf{I})^{-1} \bm{J}_f^T \bm{f}(\bm{\nu}^{\text{old}})\,,
\end{align}
where $\bm{f}$ is the vector-valued function that stacks all the individual (quadratic) residuals of $f$, and $\mu$ is the LM damping factor.
Based on empirical evidence, the LM method is generally known for rapidly decreasing the objective function with very few iterations. Hence, and in order to maintain real-time performance, in addition to efficiently evaluating the Jacobian on the GPU, we terminate the iterative optimization after $10$ iterations.
\section{Dense Matching and Depth Regression}
\label{sec:network}

In order to obtain the predictions that were described in the previous section, including predictions for segmentation, dense matching, intra-hand depth, inter-hand distance and 2D keypoints, we feed the RGB input image to a fully-convolutional neural network.
This enables us to work on entire images without requiring a potentially error-prone bounding box estimation for each hand.
Since our network is trained using a large training corpus, it successfully learns priors to handle the inherent ambiguities in monocular RGB data.
In the following, we describe our network, including the outputs, losses, and the architecture, in more detail.

\subsection{Network Outputs}
\label{sec:network_outputs}
\begin{figure*}
	\centering
	\includegraphics[width=0.195\textwidth]{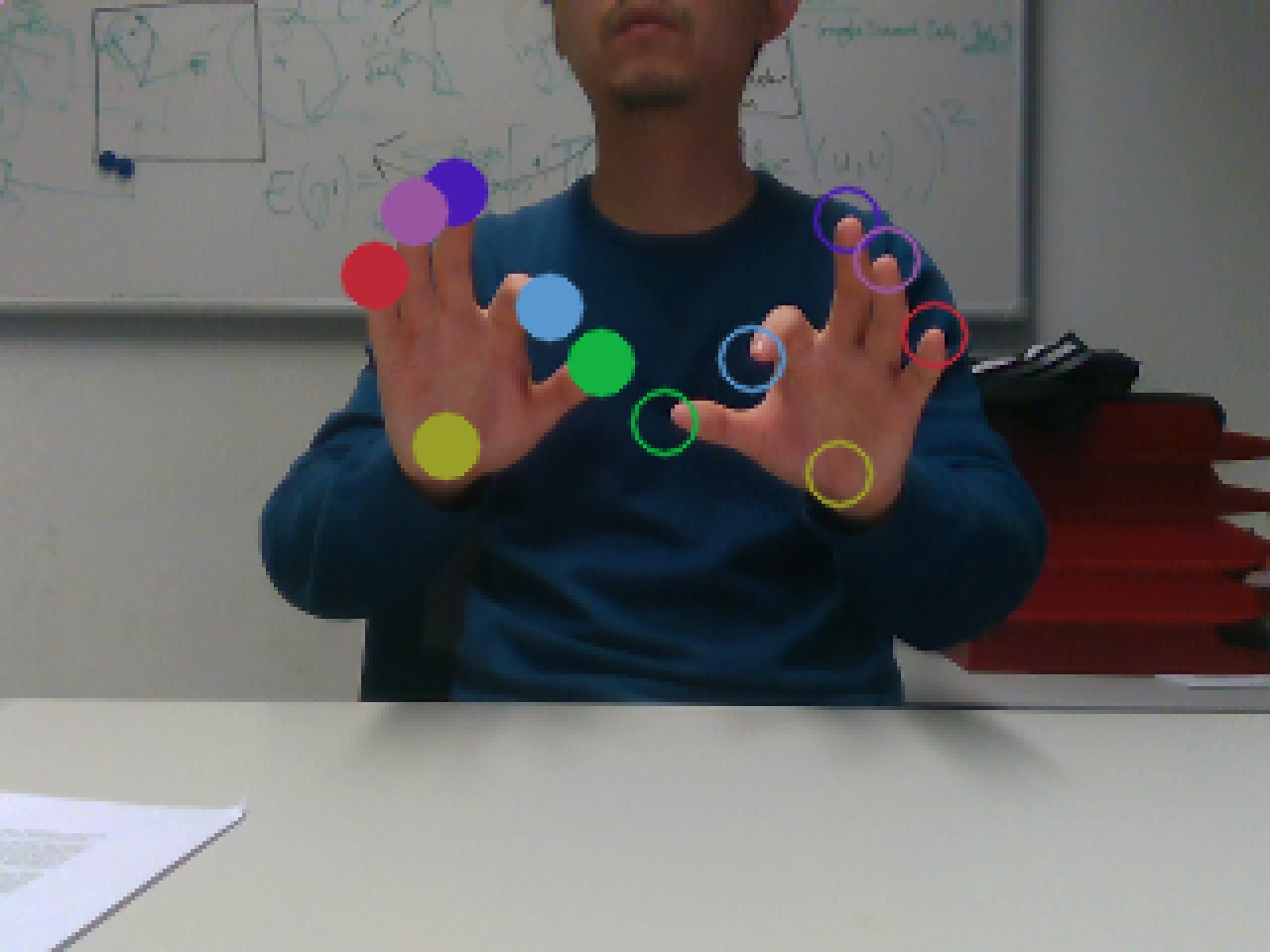}
	\includegraphics[width=0.195\textwidth]{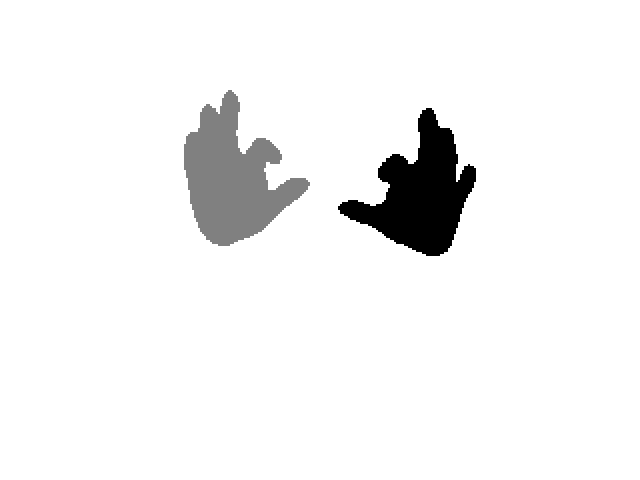}
	\includegraphics[width=0.195\textwidth]{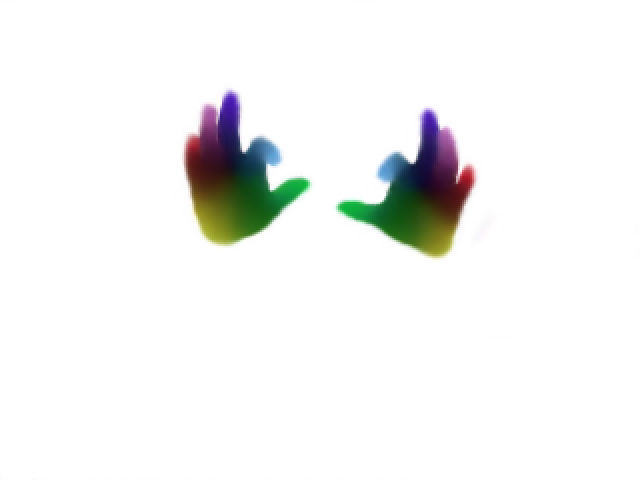}
	\includegraphics[width=0.195\textwidth]{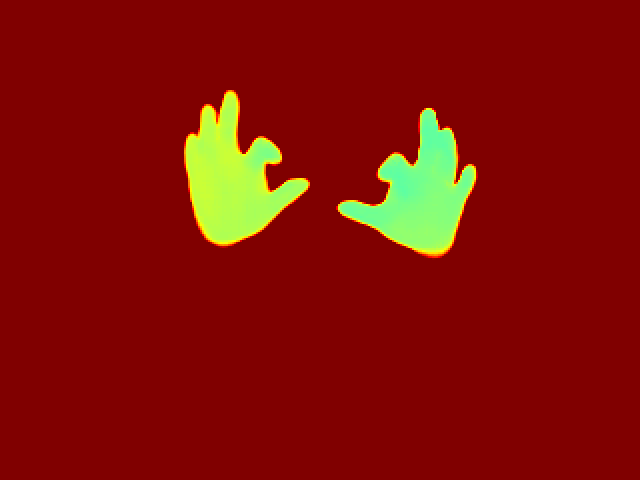}
	\includegraphics[width=0.195\textwidth]{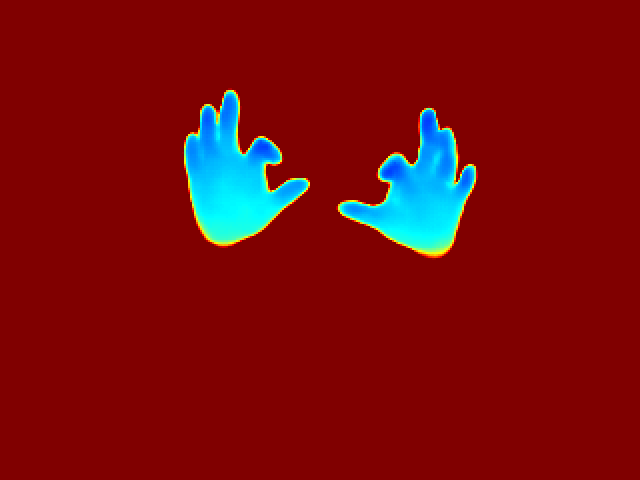}
	\\[0.03cm]
	\includegraphics[width=0.195\textwidth]{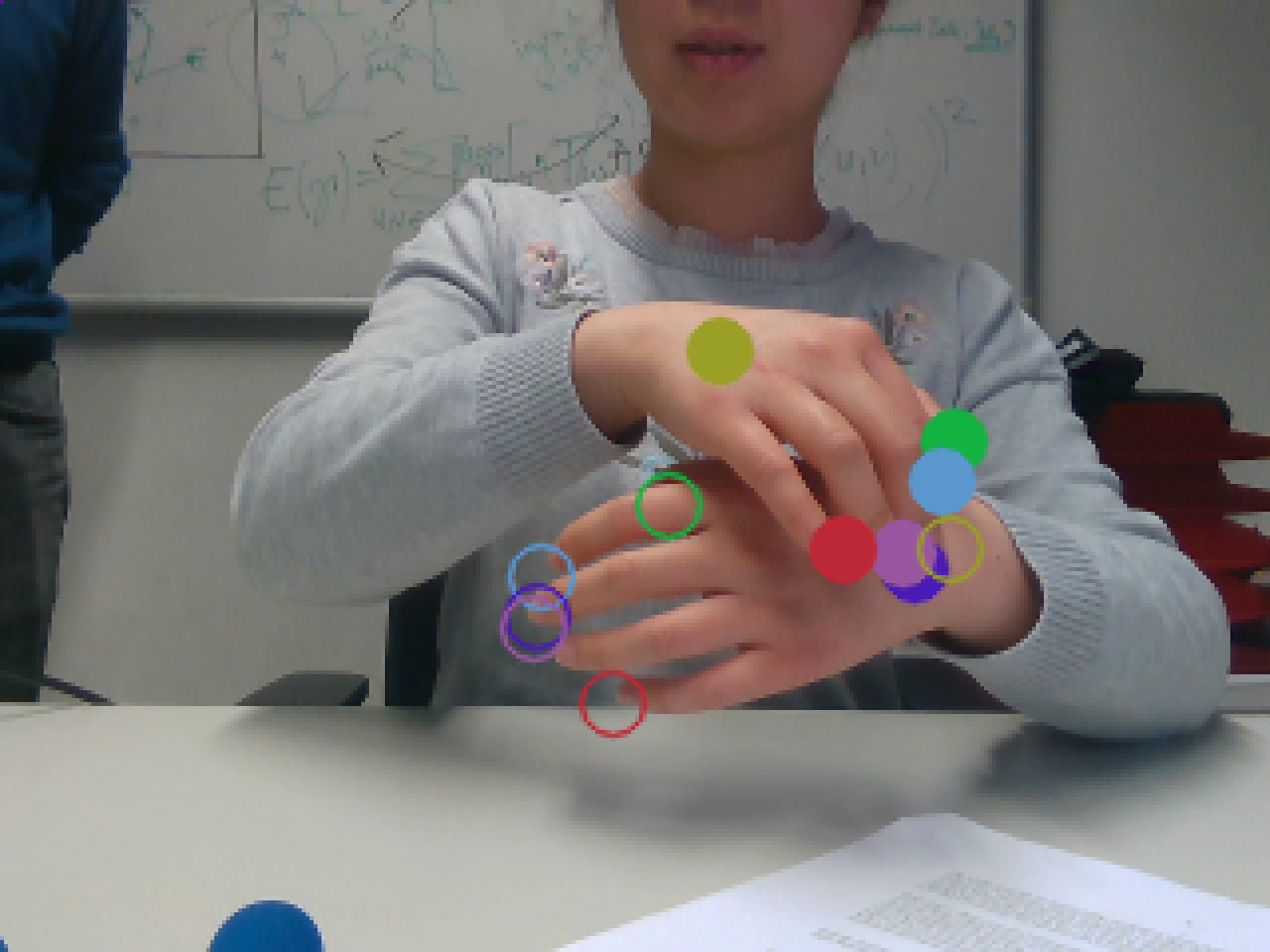}
	\includegraphics[width=0.195\textwidth]{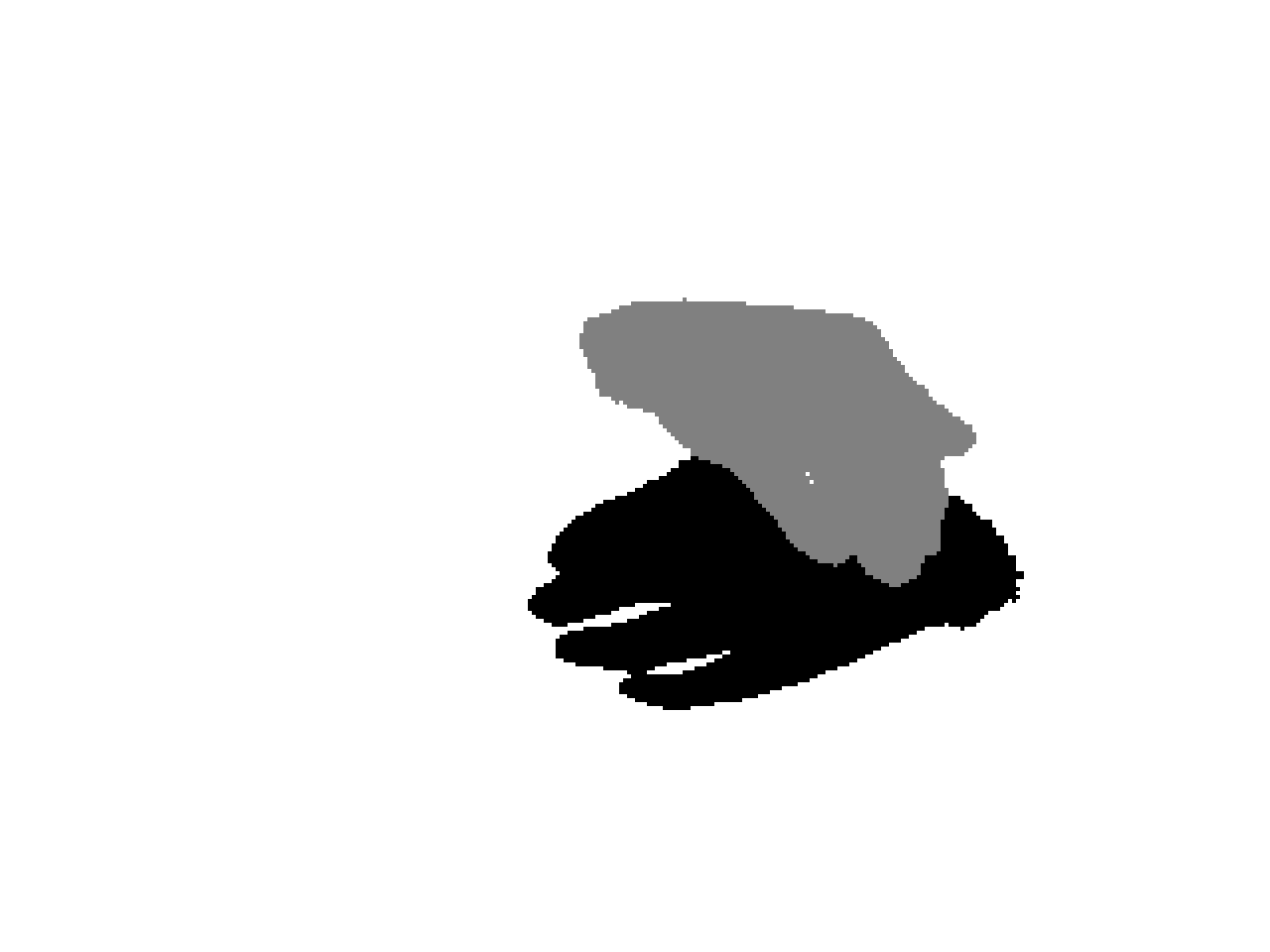}
	\includegraphics[width=0.195\textwidth]{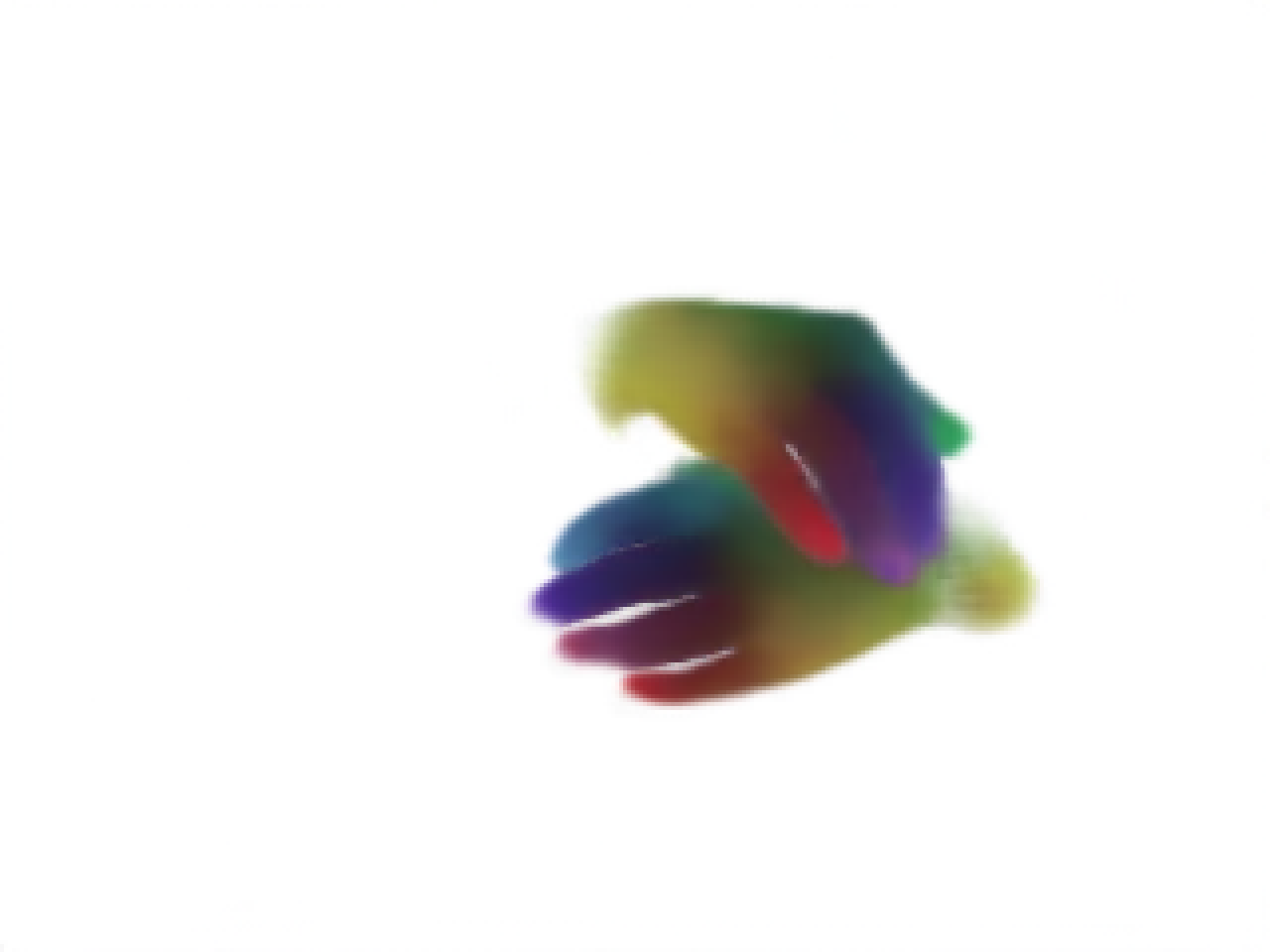}
	\includegraphics[width=0.195\textwidth]{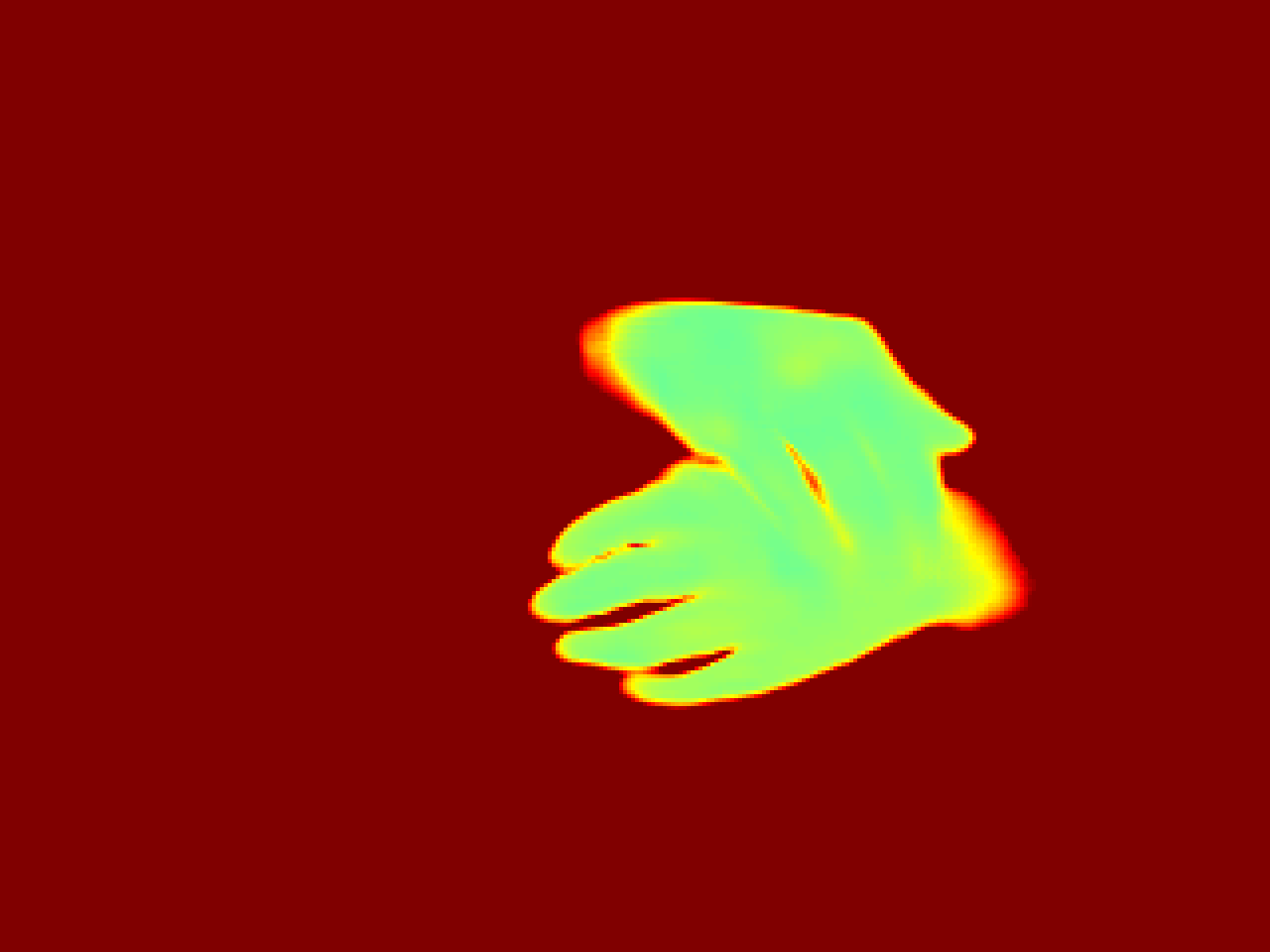}
	\includegraphics[width=0.195\textwidth]{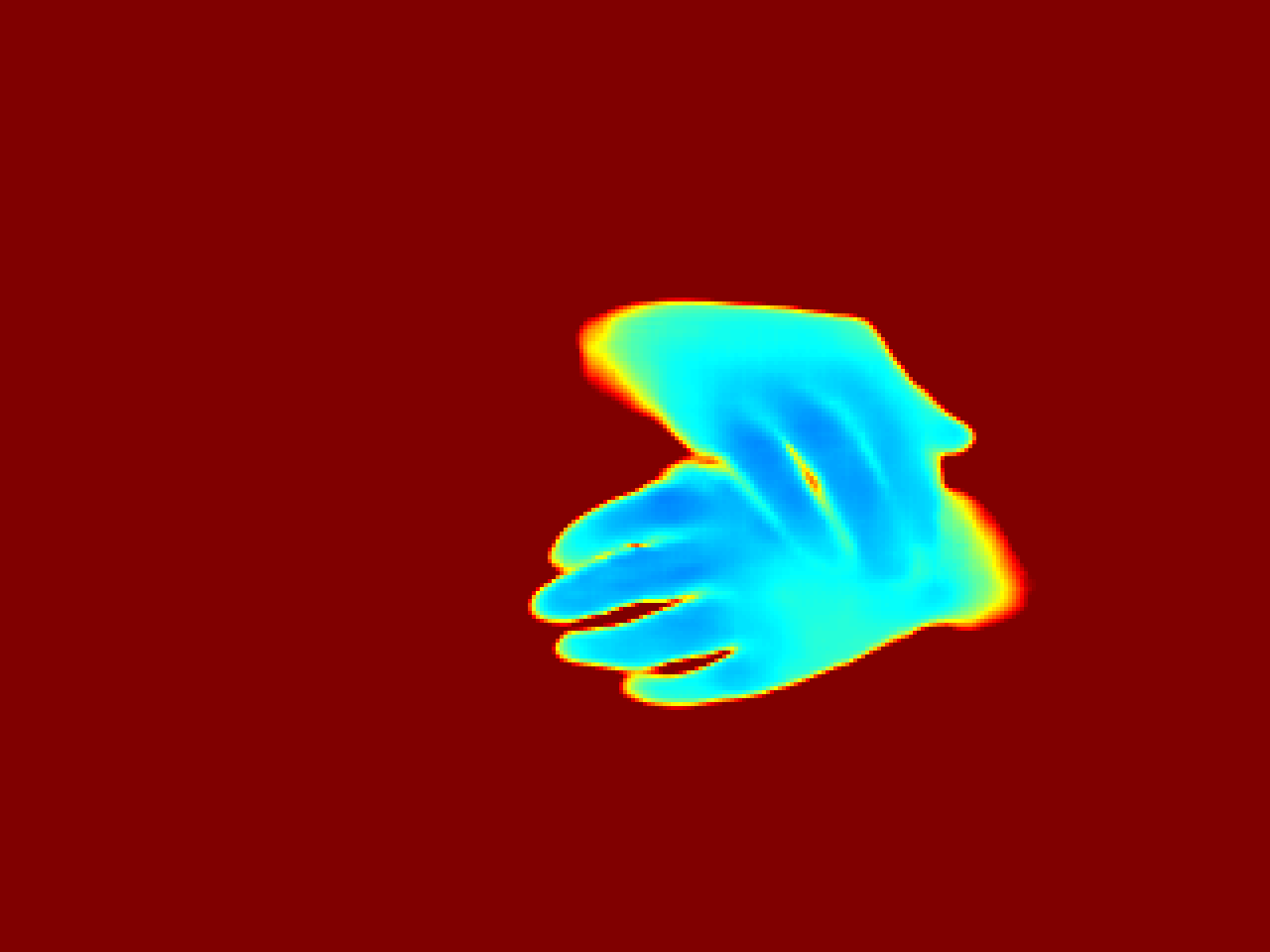}
	\\[0.03cm]	
	\includegraphics[width=0.195\textwidth]{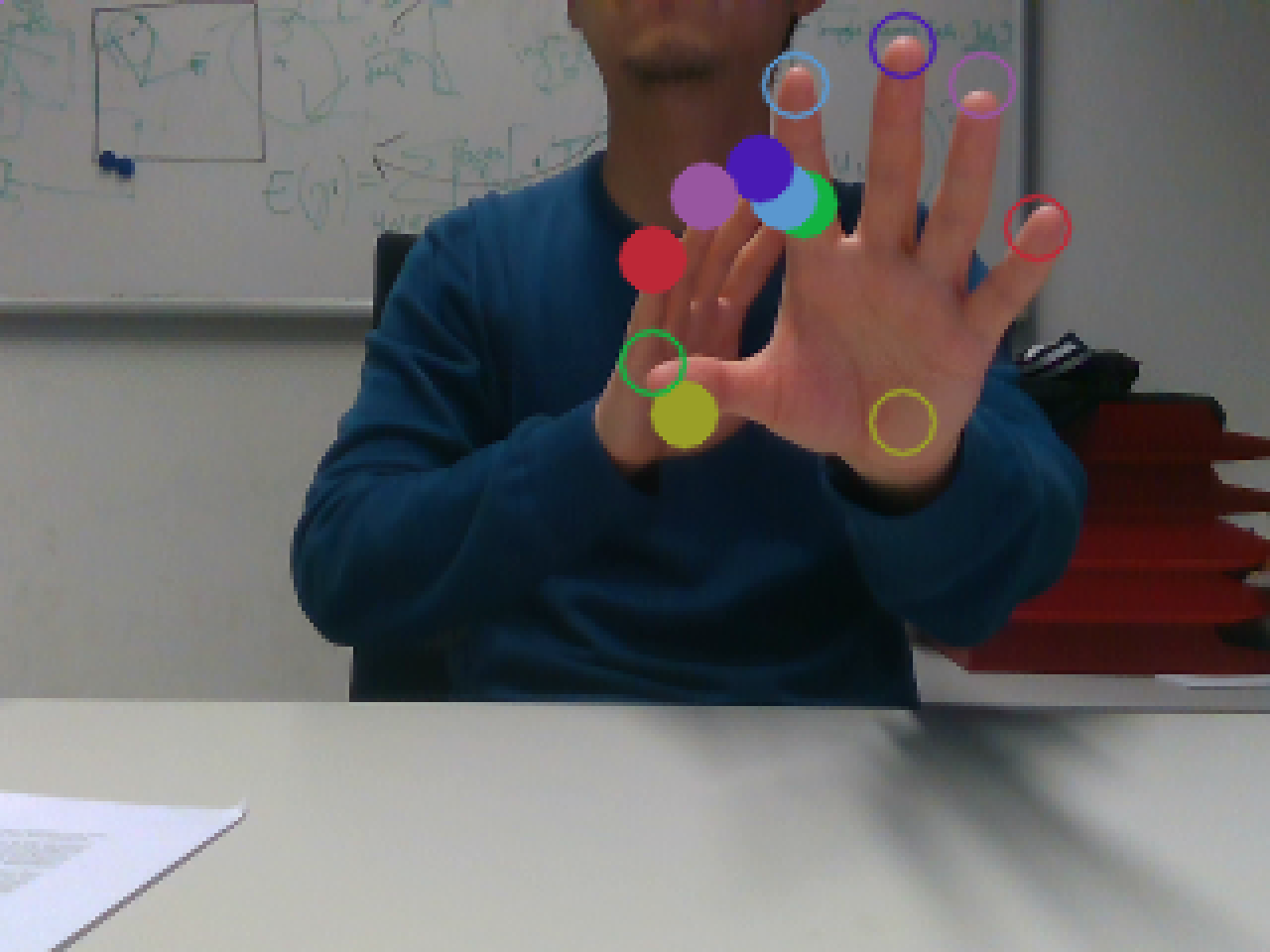}
	\includegraphics[width=0.195\textwidth]{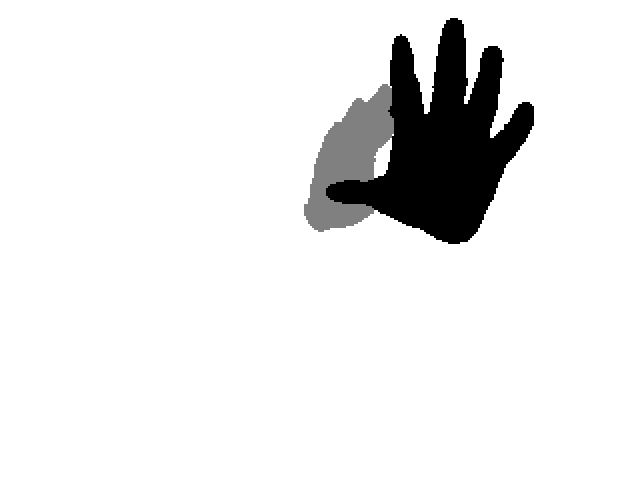}
	\includegraphics[width=0.195\textwidth]{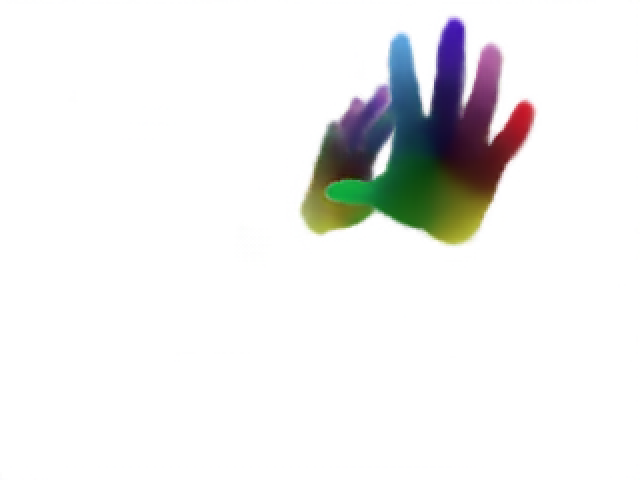}
	\includegraphics[width=0.195\textwidth]{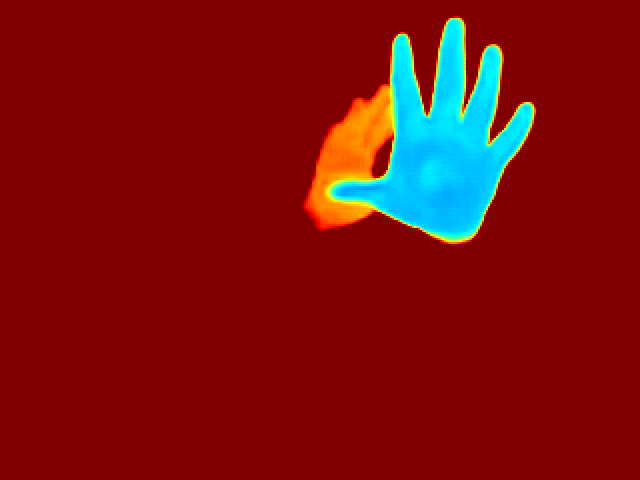}
	\includegraphics[width=0.195\textwidth]{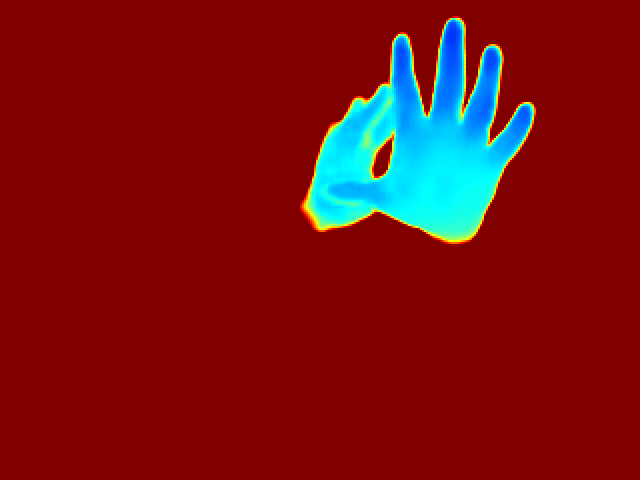}
	
	\includegraphics[width=0.195\textwidth]{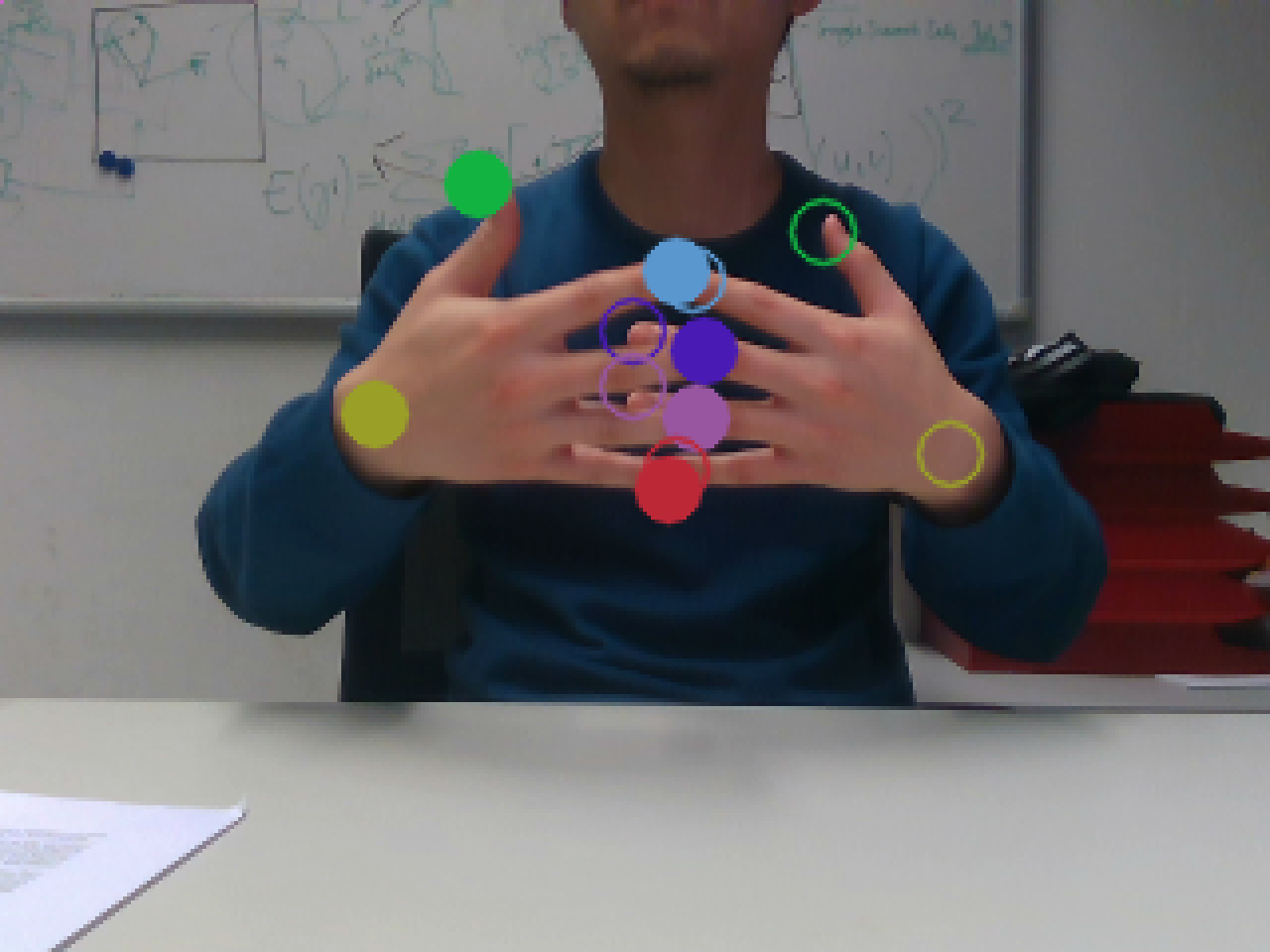}
	\includegraphics[width=0.195\textwidth]{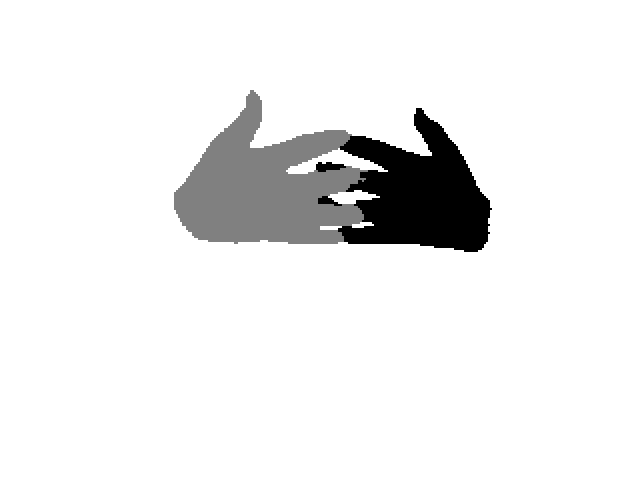}
	\includegraphics[width=0.195\textwidth]{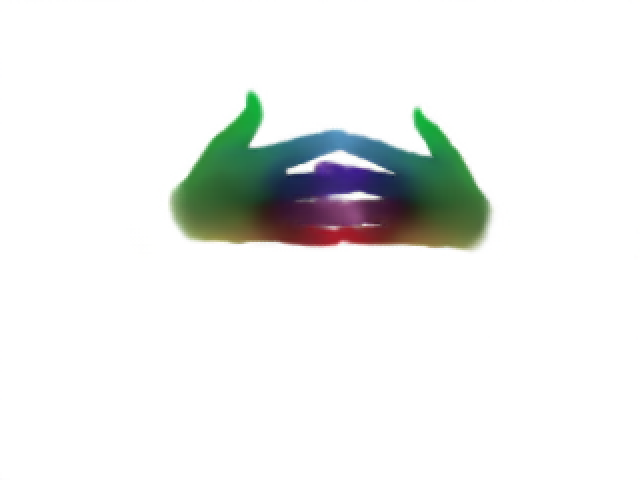}
	\includegraphics[width=0.195\textwidth]{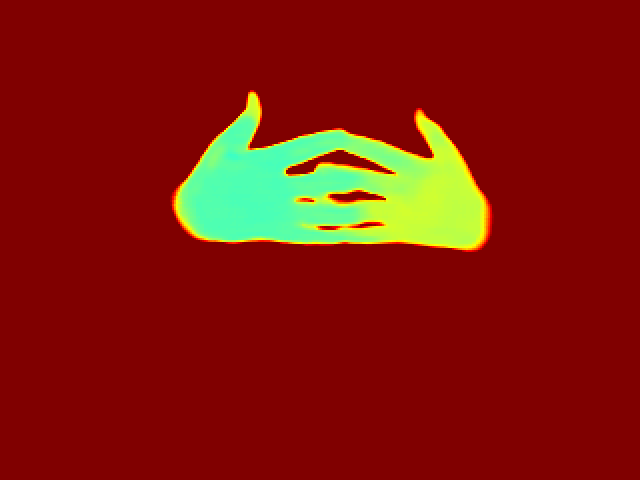}
	\includegraphics[width=0.195\textwidth]{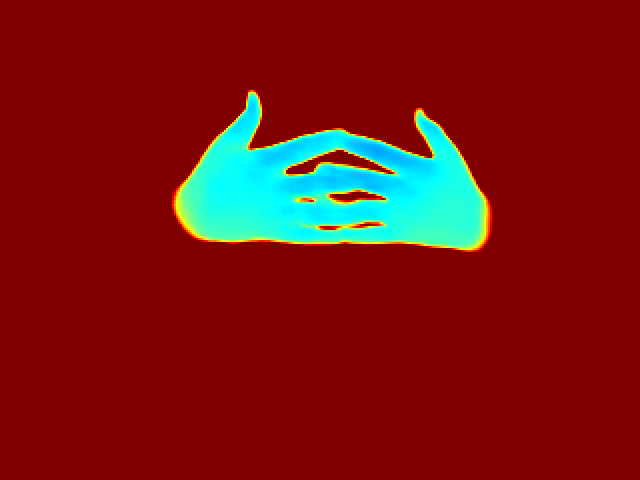}

	\caption{Visualizaiton of network outputs. From left to right: 2D keypoints, segmentation, dense matching map, inter-hand distance, intra-hand relative depth. 
	}
	\label{fig:network_outputs}
\end{figure*}

Our network architecture comprises two stages.
In the first stage our network performs per-pixel segmentation into left hand, right hand, and background pixels. 
Then, we branch into multiple subnetworks to regress dense matching, 2D keypoints, intra-hand relative depth, and inter-hand distance (the latter two using a shared multi-task subnetwork). 
The input for the second stage are both the original RGB input image, as well as the segmentation masks predicted in the first stage. 
Fig.~\ref{fig:network_outputs} shows all outputs predicted from test images.

\paragraph{Segmentation}
Let the image have height $h$ and width $w$.
Given only the RGB input image, the first-stage segmentation network predicts class-probability maps $\mathcal{S}' \in [0, 1]^{h \, \times \, w \, \times \, 3}$ for the three classes \textsc{left}, \textsc{right}, and \textsc{bg}.
We convert the probability maps to a segmentation mask $\mathcal{S} \in \{\textsc{left, right, bg}\}^{h \, \times w}$ by assigning the most probable class to each pixel.

\begin{figure}[t]
	\centering
	\subfigure{
		\includegraphics[trim={0pt 0pt 0pt 0pt} ,clip,width=0.25\columnwidth]{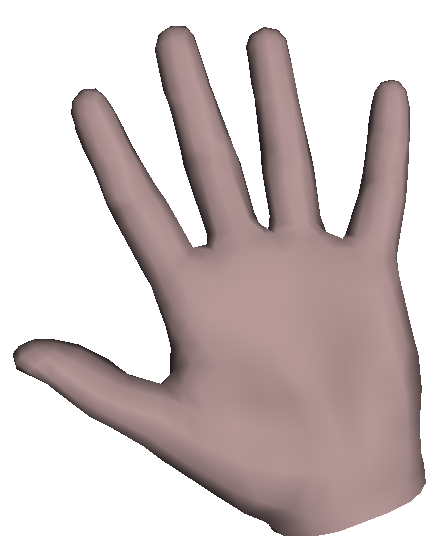}
	}
	\subfigure{
		\includegraphics[trim={0pt 0pt 0pt 0pt} ,clip,width=0.25\columnwidth]{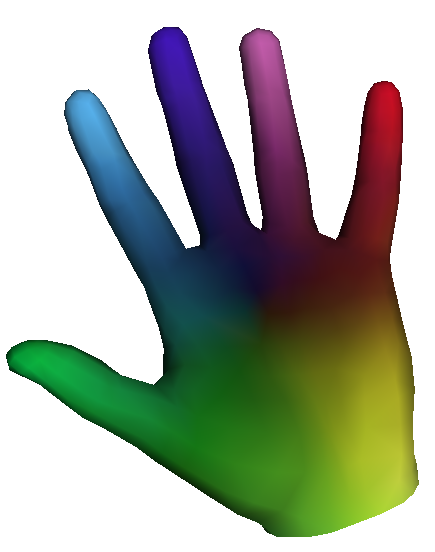}
	}
	\subfigure{
		\includegraphics[trim={0pt 0pt 0pt 0pt} ,clip,width=0.27\columnwidth]{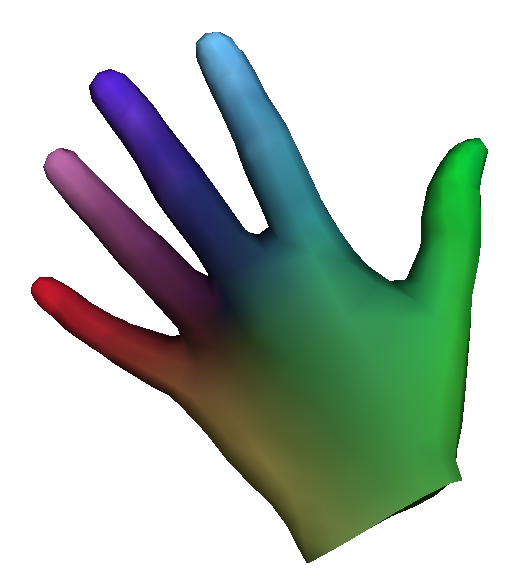}
	}
	\caption{Dense matching encoding of MANO model, front and back.}
	\label{fig:Encoding}
\end{figure}

\paragraph{Dense Matching}
The dense matching subnetwork regresses a dense matching image $\mathcal{M} \in \mathbb{R}^{h \, \times \, w \, \times \, k}$, where $k$ is the number of features.
Each pixel $\gamma = (u,v) \in \Gamma$ contains the feature vector $\mathcal{M}(\gamma) \in \mathbb{R}^k$ that uniquely determines the surface point of the 3D hand model which is visible at this pixel. 
We call the mapping from the feature vector to the 3D model surface \emph{dense matching encoding}. 
Note that the dense matching encoding is the same for the left and right hand, where we make use of the segmentation mask $\mathcal{S}$ for disambiguation.
We use the same encoding as \cite{mueller_siggraph2019} to embed the hand surface to a 3D feature space for our dense matching map. This is done using the method of \cite{bronstein2006multigrid} to approximately preserve geodesic distances in the feature space. We then map the feature space to an HSV color space cylinder which results in each finger being assigned a different hue.
We denote the extended feature vector at vertex $\bm{x}$ as $\eta'(\bm{x}) \in \mathbb{R}^{k+1}$ and define $\eta'(\bm{x}) = [\eta(\bm{x}), s(\bm{x})]$, where $\eta: \mathcal{X} \rightarrow \mathbb{R}^k$ is the original dense matching encoding defined on the mesh. The scalar $s(\bm{x})$ yields a different value $\sigma(\textsc{right})$ or $\sigma(\textsc{left})$ that encodes which hand $\bm{x}$ belongs to.
We can then measure the \emph{matching distance} between 3D hand model vertices $\bm{x}$ and pixels $\gamma$ in the image as 

\begin{equation}
\Delta_{\mathcal{M,S}}(\gamma, \bm{x}) = || \, [\mathcal{M}(\gamma),\sigma(\mathcal{S}(\gamma))] - \eta'(\bm{x}) \, ||_2 \\
\,.
\end{equation}

We formulate the dense matching map $\psi: \mathcal{X} \rightarrow \Gamma$ to establish correspondence between model vertices and the RGB image as
\begin{align}
 \psi'(\bm{x}) &= \arg\min_{\gamma \in \Gamma} \, \Delta_{\mathcal{M,S}}(\gamma, \bm{x}) 
\\
     \psi(\bm{x}) &= 
    \begin{cases}
     \psi'(\bm{x}) \, , &\text{if} \: \Delta_{\mathcal{M,S}}( \psi'(\bm{x}), \bm{x}) < t_c \\
    \emptyset \, , &\text{otherwise}
    \end{cases} 
     \, .
\end{align}
If the minimum distance of vertex $\bm{x}$ to all pixels is larger than the threshold $t_c$, this vertex is likely not visible and we set  $\psi(\bm{x}) = \emptyset$.
The calculation of the dense matching map  $\psi$ is efficiently implemented using parallel reduction in CUDA.
The dense matching encoding $\eta(\cdot)$ is defined analogously to the approach by \citet{mueller_siggraph2019} with $k = 3$.

Furthermore, we set $\sigma(\textsc{left}) = 0.0$ and $\sigma(\textsc{right}) = 0.5$.

\paragraph{Intra-Hand Relative Depth}
The network further learns to predict an intra-hand relative depth map $\mathcal{D}_{\text{intra}} \in \mathbb{R}^{h \, \times \, w}$.
For each hand pixel, it contains the estimated depth difference of this hand point to the root of the respective hand.
Note that $\mathcal{D}_{\text{intra}}$ is scale-normalized due to the inherent ambiguity in RGB images.
We multiply it with the palm length $\alpha$ to obtain the \emph{metric} relative depth map $\mathcal{Q}_{\text{intra}}$, which is used for 3D model fitting (cf.~\autoref{eq:intra}).
 
\paragraph{Inter-Hand Distance}
Our multi-task CNN also learns to estimate the distance in depth between the two hands.
Instead of predicting a single scalar, we predict the distance as image $\mathcal{D}_{\text{inter}} \in \mathbb{R}^{h \, \times \, w}$.
This allows us to use a fully-convolutional network and thereby enables feature sharing with the intra-hand depth prediction task.
Every pixel in $\mathcal{D}_{\text{inter}}$ that belongs to a hand contains the distance of its root joint from the other hand's root (in the case for only a single hand being visible, we assign a constant value to all pixels).
Note that each pixel in the output can thus be seen as member of an ensemble.
Analogous to the intra-hand relative depth, we also normalize the inter-hand distance with the size of the hand for training.
We summarize the ensemble with one relative distance value $d_h$ per hand by calculating the median over all pixels that are predicted to belong to the respective hand based on the segmentation mask $\mathcal{S}$,~i.e.
\begin{equation}
    d_h = \underset{\gamma \in \Gamma,\\\mathcal{S}(\gamma) = h}{\operatorname{median}} \mathcal{D}_{\text{inter}}(\gamma) \, .
\end{equation}
We set the robust relative distance $d_{\text{inter}} = \operatorname{mean}(d_{\text{left}}, -d_{\text{right}})$.
When the two hands are close, $d_{\text{left}}$ and $d_{\text{right}}$ can be degenerate and have the same sign. 
In this case, $d_{\text{inter}}$ is set to 0.
For the model fitting, we can then define the \emph{metric} absolute distance $q_{\text{inter}} := \alpha \cdot d_{\text{inter}}$ (cf. \autoref{eq:inter}).

\paragraph{2D Keypoints}
Let $\mathcal{J}_{\text{total}}$ be the set of all 12 keypoints, namely the fingertips and wrist of each of the two hands.
We formulate the 2D keypoint estimation as heatmap regression task.
The network outputs heatmaps $\mathcal{H} \in \mathbb{R}^{h \, \times \, w \, \times |\mathcal{J}_{\text{total}}|}$, a one-channel image for each of the keypoints.
Each ground-truth heatmap contains a Gaussian with radius $0.07 \cdot r_{\text{c}}$, where $r_{\text{c}}$ is the edge length of the larger edge of a tight hand crop, scaled to have maximum value 1, centered at the 2D keypoint position.
Note that the ground truth is also provided for occluded keypoints which enables the network at test time to predict keypoint locations under strong occlusions which are common for two-hand interactions.
We extract the maximum location of each predicted heatmap as 
\begin{equation}
    \gamma^{\max}_j = \arg\max_{\gamma \in \Gamma} \, \mathcal{H}(\gamma,j) \, .
\end{equation}
We use a threshold $t_h$ to filter out low-confidence estimates and obtain the 2D keypoint location as
\begin{align}
    \mathcal{Q}_{\text{key}}(j) = 
    \begin{cases}
    \gamma^{\max}_j \, , &\text{if} \: \mathcal{H}(\gamma^{\max}_j,j) > t_h \\
    \emptyset \, , &\text{otherwise}
    \end{cases}
     \, .
\end{align}

\subsection{Network Architecture and Training}
Our network consists of several subnetworks as shown in \autoref{fig:overview}.
Each subnetwork is a U-Net~\cite{unet2015} with 4 layers for down-sampling and 4 layers for up-sampling, resulting in a bottleneck resolution of $\frac{h}{16} \, \times \, \frac{w}{16}$.
We use skip connections between layers of the same resolution in the down- and up-sampling stream to better preserve local information.
We employ instance normalization instead of batch normalization at every layer as proposed by \citet{ulyanov2016instance}.

We use the softmax cross-entropy loss for the segmentation prediction and $\ell_2$-losses for all other outputs.
For real data, we use a loss mask to disable the losses for holes in the annotations, which are present due to the projection between the depth and color channel.
Appendix~\ref{appendix:loss_mask} describes the annotation transfer from the depth to the color image.
We train the whole network end-to-end for 400k iterations using Adam with a learning rate of $0.001$ and a beta of $0.9$.
We perform data augmentations on-the-fly to further increase the diversity of our training set (see Appendix~\ref{appendix:data_aug}).
\section{Training Data}
\label{sec:training_data}
For training our regressor in a supervised manner, for a given RGB image containing two potentially interacting hands, we ideally require a ground-truth relative depth map $\mathcal{D}_{\text{intra}}^{\text{GT}}$, the relative inter-hand distance map $\mathcal{D}_{\text{inter}}^{\text{GT}}$, a dense matching image $\mathcal{M}^{\text{GT}}$, and 2D joint position heatmaps $\mathcal{H}^{\text{GT}}$. 
Existing datasets like the \emph{Rendered Hands Dataset (RHD)} \cite{Zimmermann:2017um} or \emph{Panoptic} \cite{Joo_2017_TPAMI} only provide a subset of the required annotations (see \autoref{table:hand_datasets}) and, in particular, do not have dense matching annotations. 
The former does also not show realistic and physically plausible close two-hand interactions, an important requirement for our setting.
The recent FreiHand dataset~\cite{Zimmermann_2019_ICCV} provides crops of single hands with annotated MANO fits, sometimes even with objects, but no two-hand frames. Generating synthetic interacting hands images from these would require compositing and would lead to unrealistic interaction.
Therefore, since manual annotation of the labels we require is impossible, we propose a new set of strategies to obtain annotations for both real and synthetic images.
We add the existing datasets \emph{RHD} and \emph{Panoptic} to our own real and synthetic datasets to increase data diversity and hence improve generalization. 
\autoref{table:hand_datasets} presents a summary of the different datasets used for training, and gives details about the ground-truth annotations available in each of them.
In the following, we describe the procedure for creating our own synthetic an real datasets.
Furthermore, in Sec.~\ref{sec:ablation} we present an ablation study that demonstrates how our real data (with slightly noisy annotations) helps bridge the real-synthetic domain gap, and the perfectly annotated synthetic data mitigates influence of noise.

\paragraph{Real Data.}
We leverage the state-of-the-art depth-based two hand tracker of \citet{mueller_siggraph2019} to track sequences of two hands in interaction with an RGB-D sensor that captures synchronized color and depth images.
We record in front of a green screen to enable background augmentation as post-processing.
Mueller's approach outputs MANO \cite{Romero_siggraphasia2017} per-frame shape $\bm{\beta}$ and pose $\bm{\theta}$ parameters, which, in combination with the extrinsic parameters of the RGB and depth sensors of the camera, allows us to reproject the surface of the tracked hand to the RGB image. For details on the reprojection see Appendix~\ref{appendix:loss_mask}.
Subsequently, we are able to compute relative depth maps $\mathcal{D}_{\text{intra}}^{\text{GT}}$, inter-hand relative distance maps $\mathcal{D}_{\text{inter}}^{\text{GT}}$, and dense matching images $\mathcal{M}^{\text{GT}}$ from the real RGB image.
Additionally, we use 2D keypoint positions from \cite{Joo_2017_TPAMI} to construct 
heatmaps $\mathcal{H}^{\text{GT}}$ for supervision.
Since tracking a single hand is usually more robust and accurate than tracking two interacting hands, we also include single-hand sequences in our data.
We then employ depth-based compositing to obtain images depicting two hands, see \autoref{appendix:data_aug}.
Note that we manually cleaned bad tracking results and 2D keypoint predictions by visual inspection to ensure reasonable quality in our real data annotations.

{
\hspace{-1cm}
 \small
 \begin{table}[t]
 	\caption{
     Available annotations in existing hand tracking datasets and ours.
 	}
 	\setlength{\tabcolsep}{5pt}
 	\makebox[\linewidth][l]{
 	\begin{tabular}{lcccccc}
 		& \rot{Segmentation} & \rot{Dense Corrs.} & \rot{Intra-Hand} & \rot{Inter-Hand} & \rot{2D Keypoints}\\
 		\midrule
 		Our (Synth) & \cmark & \cmark & \cmark & \cmark & \cmark \\
 		Our (Real) & \cmark & \cmark & \cmark & \cmark & \cmark \\ 
 		RHD~\cite{Zimmermann:2017um} & \cmark & \xmark & \cmark & \cmark & \cmark\\ 
 		Panoptic~\cite{Joo_2017_TPAMI}& \xmark & \xmark & \xmark & \xmark  & \cmark\\ 
 		\bottomrule
 	\end{tabular}
 	}
 	\setlength{\tabcolsep}{6pt}
 	\label{table:hand_datasets}
 \end{table}

 }

\paragraph{Synthetic Data.} 
The above-described approach to annotate real data is not perfect. 
In some poses the depth-based tracker may exhibit tracking errors.
Also, the RGB-D camera has separate depth and RGB optics which are apart by a small baseline.
The resulting parallax leads to some occlusion-disocclusion-related holes in the annotations when reprojecting them from the depth channel to the color channel.
This makes our real data not sufficiently accurate and unable to produce annotations for highly-challenging poses.
We address this issue by complementing our real dataset by synthetically generating images with their corresponding annotations.
To this end, and similar in spirit to \citet{Zhao_TOG2013} and \citet{mueller_siggraph2019}, we employ a motion capture-driven physics-based simulation to generate physically-correct hand sequences (\textit{e.g.}, without self-collisions, with accurate inter-hand contact, and with a soft-skin layer) where two hands realistically interact in a large variety of poses.
To increase the realism and variety of simulated hand sequences, and in contrast to existing approaches that use a hand template of fixed shape and appearance in the simulation framework, we extend the surface-based parametric model of MANO to a volumetric representation that is subsequently fed into the simulation \cite{verschoor2018soft}.
This allows us to synthesize complex hand motions driven by a motion capture sequence, including 2D keypoint positions and heatmaps $\mathcal{H}^{\text{GT}}$, dense correspondence images $\mathcal{C}^{\text{GT}}$, relative depth maps $\mathcal{D}_{\text{intra}}^{\text{GT}}$, and relative inter-hand distance maps $\mathcal{D}_{\text{inter}}^{\text{GT}}$, with varying subject identities.
We can therefore generate data with varying hand shapes.

Additionally, we further extended the MANO model with photorealistic appearances by a standard texture mapping approach.
Hand textures were generated by reprojecting multi-camera imagery into a still hand image to which the MANO model was fitted.
In practice, we generate 10 different hand textures, which include a variety of actors of different ethnicities, genders, and hand shapes.
The ability to render physically plausible two-hand interactions for various hand shapes and appearances enables our approach to generalize better to real world scene diversity.

\section{Experiments}
\label{sec:eval}

\begin{figure}[t]
    \includegraphics[width=\linewidth]{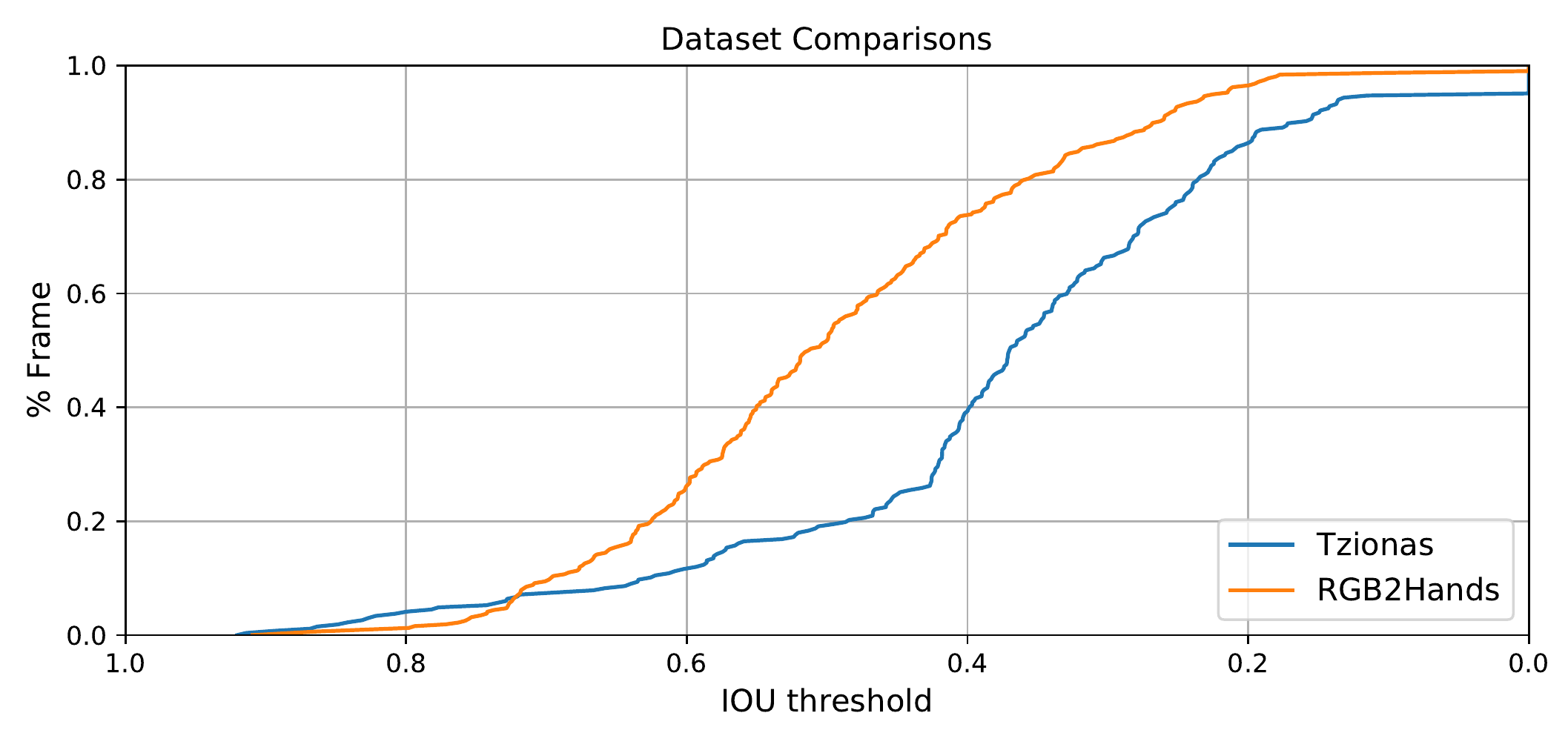}
    \caption{
    The two-hand \textsc{Tzionas} dataset has significantly fewer frames with strongly interacting and overlapping hands compared to our \textsc{RGB2Hands} dataset. We plot the percentage of frames (y-axis) where the overlap (in terms of the \emph{intersection over union}, IOU) of the left and right hand bounding box is greater than a certain threshold (x-axis).
    }
    \label{fig:datasets_IoU} 
\end{figure}

\begin{figure*}[t] 
\minipage[t]{0.32\textwidth}
  \includegraphics[width=\linewidth,  draft=false]{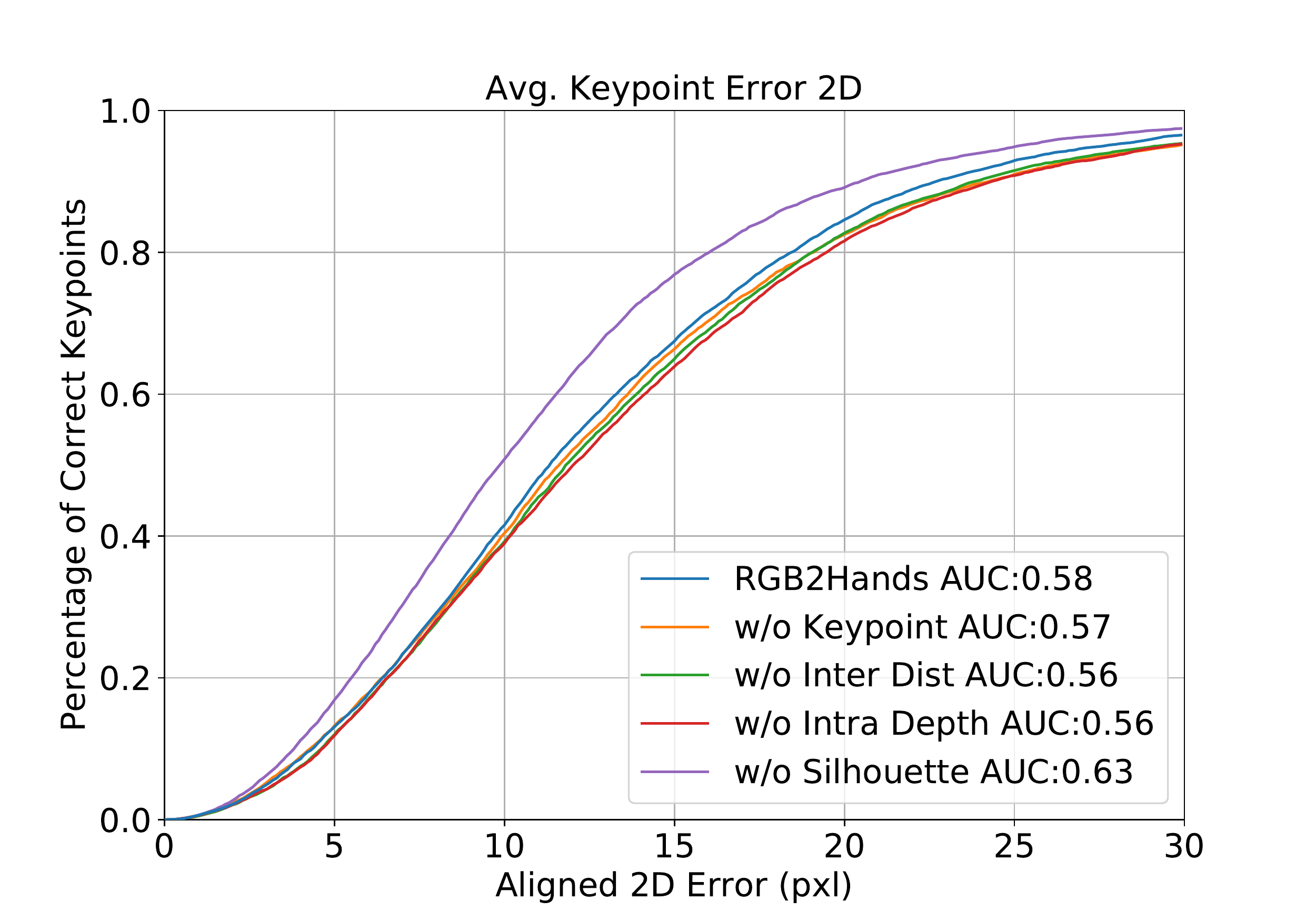}
  \includegraphics[width=\linewidth,  draft=false]{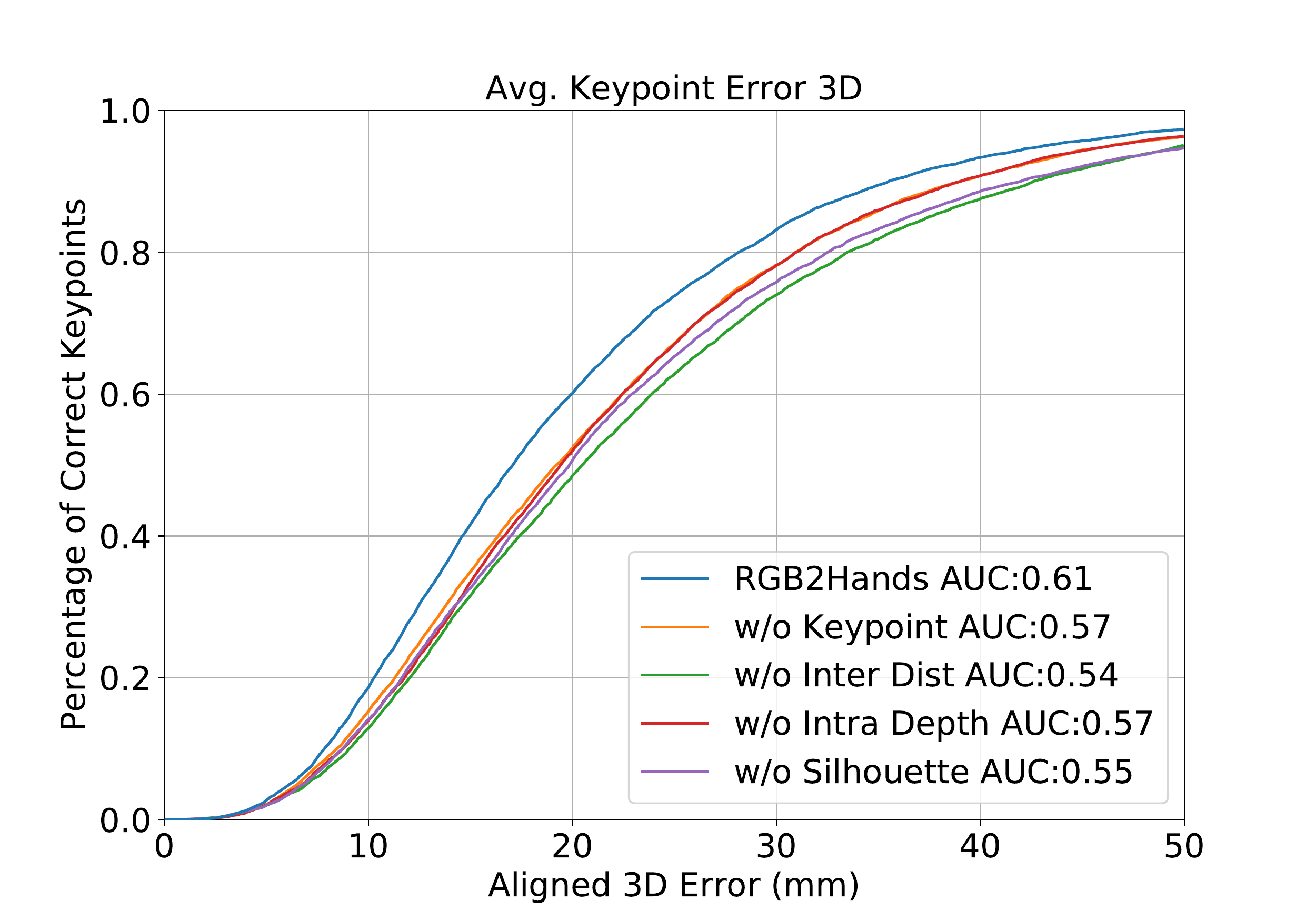}
    \caption{
    Energy term ablation study on the \textsc{RGB2Hands} dataset. 
    All terms except the silhouette term improve the results in 2D and 3D.
    The silhouette term improves the 3D error at the cost of 2D error.
    We hypothesize that it reshapes the energy landscape to have fewer local minima with accurate 2D but inaccurate 3D pose.
    }
    \label{table:ablation_energy} 
\endminipage\hfill
\minipage[t]{0.32\textwidth}
    \includegraphics[width=\linewidth,  draft=false]{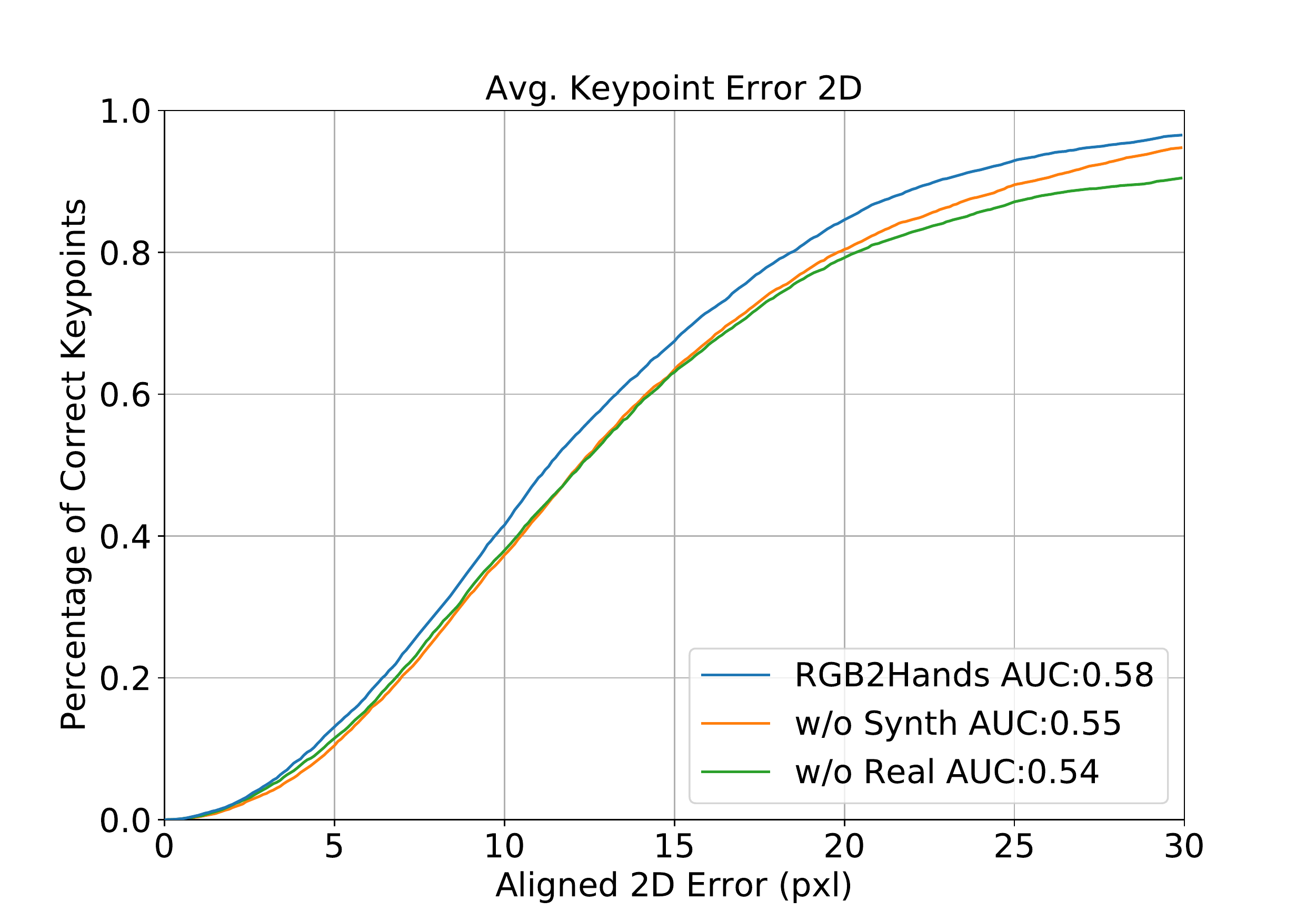}
    \includegraphics[width=\linewidth,  draft=false]{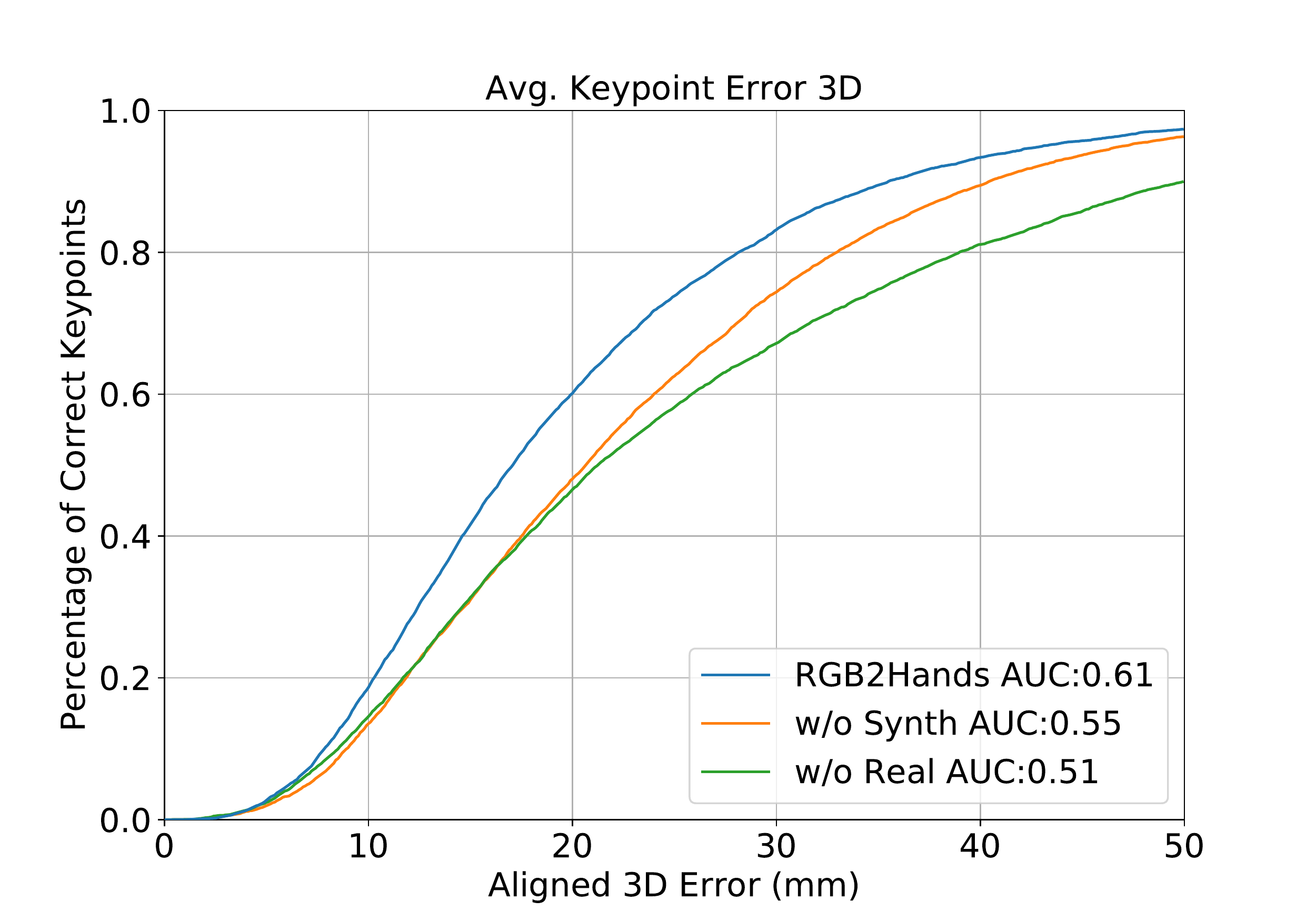}
    \caption{
    Training data ablation study on the \textsc{RGB2Hands} dataset. Training with all datasets (blue line) outperforms version where we leave out the synthetic dataset (orange line), or the real dataset (green line).
    }
    \label{fig:ablation_data} 
\endminipage \hfill
\minipage[t]{0.32\textwidth}
    \includegraphics[width=\linewidth,  draft=false]{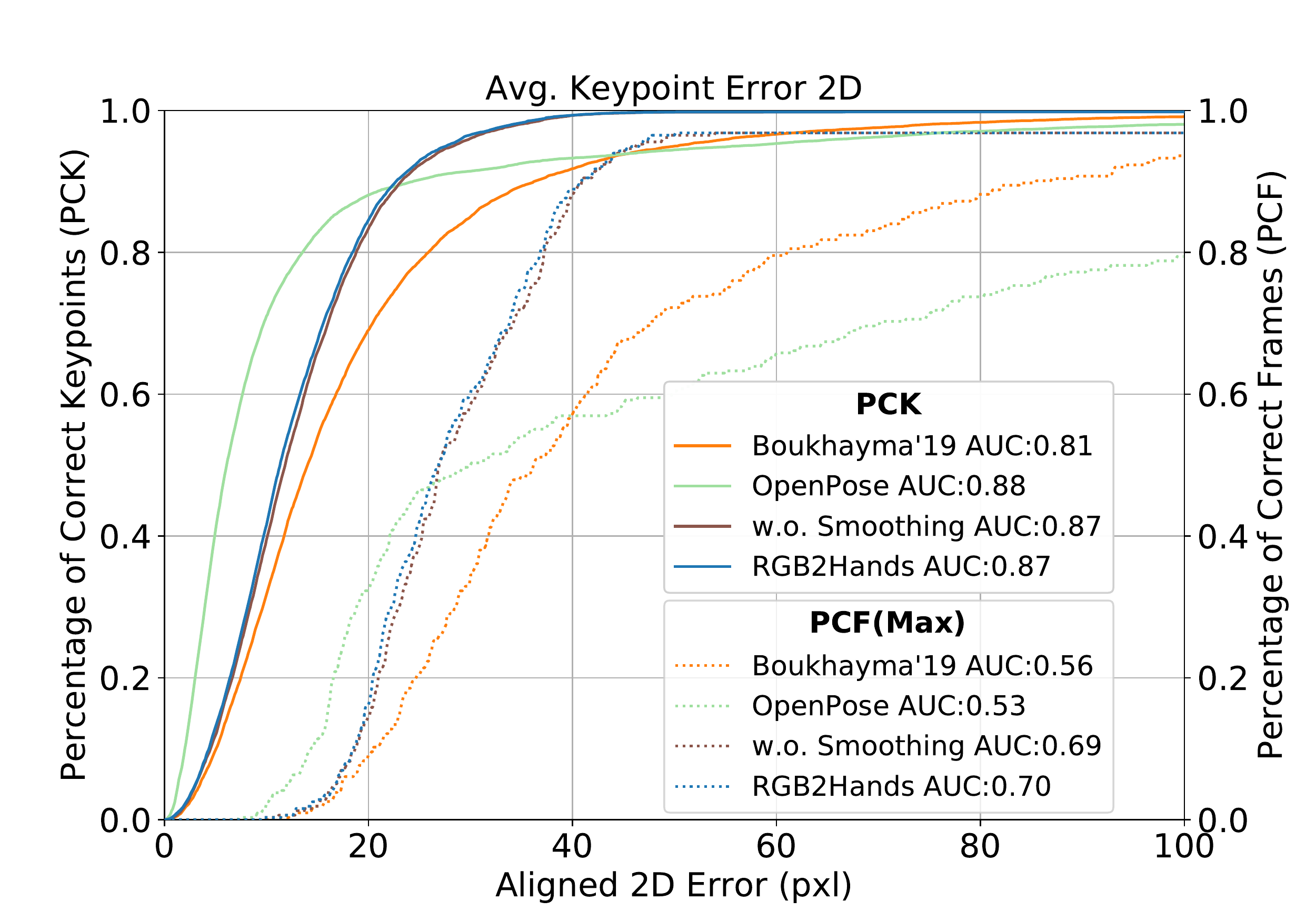}
  \includegraphics[width=\linewidth,  draft=false]{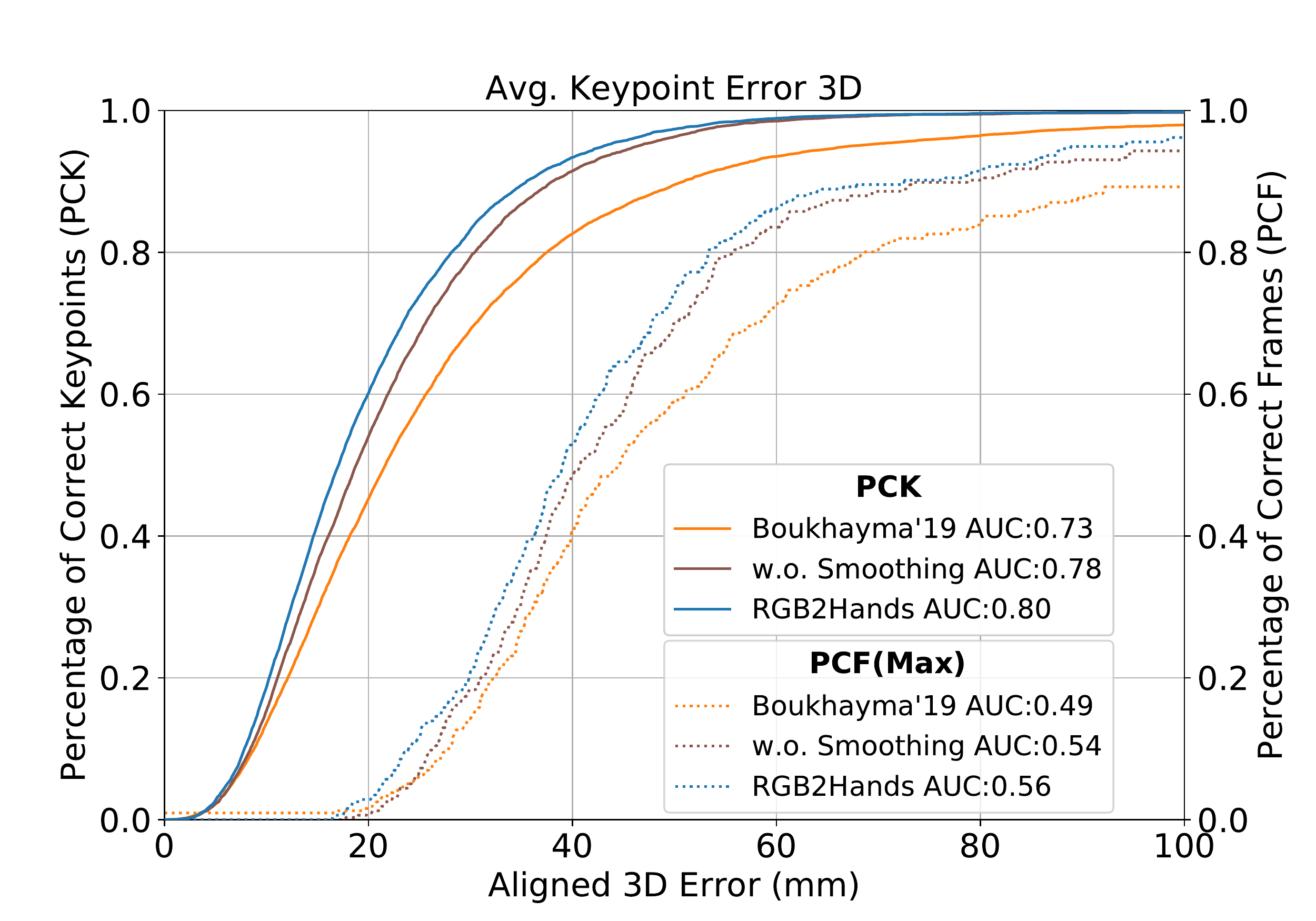}
  \caption{
  Quantitative comparison of our method to \citet{Boukhayma_2019_CVPR} and OpenPose on the \textsc{RGB2Hands} dataset. 
  Our method significantly outperforms the method by Boukhayma et al.
  Both obtain dense 3D results whereas OpenPose only estimates sparse 2D keypoints and exhibits higher per-frame maximum keypoint error (dotted lines).
  }
    \label{fig:comparison_quant} 
\endminipage
\end{figure*}

In this section we experimentally evaluate our proposed RGB two-hand tracking approach in order to demonstrate its merits. 
We first introduce the datasets and metrics used in our evaluation.
Subsequently, we conduct an ablation study that evidences the importance of the individual components. 
Afterwards, we compare our method quantitatively and qualitatively to other related works. Moreover, we present additional qualitative two-hand tracking results.
We refer the reader to our supplementary video for animated results.

\subsection{Datasets and Metrics}
Although the dataset by \citet{tzionas_ijcv2016}  is commonly used to evaluate two-hand tracking methods, we have found that it is not well-suited for evaluating two-hand tracking methods with \emph{strong interactions}. This is because in their dataset 
only very few frames actually exhibit close two-hand interactions. For a more comprehensive evaluation of challenging interaction settings, we therefore introduce a new benchmark dataset, \textsc{RGB2Hands}, which exhibits stronger interactions and more overlap between the left and right hand.
In Fig.~\ref{fig:datasets_IoU},  we illustrate that  our \textsc{RGB2Hands} dataset contains more frames with stronger hand-hand interactions compared to the dataset by \citet{tzionas_ijcv2016}, which we measure in terms of the overlap of the bounding box from the left and right hand.

In the following, we present details of both datasets as well as the evaluation metrics.

\paragraph{\textsc{Tzionas} Dataset}
The \textsc{Tzionas} dataset contains 7 two-hand sequences with a total number of 1{,}307 RGB-D frames.
2D annotations on the depth image are provided every 5th frame for the 14 interior joints of each hand when visible.
The camera calibration can be used to obtain 3D annotations by backprojection.

\paragraph{\textsc{RGB2Hands} Dataset}

Our new dataset \textsc{RGB2Hands} has a total of 1{,}724 frames which are divided into 4 sequences, where each sequence contains between 316 and 572 frames.
To enable 3D evaluation, we recorded synchronized depth data. Using the depth camera calibration, 3D annotations can be obtained for the visible keypoints by backprojection.
For quantitative comparisons, out of the 4 sequences, at least every 5th frame was annotated starting from the beginning of the interaction, resulting in a total of 319 annotated frames. 
The annotation was performed manually, where annotators were asked to identify the 14 interior joints of each hand as done for previous datasets~\cite{tzionas_ijcv2016}.
If the location of an occluded joint could be inferred with high confidence, annotators marked this location while also flagging the occlusion to signify that depth cannot be recovered for 3D evaluation.
If no reliable guess was possible, this joint was not annotated. 
Note that this is an advantage over the \textsc{Tzionas} dataset where only visible joints are annotated.

\paragraph{Metrics}
For our quantitative comparisons in 2D and 3D we use two metrics to compare the errors between the annotated ground-truth keypoints and corresponding estimates obtained by our (or other) methods. 
First, we use the mean per-keypoint error in pixels for 2D or in millimeters for 3D.
Second, to enable a more fine-grained analysis, we also employ the \emph{Percentage of Correct Keypoints (PCK)} metric in 2D and 3D.
A keypoint estimate is counted as correct if its distance from the ground truth is less than $t_{\text{PCK}}$.
By varying the threshold $t_{\text{PCK}}$ on the horizontal axis, and showing the respective value on the vertical axis, a PCK curve is plotted.
To address the inherent depth-scale ambiguity of RGB images in the 3D evaluation, the estimated keypoints were aligned to the ground truth using Procrustes analysis without rotation.
Note that the alignment is performed for both hands jointly, i.e. a single translation and scale value is estimated for both. 
Hence, our aligned 3D error still captures the quality of the relative hand placement in 3D.

\subsection{Ablation Study}
\label{sec:ablation}

\begin{figure}[t]
	\centering
		\subfigure[Without silhouette]{
		\includegraphics[trim={60pt 90pt 220pt 280pt},clip,width=0.45\columnwidth]{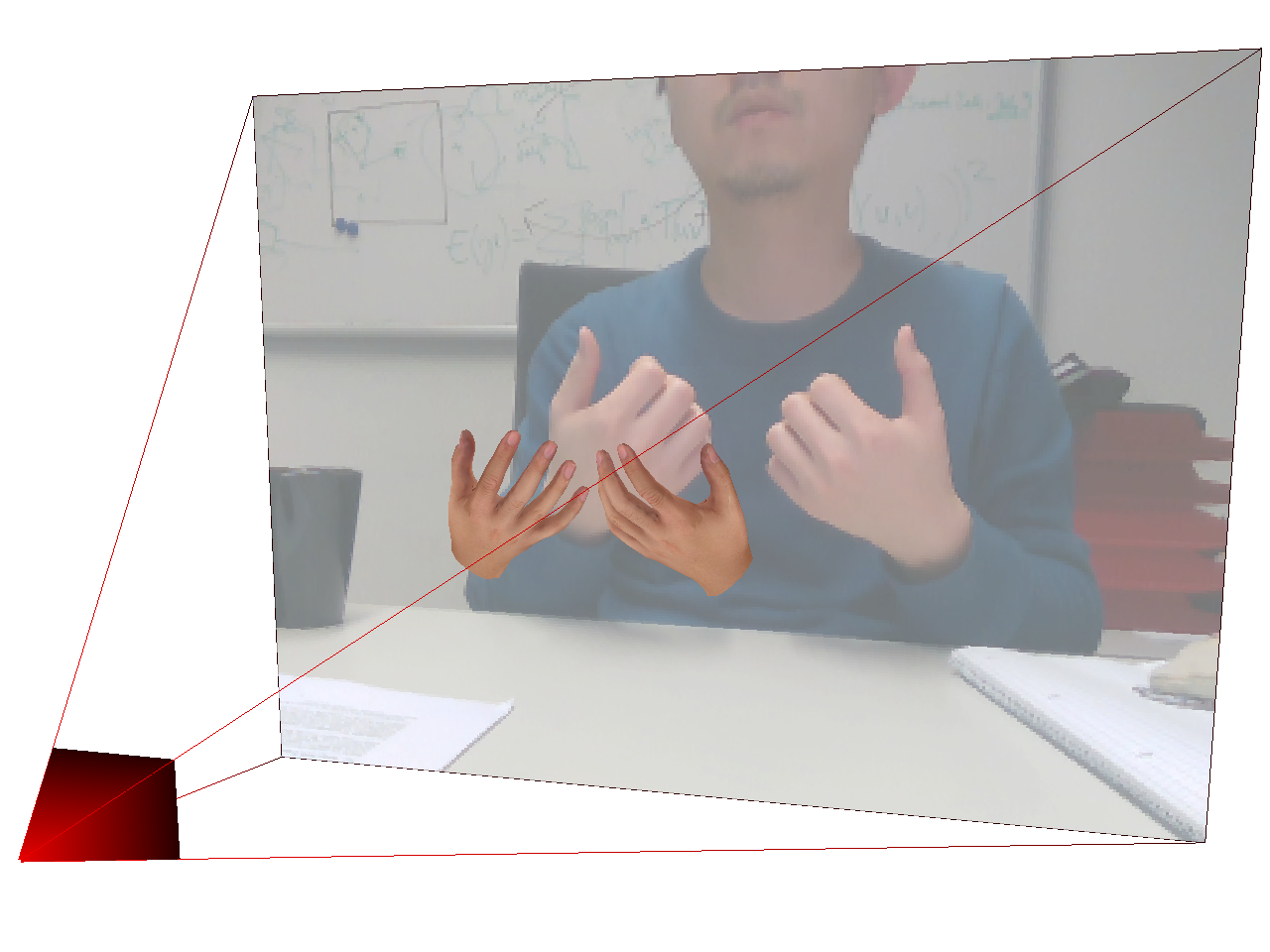}
	}
	\subfigure[With silhouette]{
		\includegraphics[trim={60pt 90pt 220pt 280pt},clip,width=0.45\columnwidth]{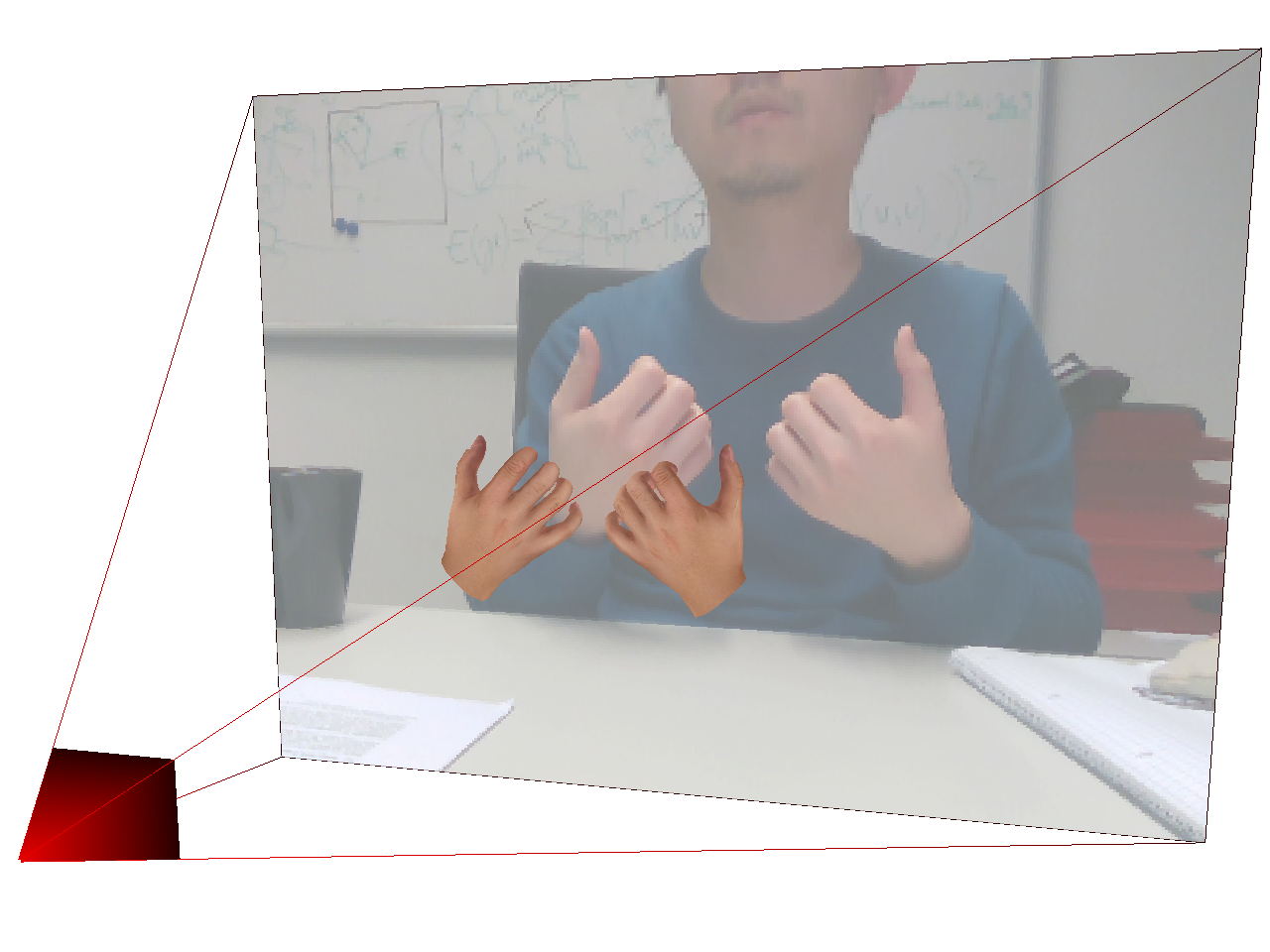}
	}
	\subfigure[Without 2D keypoints]{
		\includegraphics[trim={60pt 90pt 220pt 280pt},clip,width=0.45\columnwidth]{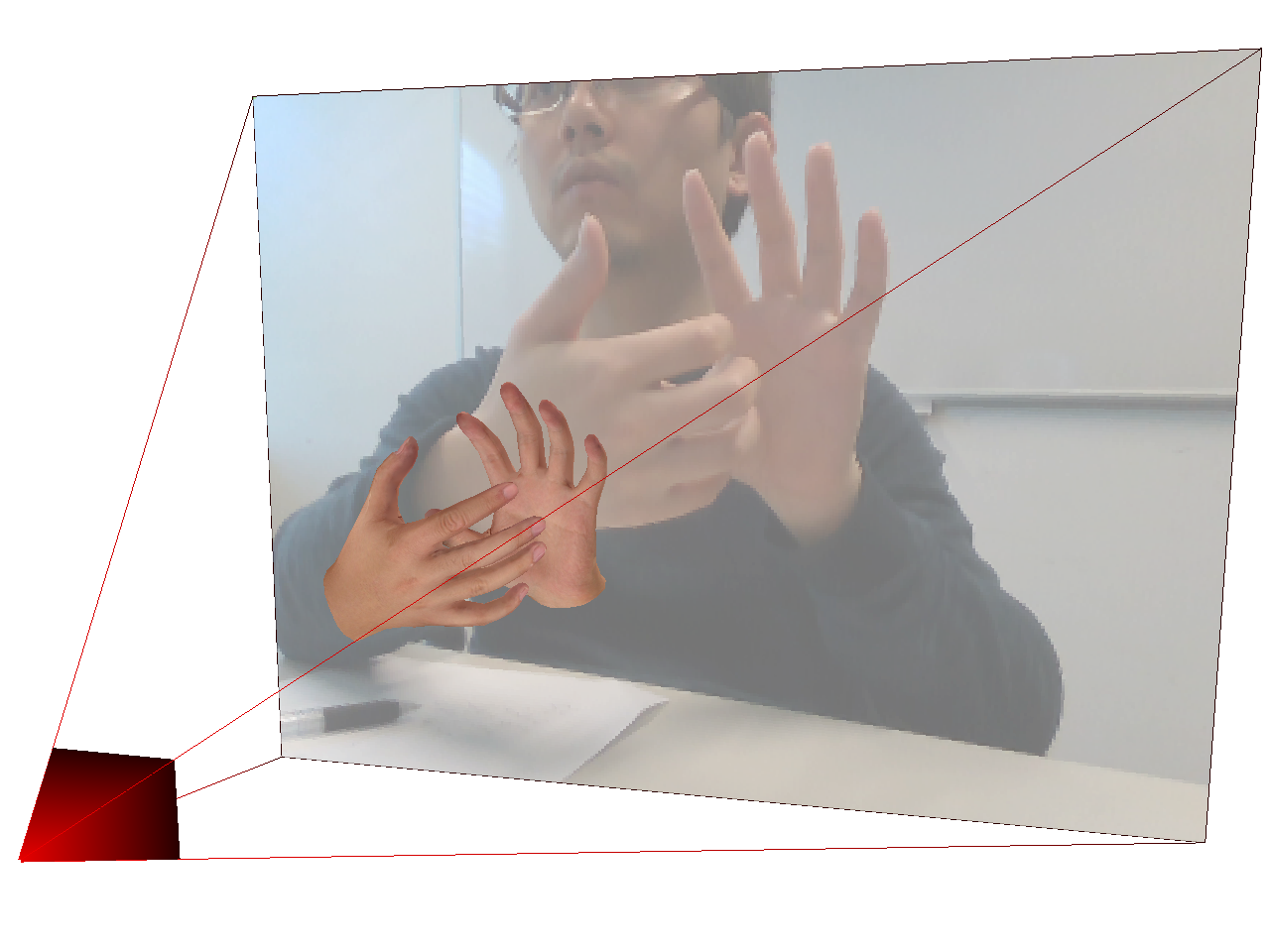}
	}
	\subfigure[With 2D keypoints]{
		\includegraphics[trim={60pt 90pt 220pt 280pt},clip,width=0.45\columnwidth]{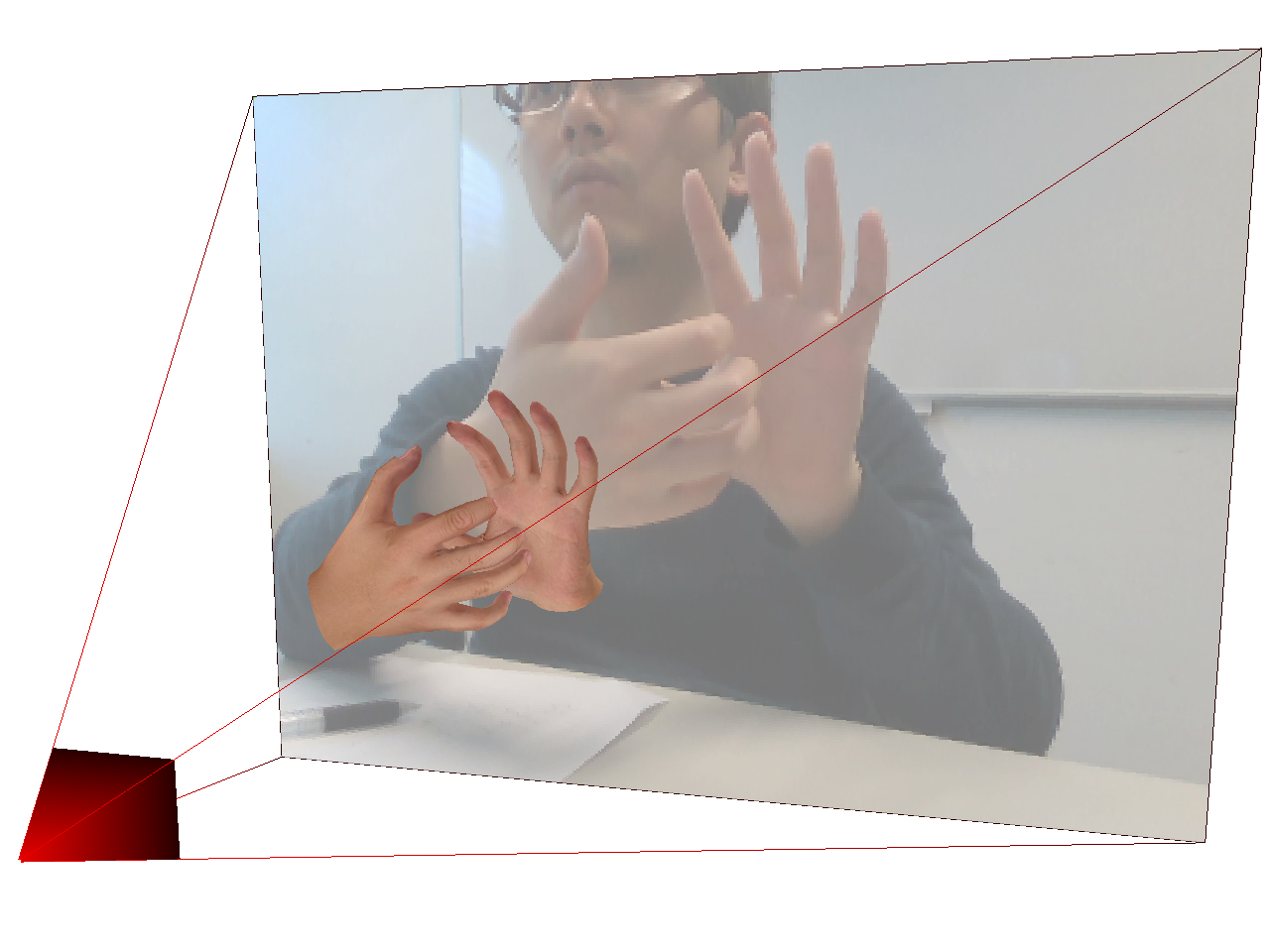}
	}
	\subfigure[Without inter-hand distance]{
		\includegraphics[trim={60pt 90pt 220pt 280pt},clip,width=0.45\columnwidth]{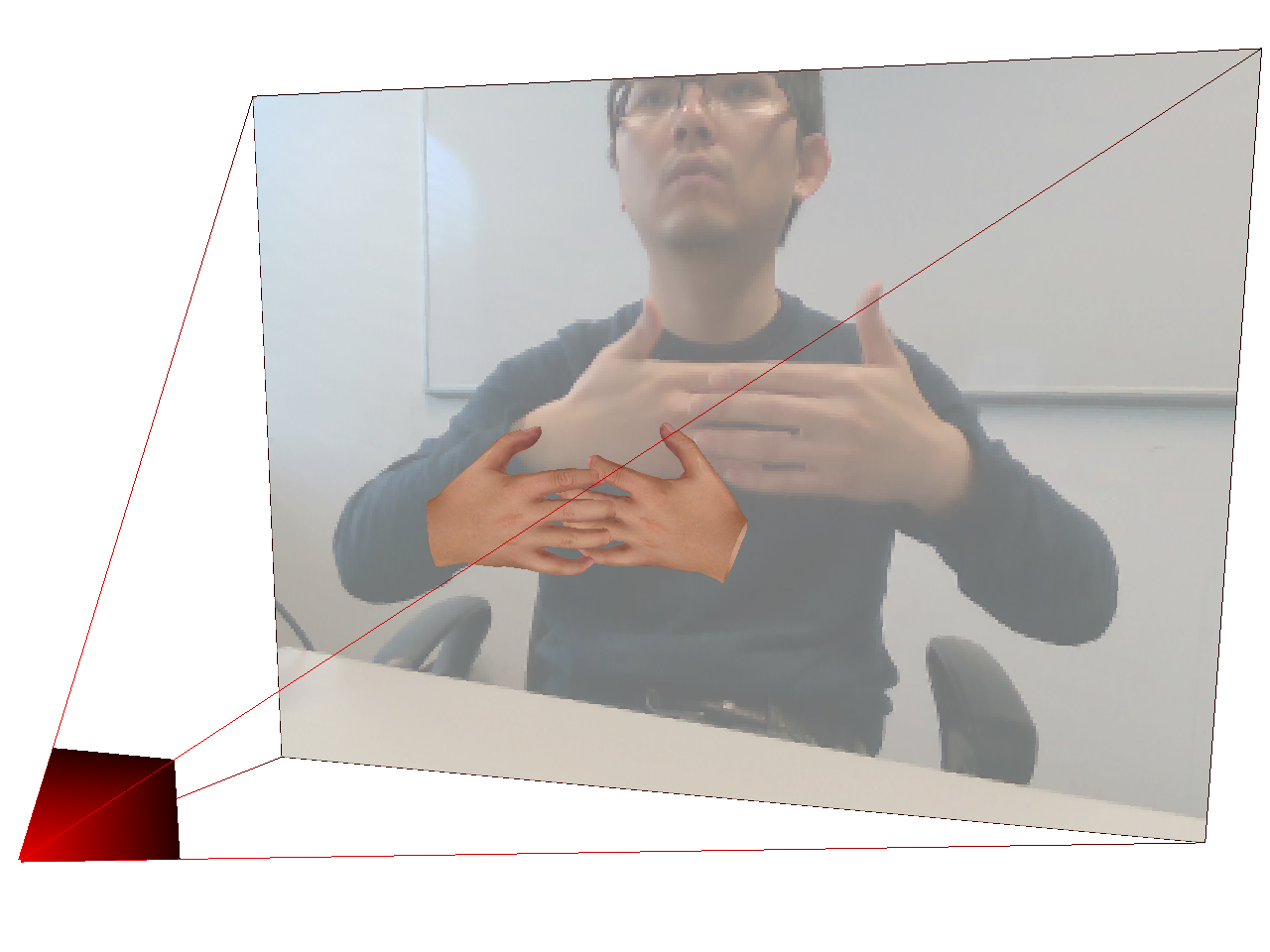}
	}
	\subfigure[With inter-hand distance]{
		\includegraphics[trim={60pt 90pt 220pt 280pt},clip,width=0.45\columnwidth]{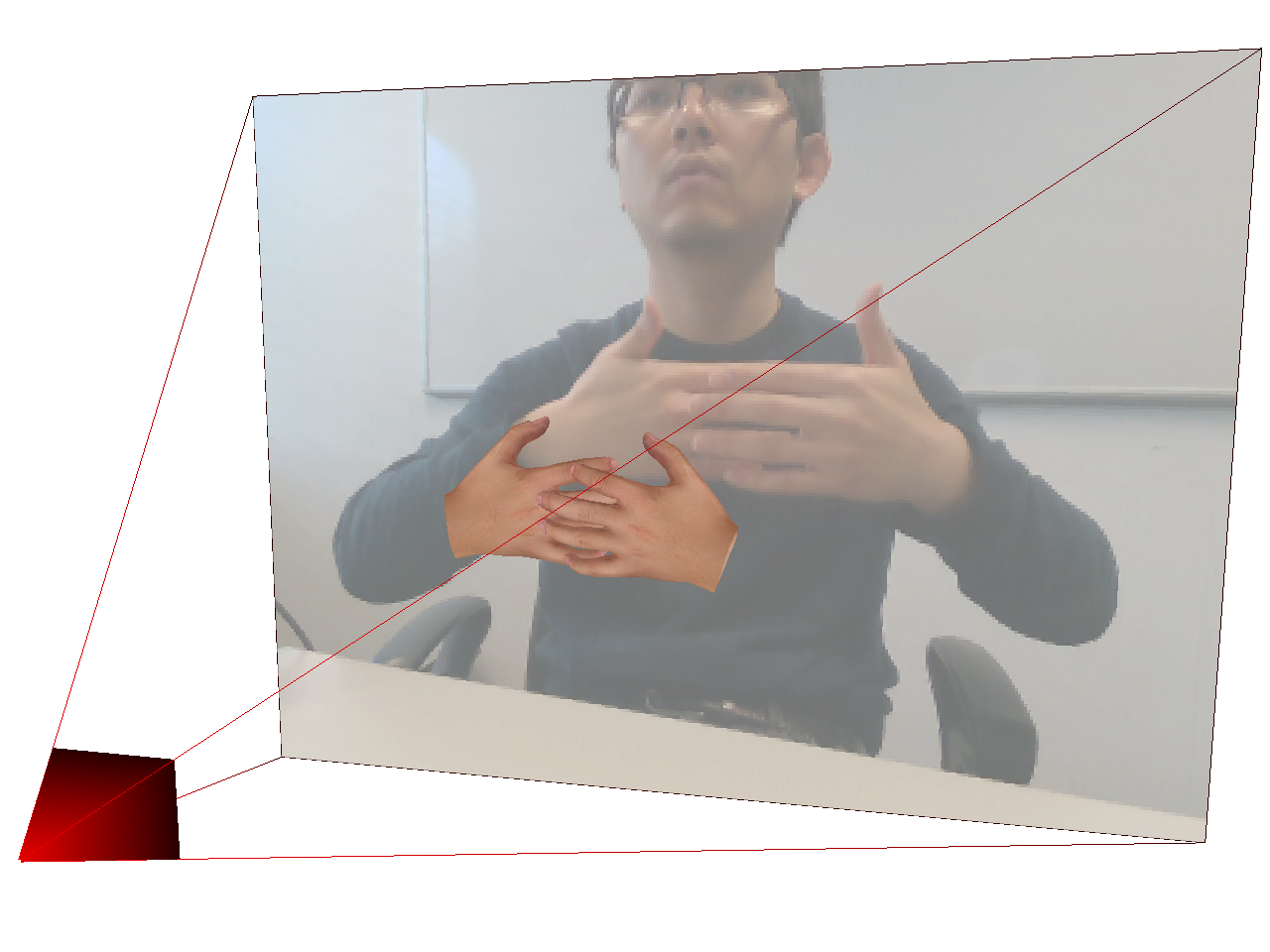}
	}
	\subfigure[Without intra-hand distance]{
		\includegraphics[trim={60pt 90pt 40pt 100pt},clip,width=0.45\columnwidth]{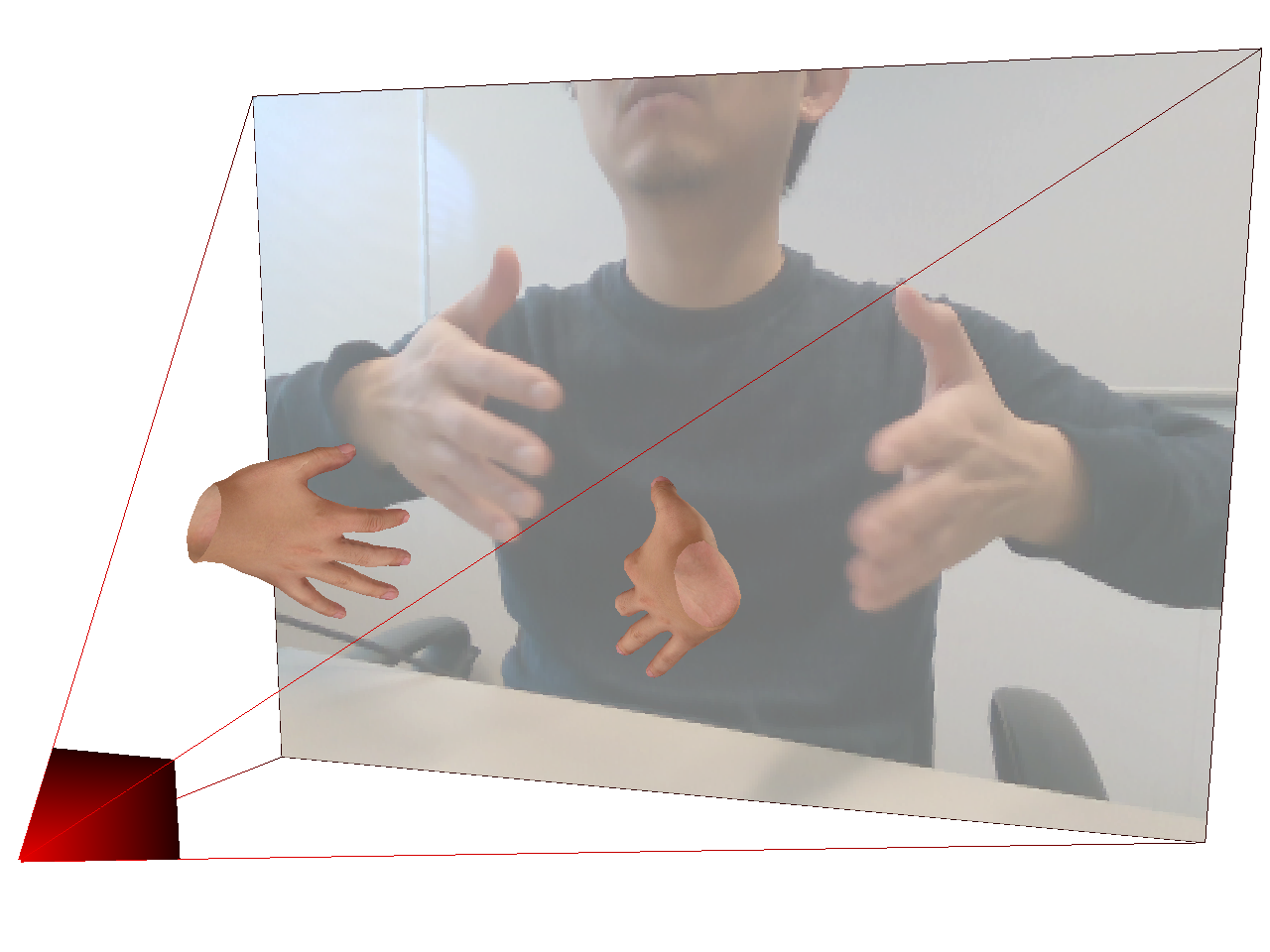}
	}
	\subfigure[With intra-hand distance]{
		\includegraphics[trim={60pt 90pt 40pt 100pt},clip,width=0.45\columnwidth]{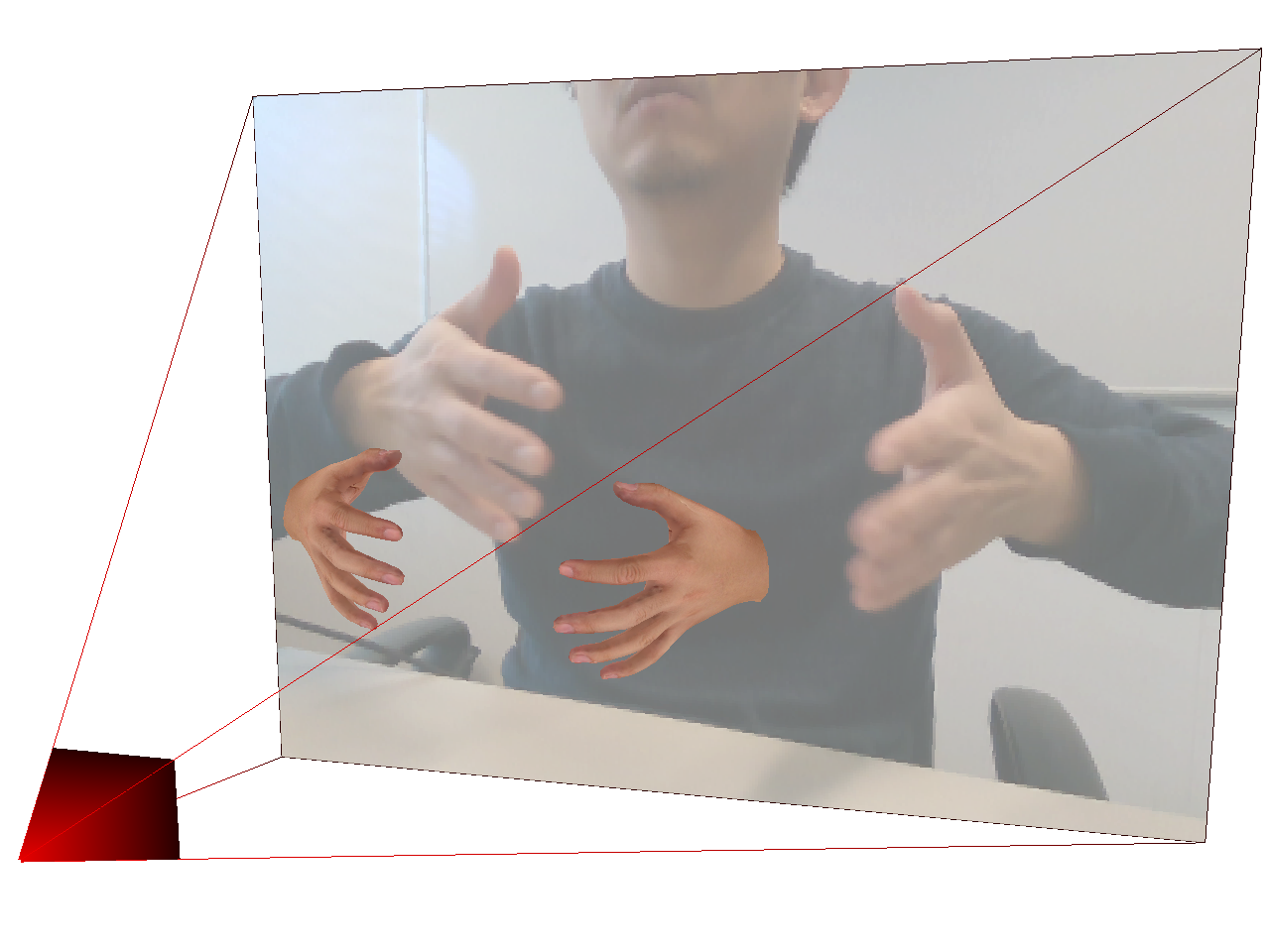}
	}
	\caption{Qualitative results for the energy term ablation study.}
	\label{fig:qualitative-ablation-study}
\end{figure}
For our ablation experiments we consider different settings, which we evaluate based on our \textsc{RGB2Hands} dataset. To be more specific, we analyze the effects of
\begin{enumerate}[(i)]
    \item the individual terms in our fitting energy $f$ in~\eqref{eq:fit},
    \item the importance of using our real and our synthetic dataset.
\end{enumerate}

\paragraph{Fitting Energy Terms:}
In Fig.~\ref{table:ablation_energy} we show PCK curves across all sequences when leaving out one of the terms in our fitting function, compared to using the whole function in~\autoref{eq:fit}. 
All of the terms improve the 3D error. It is notable that the silhouette term improves the 3D error at the cost of 2D keypoint error.
We hypothesize that the energy function without silhouette term has local minima with accurate 2D keypoints, but inaccurate 3D pose, which the silhouette term helps to escape from.
Additionally, in Fig.~\ref{fig:qualitative-ablation-study} and in the supplementary video we present qualitative results of this ablation study.
To this end, we show tracking results with and without individual terms of our optimization problem. 

\paragraph{Importance of Our Datasets:}
Additionally, we have analyzed the behavior of our hand tracker when training our prediction networks either without our real dataset, or without our synthetic dataset, respectively, see Fig.~\ref{fig:ablation_data}. When not using our real dataset, or when omitting our synthetic dataset, the PCK curves drop substantially (see green and orange lines in Fig.~\ref{fig:ablation_data}), compared to using both datasets (blue line).

\subsection{Comparison to Other Methods}
In this section we perform a comparison of our method to existing depth-based as well as RGB-based methods on the \textsc{RGB2Hands} and the \textsc{Tzionas} dataset.
Specifically, for depth-based methods, we show comparisons to \citet{tzionas_ijcv2016} and \citet{mueller_siggraph2019}.
For RGB-based methods, since there is no hand tracking system that was explicitly designed for such input modality
for the scenario of two closely interacting hands,
we show comparisons to the single-hand method by~\citet{Boukhayma_2019_CVPR}. 
For a fair comparison, we follow their procedure of cropping the image around the hand based on OpenPose keypoint predictions~\cite{simon2017hand,cao2018openpose}, and subsequently estimate MANO pose and shape parameters, the 2D location in the image, and the weak-perspective scale.
We apply this approach for each hand independently, horizontally flipping the left hand images since their method was designed for right hands only. Although OpenPose does not respect a valid 3D hand geometry, and merely obtains 2D keypoint positions, for the sake of completeness we also compare to the plain OpenPose predictions.

\paragraph{Comparison on \textsc{Tzionas} Dataset}
In Table~\ref{table:comparison_tzionas} we show quantitative comparisons to \citet{tzionas_ijcv2016}, \citet{mueller_siggraph2019}, \citet{Boukhayma_2019_CVPR}, and OpenPose.
Although in terms of mean error our method performs worse than the depth-based method by Tzionas et al., we emphasize that theirs is an offline method that is about 100 times slower than ours.
However, our result is close to the depth-based real-time method by Mueller et al., despite the fact that they use much richer input data that contains 3D information.
In comparison to the RGB-only method by \citet{Boukhayma_2019_CVPR}, in terms of mean error we achieve results that are on par,
while their method is significantly slower and thereby not real-time capable. 
In contrast to all other methods, the RGB-based OpenPose is trained to regress 2D keypoint locations which exactly matches the evaluated metric and hence yields a better result.
However, we point out that such 2D predictions generally do not represent plausible hand poses, which we will also highlight in the subsequent comparison using our \textsc{RGB2Hands} dataset.
Contrary to our full-frame method, the other two RGB-only methods require bounding boxes to obtain a hand crop.
In consequence, there are 13 frames in the dataset for which no estimates are available due to missing bounding box detection.
Our method also outputs global 3D pose and shape (up to a single scale factor) and runs much faster compared to the other RGB-only methods.
As shown in Fig.~\ref{fig:datasets_IoU}, the \textsc{Tzionas} dataset does not contain many frames with strongly interacting and overlapping hands.
This is the main reason why the evaluated crop-based single-hand RGB methods succeed on this dataset.
The advantages of our method become more apparent when compared on more challenging interaction scenarios, which we present next.

{
 \begin{table*}[t]
 	\caption{
     We compare the mean error and properties of our method to several depth-based and RGB-based hand pose estimation methods on the \textsc{Tzionas} and the \textsc{RGB2Hands} datasets. 
     Note that our method performs on par with the depth-based real-time method by \citet{mueller_siggraph2019}, despite the significantly richer 3D information that the depth-based method uses.
     The other RGB-only methods require hand crops and hence miss frames due to failed detections. 
     Please note that the mean errors are calculated over all detected keypoints and hence do not include a penalty for missed frames.
     Furthermore, the other RGB-only methods are slower and thereby not applicable to the real-time tracking settings. 
     We note that OpenPose offers parameters that enable faster processing. Although in this case it is able to run at 13 FPS, it is significantly less reliable and leads to a total of 156 missed frames on \textsc{RGB2Hands} (opposed to 20 missed frames when running at 2 FPS).
 	}
 	\setlength{\tabcolsep}{6pt}
 	{
 	\centering
 	\begin{tabular}{lccccclll}
 	\toprule
 	\multirow{3}{*}{Method} & \multicolumn{2}{c}{\textsc{Tzionas} Dataset} & \multicolumn{3}{c}{\textsc{RGB2Hands} Dataset} & \multicolumn{3}{c}{Properties} \\ 
 	\cmidrule(lr){2-3} \cmidrule(lr){4-6} \cmidrule(lr){7-9}
 		 & 2D Error & Missed & 2D Error & 3D Error & Missed & \multirow{2}{*}{Input} & \multirow{2}{*}{Output} & Runtime \\
 		 & (pixels) & Frames & (pixels) & (mm) & Frames & & & (ms/frame) \\
 		\midrule
 		\citet{tzionas_ijcv2016} & 5.04  & 0 & - & - & - & Depth & global 3D & 4960 \\
 		\citet{mueller_siggraph2019} & 10.80 & 0 & - & - & - & Depth & global 3D & 33 \\ 
 		\midrule
 		\citet{Boukhayma_2019_CVPR} & 12.91  & 13 & 19.31 & 27.47 & 20 & RGB & weak-persp. 3D & (516) + 16 \\ 
 		OpenPose~\cite{cao2018openpose} & 9.68  & 13 & 13.32 & - & 20 & RGB & 2D keypoints &  516 \\ 
 	    Ours & 13.31  & 0 & 13.43 & 20.02 & 0 & RGB & global 3D (up to scale) & 47 \\
 		\bottomrule
 	\end{tabular}
 	}
 	\setlength{\tabcolsep}{6pt}
 	\label{table:comparison_tzionas}
 \end{table*}
}

\newcommand{\figwidth}{0.185\textwidth}
\begin{figure*}
%
%
\minipage{0.02\textwidth}
\rotatebox{90}{OpenPose}
\endminipage
\minipage{\figwidth}
\includegraphics[width=\linewidth]{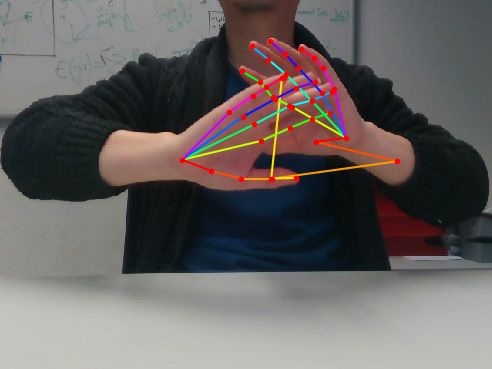}
\endminipage\hfill
\minipage{\figwidth}
\includegraphics[width=\linewidth,draft=false]{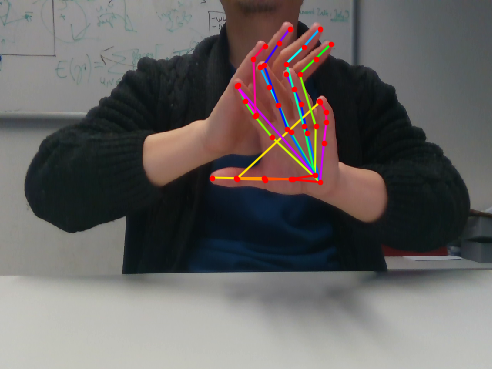}
\endminipage\hfill
\minipage{\figwidth}
\includegraphics[width=\linewidth,draft=false]{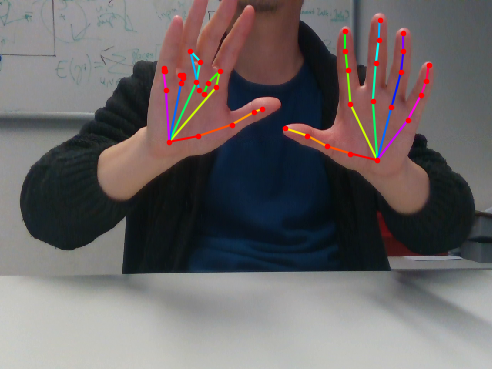}
\endminipage\hfill
\minipage{\figwidth}
\includegraphics[width=\linewidth]{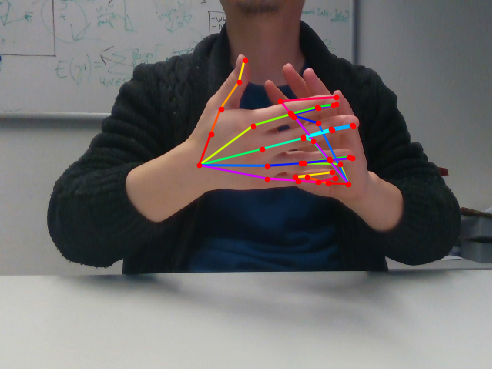}
\endminipage\hfill
\minipage{\figwidth}
\includegraphics[width=\linewidth]{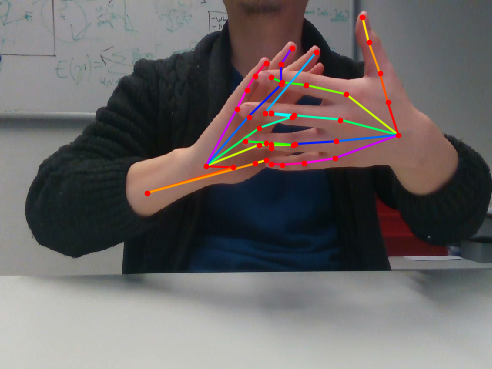}
\endminipage\hfill
%
%
\minipage{0.02\textwidth}
\rotatebox{90}{Boukhayma'19}
\endminipage
\minipage{\figwidth}
\includegraphics[width=\linewidth]{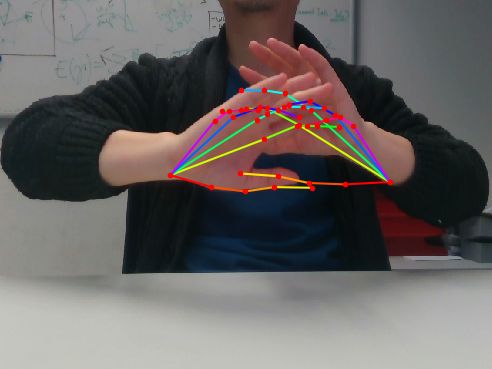}
\endminipage\hfill
\minipage{\figwidth}
\includegraphics[width=\linewidth,draft=false]{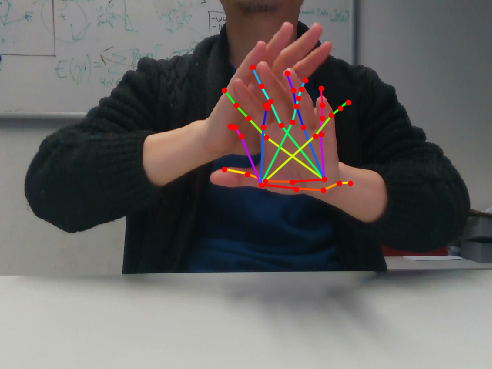}
\endminipage\hfill
\minipage{\figwidth}
\includegraphics[width=\linewidth,draft=false]{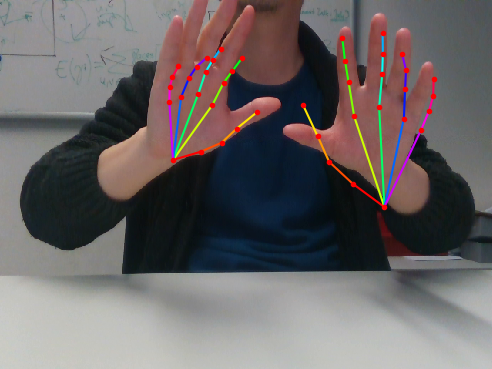}
\endminipage\hfill
\minipage{\figwidth}
\includegraphics[width=\linewidth]{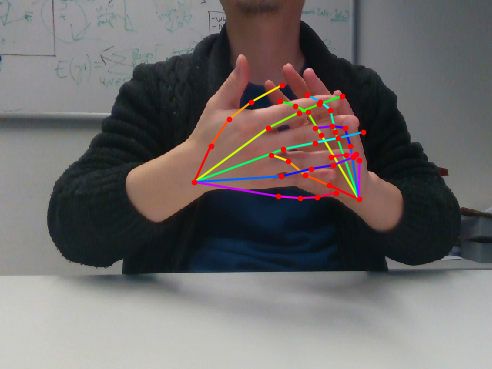}
\endminipage\hfill
\minipage{\figwidth}
\includegraphics[width=\linewidth]{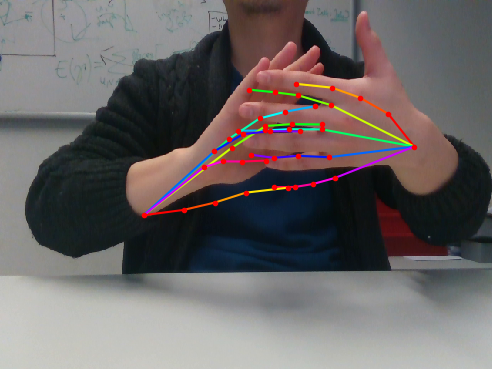}
\endminipage\hfill
\minipage{0.02\textwidth}
\rotatebox{90}{RGB2Hands}
\endminipage
\minipage{\figwidth}
\includegraphics[width=\linewidth]{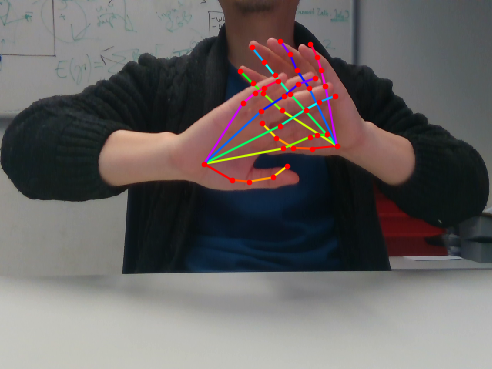}
\endminipage\hfill
\minipage{\figwidth}
\includegraphics[width=\linewidth,draft=false]{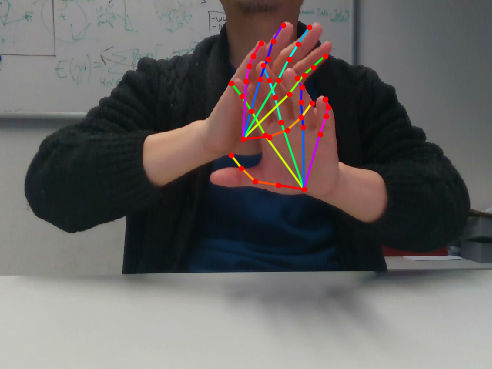}
\endminipage\hfill
\minipage{\figwidth}
\includegraphics[width=\linewidth,draft=false]{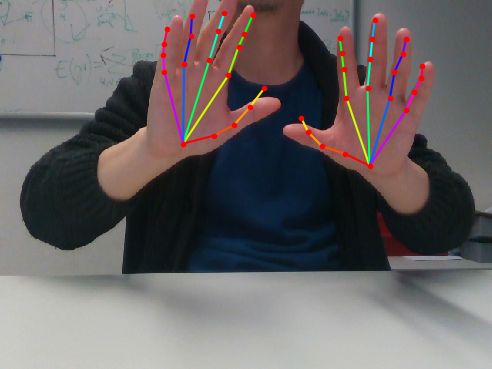}
\endminipage\hfill
\minipage{\figwidth}
\includegraphics[width=\linewidth]{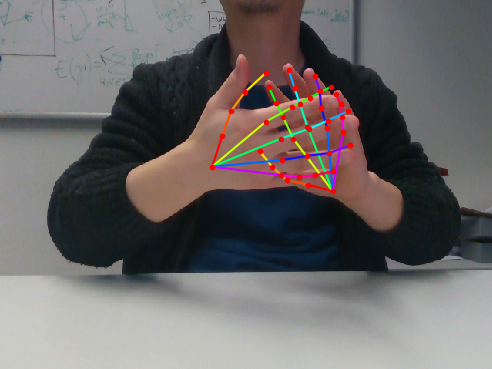}
\endminipage\hfill
\minipage{\figwidth}
\includegraphics[width=\linewidth]{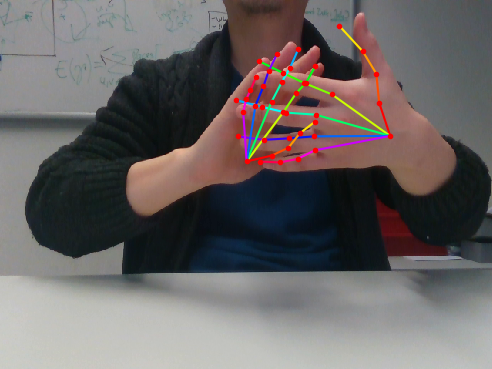}
\endminipage\hfill
\caption{Qualitative comparison of our RGB2Hands approach to \citet{Boukhayma_2019_CVPR} and OpenPose.}
\label{fig:comparison_qual} 
\end{figure*}

\paragraph{Comparison on \textsc{RGB2Hands} Dataset}
We created the \textsc{RGB2Hands} dataset to enable evaluation of more challenging hand interactions than previously seen in other datasets.
In Fig.~\ref{fig:comparison_quant} we show quantitative results, where it can be seen that our method (blue line) leads to  substantially better PCK curves than the method by \citet{Boukhayma_2019_CVPR} (orange line).
Although OpenPose appears to produce good results in terms of the percentage of correct individual keypoints (Fig.~\ref{fig:comparison_quant}, solid line), its percentage of correct frames (PCF), where a frame is considered correct if the maximum keypoint error is under a threshold, is substantially lower compared to others (Fig.~\ref{fig:comparison_quant}, dotted line).
This confirms that OpenPose is often accurate for some of the keypoints in a frame while producing large errors for harder (e.g., occluded) keypoints in the same frame.
This in turn is a strong indicator that the predicted 2D hand keypoints do not constitute a plausible hand pose due to the missing 3D model constraint.
This can also be seen in Fig.~\ref{fig:comparison_qual}, where we show qualitative results.
In addition, OpenPose and hence also the method by \citet{Boukhayma_2019_CVPR} fail to detect the hands completely in several frames (see Table~\ref{table:comparison_tzionas}).
Lastly, since competing methods do not perform temporal filtering, we show that our method without temporal smoothing (``w.o. Smoothing'' in Fig.~\ref{fig:comparison_quant}) still outperforms the competitors.

This evaluation on our $\textsc{RGB2Hands}$ dataset validates the need for methods that are specifically tailored to handle two strongly interacting hands.
Running single-hand methods on crops of the two hands individually cannot jointly reason about the two hands, which is crucial for effectively dealing with close interactions.

\begin{figure*}[t]
    \includegraphics[width=\linewidth,  draft=false,trim={0 2cm 0 0},clip]{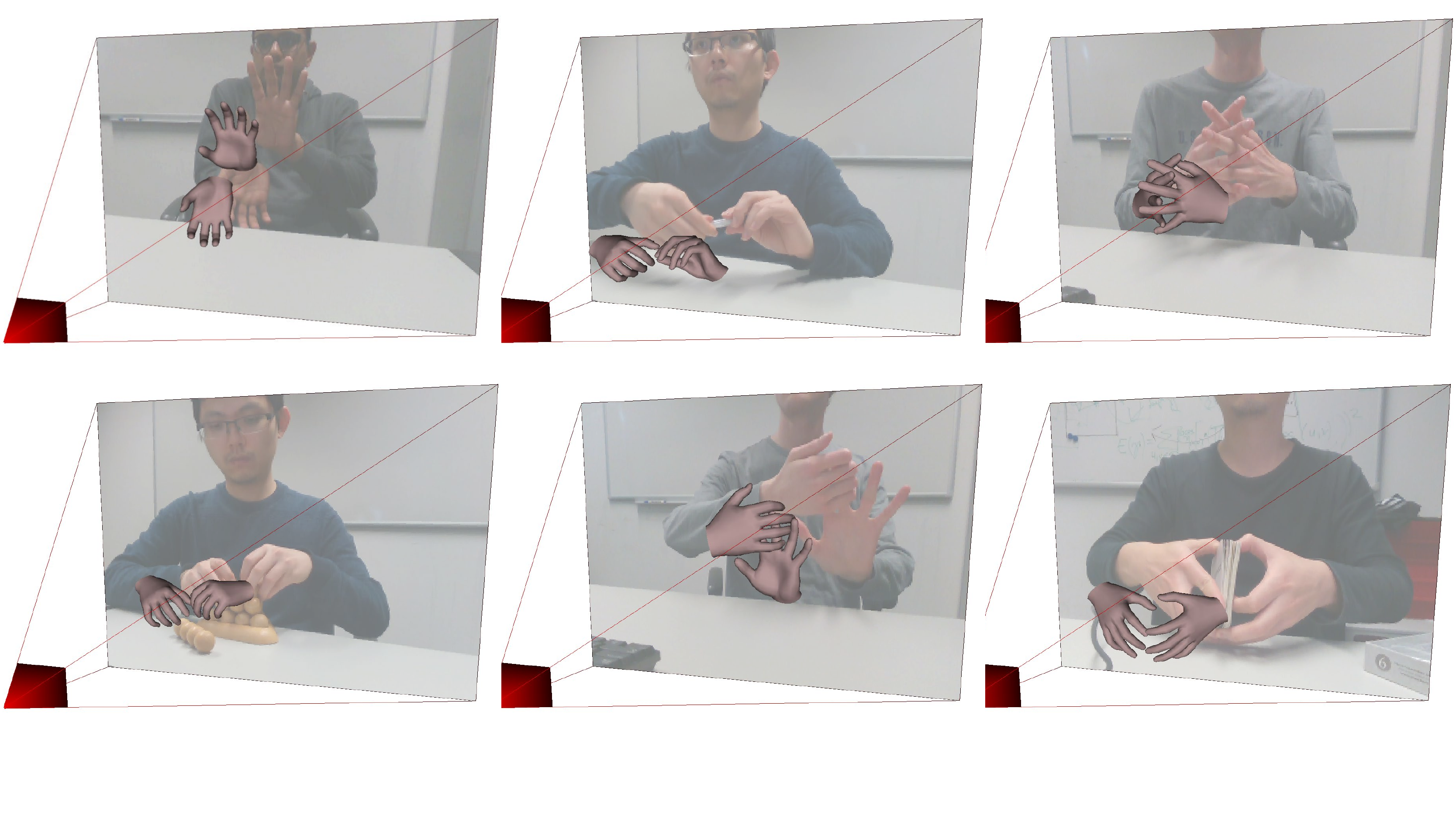}
    \caption{Additional results of our RGB2Hands method.}
    \label{fig:additional_results} 
    \vspace{0.35cm}
\end{figure*}

\begin{figure*}[t]
    \includegraphics[width=\linewidth,trim={0 5cm 0 0},clip,draft=false]{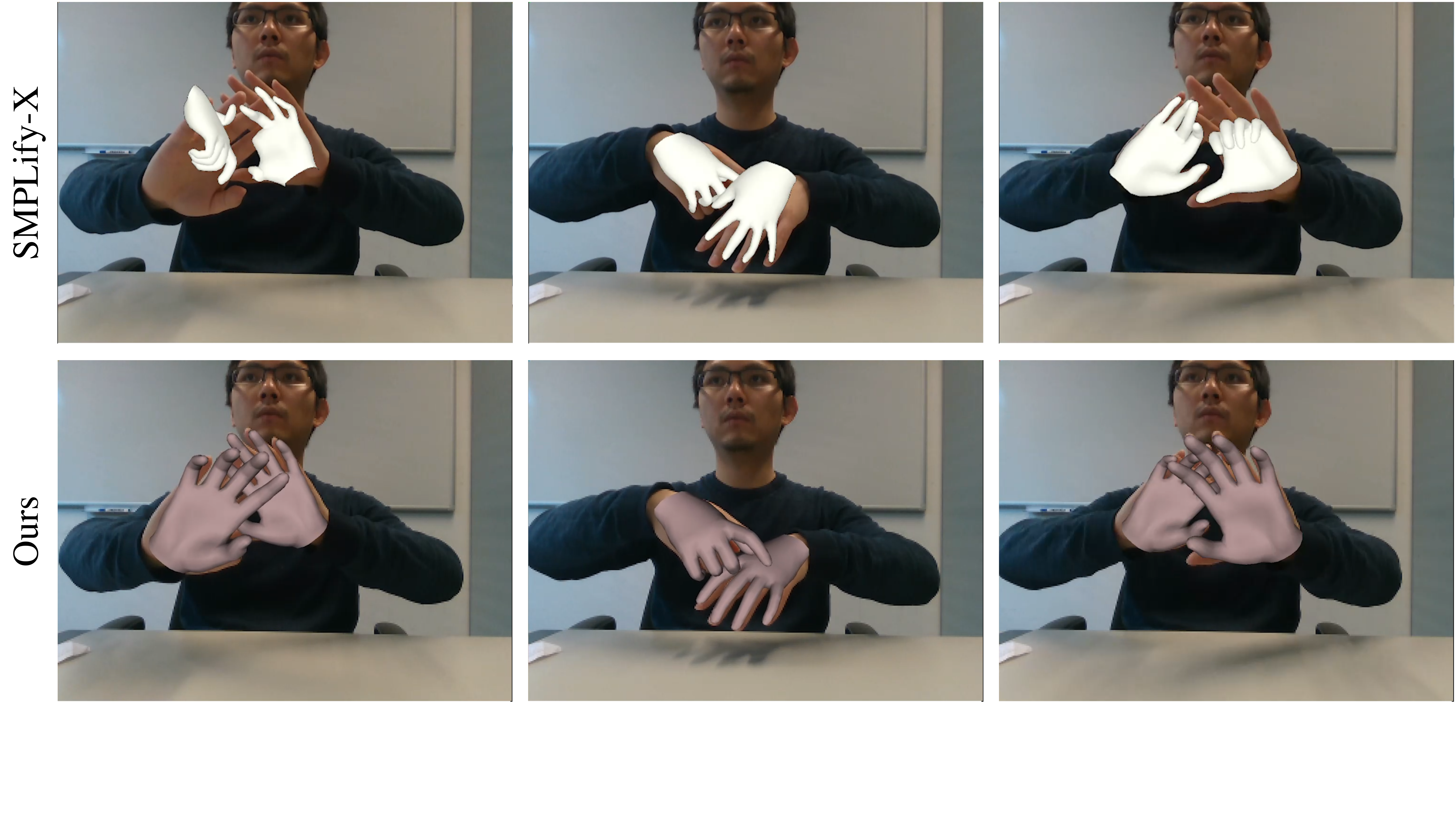}
    \caption{
    Qualitative comparison to SMPLify-X \cite{pavlakos2019SMPLX}.
    SMPLify-X fails when not the entire body is visible since it fits a full body model which implicitly regularizes the rigid hand pose (left).
    However, even when the hand orientation is estimated correctly, it still fails because it does not explicitly consider overlapping and interacting hands.
    Note that we render only the hand vertices of the full body mesh model estimated by SMPLify-X.
    }
    \label{fig:comp_smplx} 
\end{figure*}

\paragraph{Qualitative Comparison to SMPLify-X} 
In Fig.~\ref{fig:comp_smplx}, we compare qualitatively to SMPLify-X \cite{pavlakos2019SMPLX} which fits a full human body model to monocular RGB images.
Such methods rely on the estimated body pose to detect the hand and to regularize the hand orientation.
As such, our method is more stable when the body is not fully visible. 
SMPLify-X does not explicitly address overlapping or interacting hands and hence also fails when the hand detection and orientation are correctly estimated.

\subsection{Additional Qualitative Results}
\begin{figure}[t]
	\centering
		\subfigure{
		\includegraphics[trim={60pt 130pt 220pt 220pt},clip,width=0.45\columnwidth]{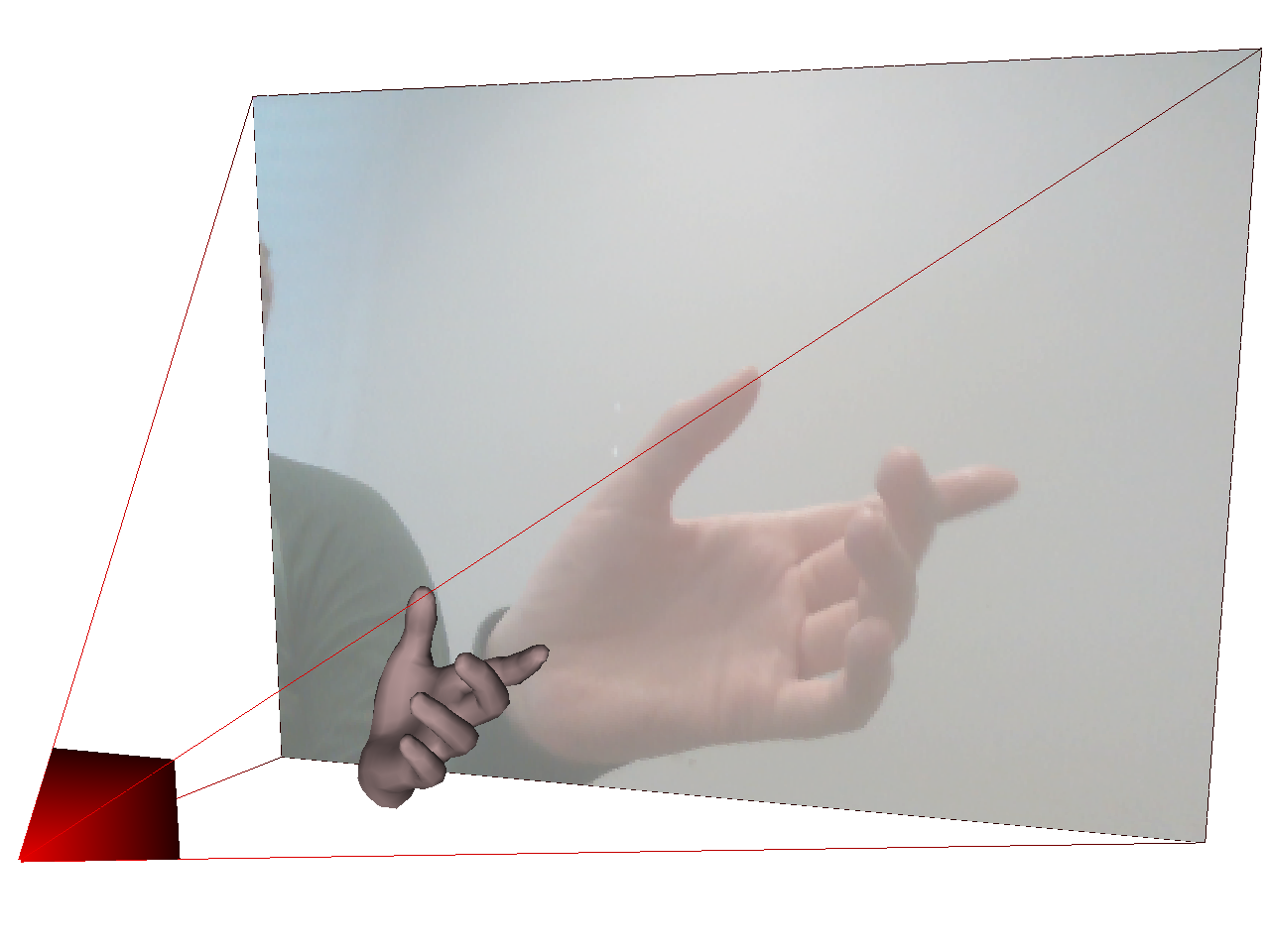}
	}
	\subfigure{
		\includegraphics[trim={60pt 130pt 220pt 220pt},clip,width=0.45\columnwidth]{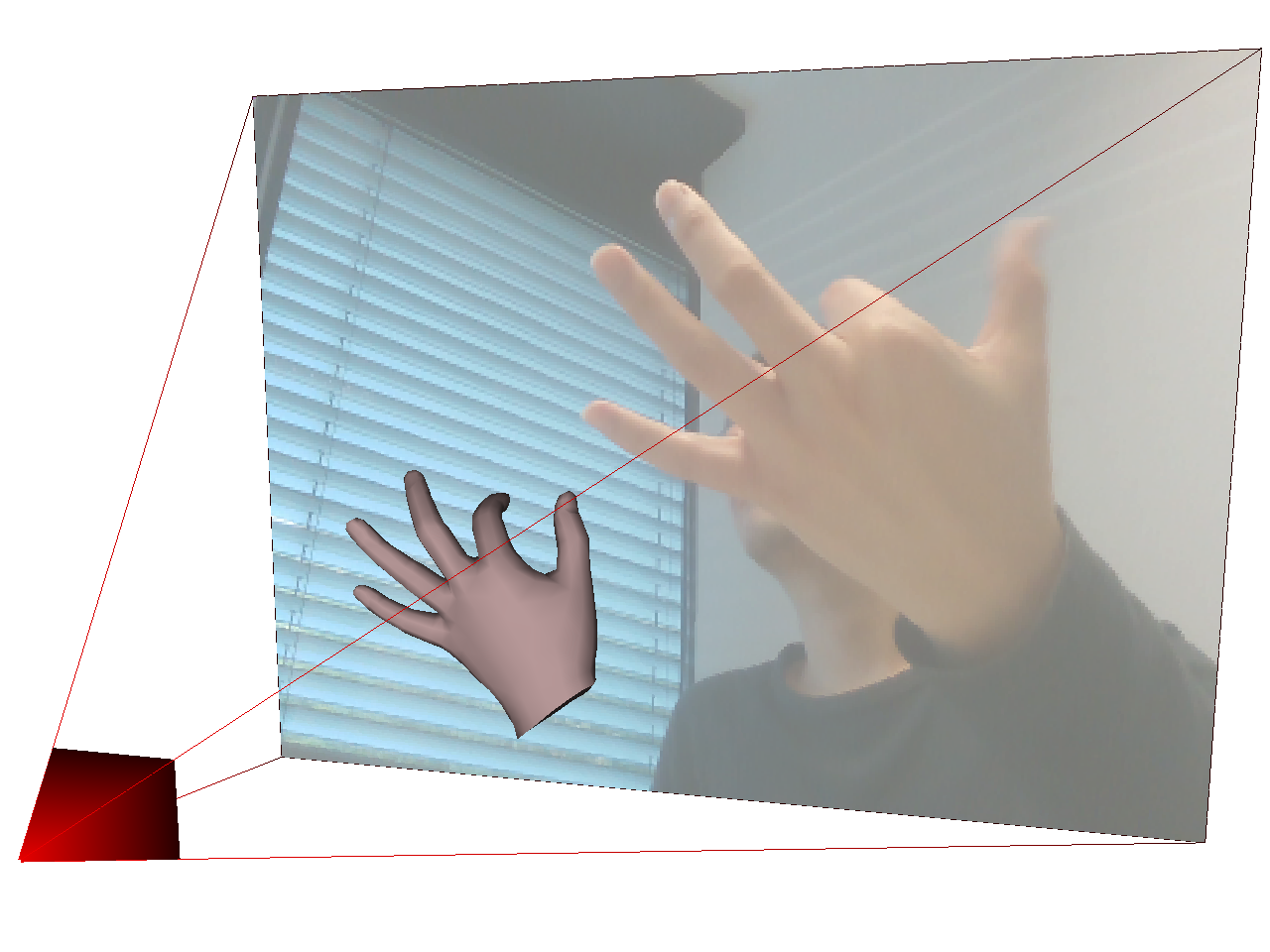}
	}
	\subfigure{
		\includegraphics[trim={60pt 150pt 220pt 220pt},clip,width=0.45\columnwidth]{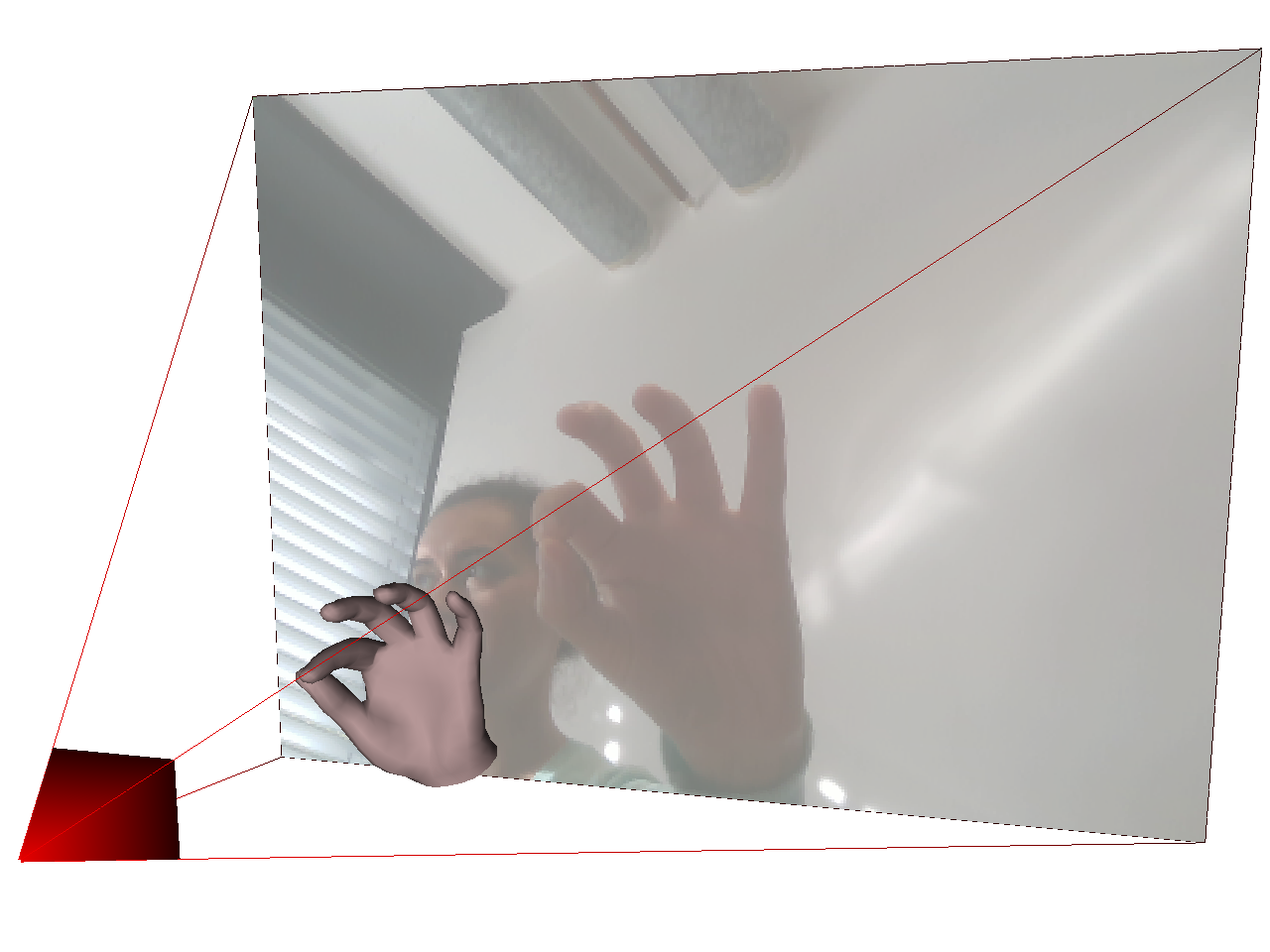}
	}
	\subfigure{
		\includegraphics[trim={60pt 150pt 220pt 220pt},clip,width=0.45\columnwidth]{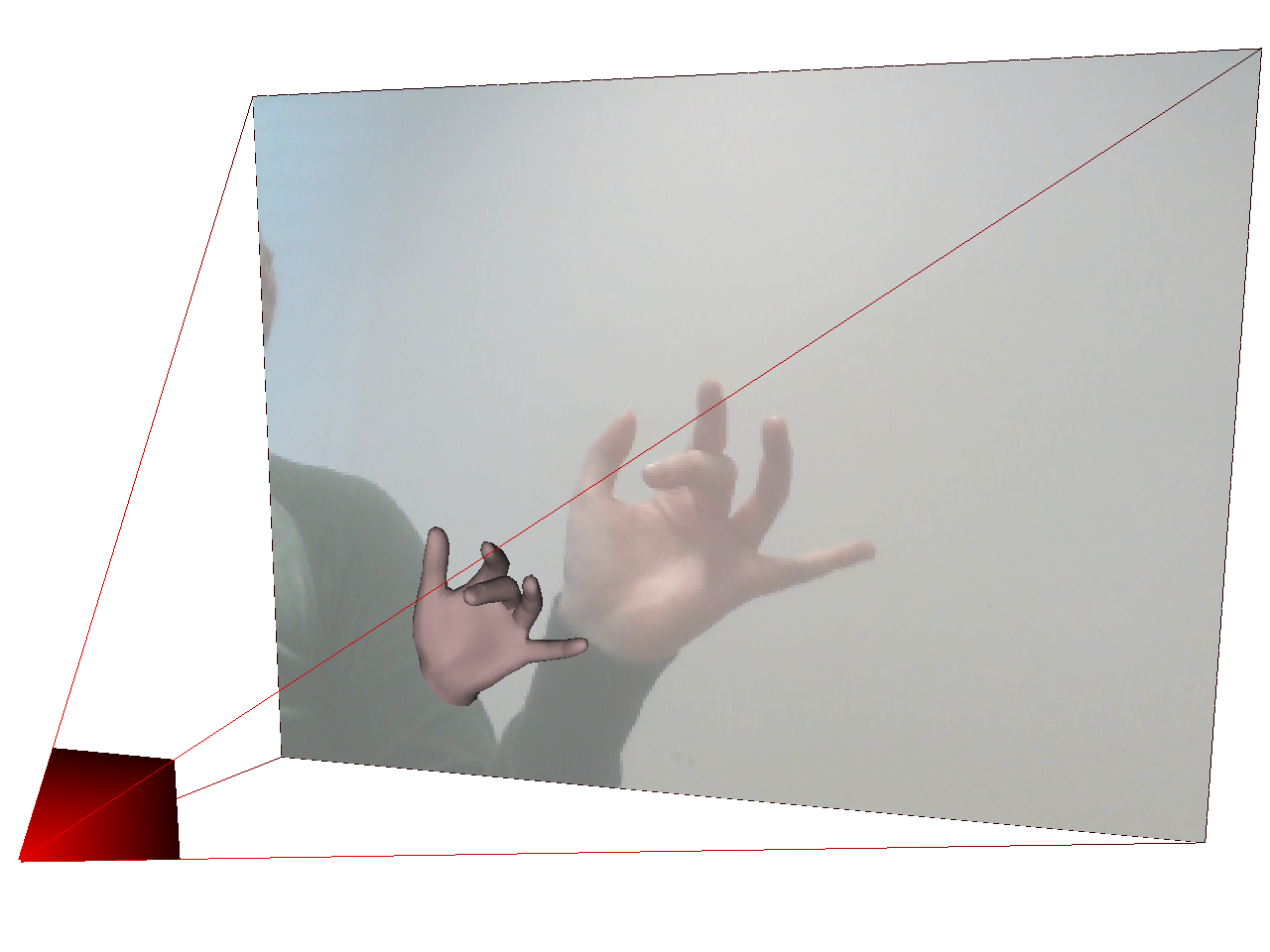}
	}
	\subfigure{
		\includegraphics[trim={60pt 150pt 220pt 220pt},clip,width=0.45\columnwidth]{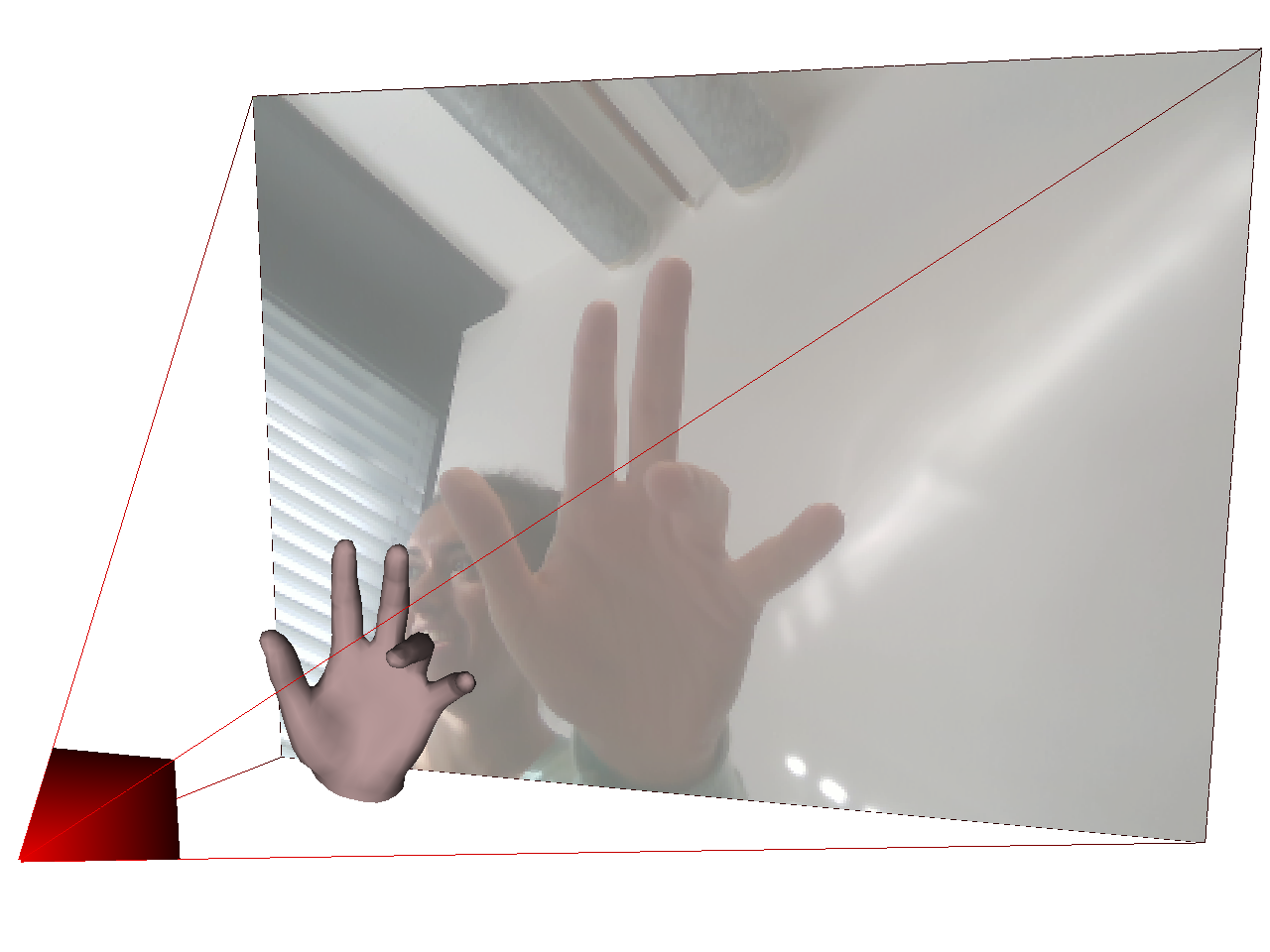}
	}
	\subfigure{
		\includegraphics[trim={60pt 150pt 220pt 220pt},clip,width=0.45\columnwidth]{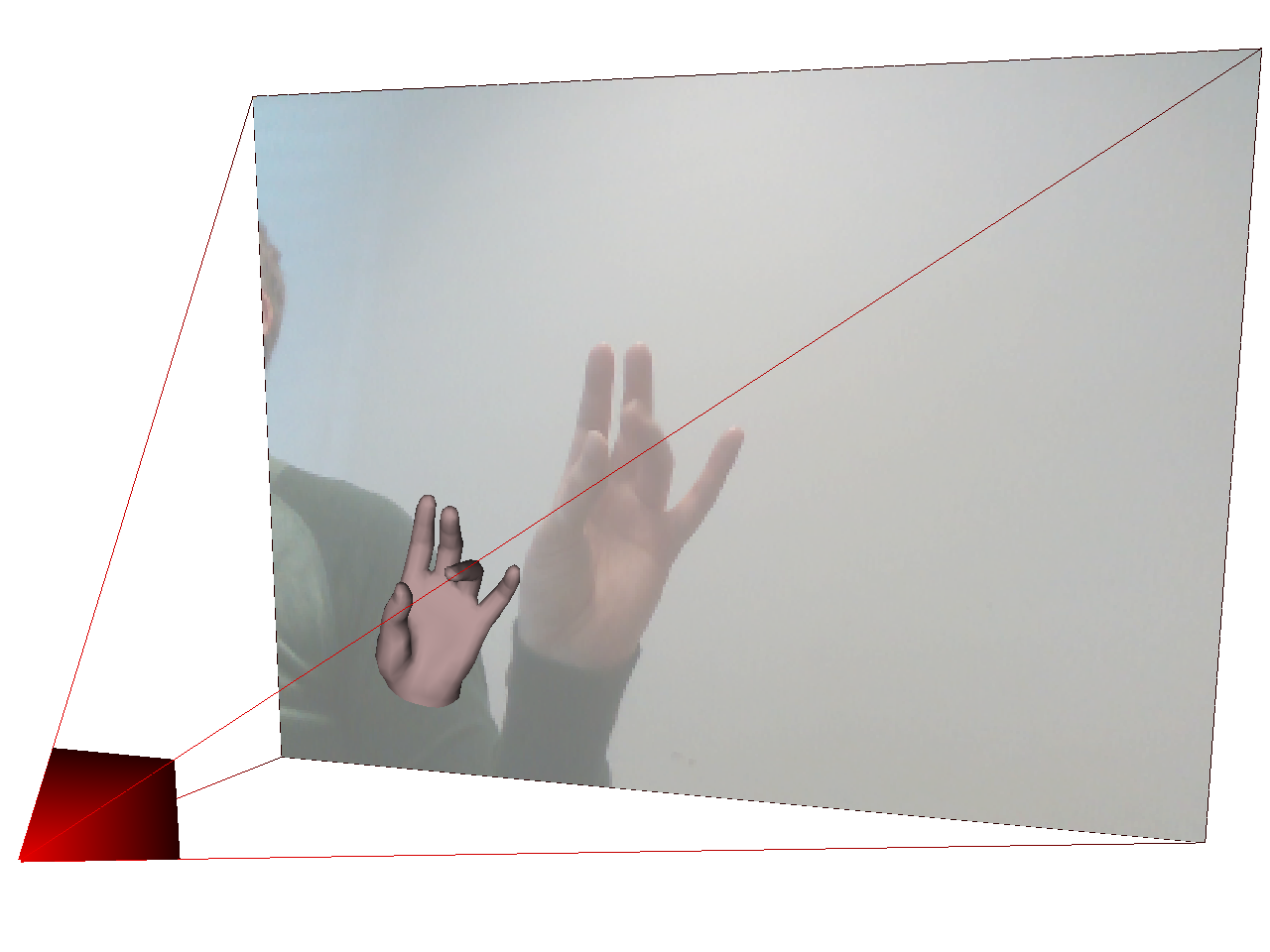}
	}
	\caption{Results of our RGB2Hands method on single hand scenes.}
	\label{fig:additional_results_single_hand}
\end{figure}

Next, we demonstrate the global 3D tracking of two interacting hands in various involved settings.
The purpose of this section is to demonstrate the generality and the wide scope of hand tracking scenarios and non-trivial two-hand interactions that our method is capable of handling in real time. We also show results on single hand scenes to emphasize that our formulation does not require both hands to be present.
The results are shown in Fig.~\ref{fig:teaser}, Fig.~\ref{fig:additional_results},and Fig.~\ref{fig:additional_results_single_hand} as well as the supplementary video.

\section{Discussion \& Future Work}
Overall we have presented compelling 3D tracking and reconstruction results on challenging sequences of two interacting hands. 
One important property of our approach is that it directly works on the full input image, rather than explicitly localizing a hand first, and then using a cropped image for further processing. 
This is in contrast to existing single-hand methods, both RGB-based and depth-based, which could in principle also be applied to the tracking of two hands (by localizing and processing each hand individually).
However, these methods oftentimes fail in the case of heavy hand-hand interactions, since in this case it is not possible to obtain a reliable crop, or the visibility of parts of the other hand lead to errors due to severe self-similarities. 

\begin{figure}[t]
	\centering
	\subfigure{
		\includegraphics[trim={10pt 10pt 10pt 0},clip,width=0.45\columnwidth]{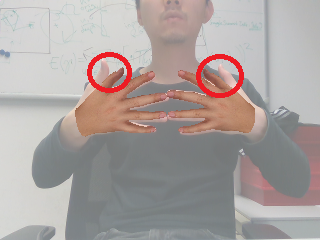}
	}
	\subfigure{
		\includegraphics[trim={40pt 30pt 30pt 20pt} ,clip,width=0.45\columnwidth]{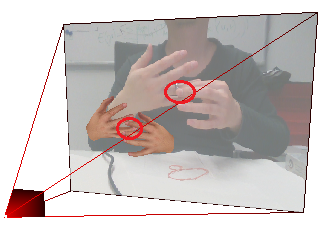}
	}
	\caption{Example Failure Cases}
	\label{fig:FailureCases}
\end{figure}

Despite the overall good performance of our method, particularly for close hand-hand interaction settings, there are also some downsides that we aim to address in the future. Currently, our method may not always be able to correctly track very fast hand motions, since in this case motion blur may lead to unreliable predictions of the neural network. One potential way to address this is to also include data with simulated motion blur, so that the neural network is able to deal with such cases. Moreover, 
it is difficult to find a good trade-off between the MANO pose prior and the other energy terms, so that one has to sacrifice either pose variability or pose plausibility. This is most noticeable for thumb articulations (Fig.~\ref{fig:FailureCases}, left). This could for example be addressed by equipping the MANO model with a kinematic skeleton, and then enforcing explicit joint limit constraints while still using the pose space to capture correlations in joint articulations. 
Due to inherent depth ambiguity, our method may also have difficulties reconstructing interactions where high precision in relative hand positioning is required; e.g. slotting a ring onto a finger (see Fig.~\ref{fig:FailureCases}, right). For such tasks, additional cues from a depth sensor or a stereo camera might be requires.
It would be interesting as well to explore the explicit use of the temporal dimension, so that for example hand shape information can be integrated over time, in a similar spirit to bundle adjustment in multi-view reconstruction.
Moreover, temporal neural network architectures can be used to obtain temporally smoother predictions and thus further improve temporal tracking consistency. Another open point is optimizing for person-specific hand textures based on a parametric hand texture space.

\section{Conclusion}
We have presented the first approach that is specifically tailored towards tracking and reconstruction of two hands in interaction in global 3D from only RGB images. 
A major challenge in this setting are depth ambiguities, which we have addressed by combining two strong priors, one in form of a parametric 3D hand model, and the other one in form of a multi-task neural network predictor that is trained based on a large body of real and synthetic training data. 
For training, we have proposed to combine existing datasets with two datasets that we created specifically for our task. 
The first one is a real dataset for which (potentially noisy) annotations were obtained based on RGB-D frames.
It is complemented by a new synthetic dataset that models physically correct hand interactions while taking hand variability in terms of shape and appearance into account. 
Moreover, we have introduced a new benchmark dataset, \textsc{RGB2Hands}, that contains annotated sequences showing significantly stronger interactions between two hands in comparison to previous benchmarks. 
We demonstrated that our proposed approach outperforms previous RGB-only methods in complex hand-hand interaction settings, both quantitatively and qualitatively, and even performs on par with a state-of-the-art depth-based real-time approach. 
In summary, we have presented the first real-time approach that captures 3D motion and geometry of difficult two-hand interactions from monocular color video.

\begin{acks}
The work was supported by the ERC Consolidator Grants 4DRepLy (770784) and TouchDesign (772738) and Spanish Ministry of Science (RTI2018-098694-B-I00
VizLearning).
\end{acks}

\bibliographystyle{ACM-Reference-Format}
\bibliography{references}

\appendix

\section{Annotation Transfer from Depth to RGB}
\label{appendix:loss_mask}

Here, we describe how we transfer annotations from the depth image to the RGB image, where both were acquired with a calibrated RGB-D camera (cf.~Sec.~\ref{sec:training_data}). To this end, based on camera calibration parameters of the depth and RGB cameras, we first transfer points on the depth image to 3D, and then project these onto the RGB image. 
However, discrepancies exist between the two images due to an offset between the sensors (baseline), and due to object boundary artifacts of the structured-light depth sensor.
To prevent the neural network from learning this, we employ a foreground mask in the respective loss.
The final foreground mask is based on computing the intersection of two masks; a foreground mask obtained from the depth image, and a foreground mask that is obtained from the RGB image. For obtaining the former, the depth image is first thresholded, and subsequently processed by a morphological closing operation to correct for boundary artifacts.
The foreground mask of the RGB image is obtained by color thresholding, which is straightforward due to our green-screen setup. By using such a mask, for training the multi-task CNN we only penalize those regions that we trust, while not penalizing regions with potentially missing annotations.

\section{Data Augmentation}
\label{appendix:data_aug}

Our real dataset contains sequences of single-hand tracking scenarios with a green-screen to allow background augmentation.
Additionally, a color-based segmentation is performed to separate hand and body.
For compositing two-hand scenes, two masked hand images from the same person are chosen at random, and flipped accordingly to obtain a pair of left and right hands. The hands are then independently translated from their initial location. Whenever the two hands are overlapping, the depth channel is taken into account to ensure a plausible occlusion when the hands project onto the same location in the image. 

\section{Hyperparameters}
\label{sec:hyperparams}

In the following, we provide the values of hyperparameters:
$\lambda_{\text{d}}=0.003$,
$\lambda_{\text{sil}}=0.0045$,
$\lambda_{\text{key}}=0.005$,
$\lambda_{\text{intra}}=0.3$,
$\lambda_{\text{inter}}= 0.1$,
$\lambda_{\beta} = 0.025$,
$\lambda_{\theta} = 0.0375$,
$\lambda_{\tau} = 0.3$,
$\lambda_{\text{sym}} = 0.5$,
$\lambda_{\mathcal{N}} = 4.6 \cdot 10^5$,
$\lambda_{s} = 10^3$,
$t_{\theta} = 0.1$,
$t_r = 2.3$,
$t_c = 0.04$,
$t_h = 0.7$,
$\mu = 1$.

\section{Automatic Error Recovery}
Due to the severe ambiguities in monocular RGB data, like the concave-convex ambiguity, the optimization of the parameters $\bm{\nu}$
does not always yield a correct result.
Instead, fingers might get bent in the wrong direction to fit the projection in the input. These poses are difficult to escape using a local optimizer.
Hence we detect these unnatural poses by observing the magnitude of $\bm{\theta}_h, h \in \{\text{left, right}\}$.
If this magnitude exceeds the threshold $t_r$, we reset the pose parameters to the closest pose that is still within the threshold.

\end{document}